\documentclass[review,5p,times,twocolumn]{elsarticle}

\usepackage{mypreamble}
\newacronym{absa}{ABSA}{Aspect-based Sentiment Analysis}
\newacronym{nlp}{NLP}{Natural Language Processing}
\newacronym{sa}{SA}{Sentiment Analysis}
\newacronym{asa}{ASA}{Arabic Sentiment Analysis}
\newacronym{dl}{DL}{Deep Learning}
\newacronym{ml}{ML}{Machine Learning}
\newacronym{ca}{CA}{Classical Arabic}
\newacronym{msa}{MSA}{Modern Standard Arabic}
\newacronym{da}{DA}{Dialectal Arabic}
\newacronym{ea}{E\#A}{Entity-Attribute}
\newacronym{ote}{OTE}{Opinion Target Expression}
\newacronym{slr}{SLR}{Systematic Literature Review}
\newacronym{acd}{ACD}{Aspect Category Detection}
\newacronym{asc}{ASC}{Aspect Sentiment Classification}
\newacronym{semeval}{SemEval}{Semantic Evaluation}
\newacronym{word2vec}{Word2vec}{word-to-vector}
\newacronym{CBOW}{CBOW}{continuous bag-of-words}
\newacronym{glove}{GloVe}{global vector}
\newacronym{oov}{OOV}{out-of-vocabulary}
\newacronym{bi-lstm}{Bi-LSTM}{bi-directional long-short-term memory}
\newacronym{lstm}{LSTM}{long-short-term memory}
\newacronym{gru}{GRU}{Gated-Recurrent Unit}
\newacronym{bi-gru}{Bi-GRU}{Bi-Directional Gated-Recurrent Unit}
\newacronym{r-cnn}{R-CNN}{Regional Convolutional Neural Network}
\newacronym{rcnn}{R-CNN}{Regional Convolutional Neural Network}

\newacronym{elmo}{ELMo}{Embeddings from Language Model}
\newacronym{gpt}{GPT}{Generative Pre-trained Transformer}
\newacronym{ulmfit}{ULMFiT}{Universal Language Model Fine-tuning for Text Classification}
\newacronym{bert}{BERT}{Bidirectional Encoder Representations from Transformers}
\newacronym{t5}{T5}{Text-To-Text Transfer Transformer}
\newacronym{mlm}{MLM}{Masked Language Model}
\newacronym{nsp}{NSP}{Next Sentence Prediction}
\newacronym{rnn}{RNN}{Recurrent Neural Network}
\newacronym{cnn}{CNN}{Convolutional Neural Network}
\newacronym{mbert}{mBERT}{Multilingual BERT}
\newacronym{svm}{SVM}{Support Vector Machine}
\newacronym{crf}{CRF}{Conditional Random Field}
\newacronym{haad}{HAAD}{Human-Annotated Arabic Dataset}
\newacronym{labr}{LABR}{Large-Scale Dataset of Arabic Book Reviews}
\newacronym{hard}{HARD}{Hotel Arabic-Reviews Dataset}

\newacronym{ian}{IAN}{Interactive Attention Network}
\newacronym{indylstm}{IndyLSTM}{Independent Long Short-Term Memory}
\newacronym{pos}{POS}{Part-Of-Speech}
\newacronym{gun}{GUN}{Gated Unit Network}
\newacronym{tha}{THA}{Truncated History Attention}
\newacronym{ner}{NER}{Named Entity Recognition}
\newacronym{ctrl}{Ctrl}{Controlled Convolutional Neural Network}
\newacronym{de}{DE}{domain-Specific word embedding}
\newacronym{tdlstm}{TD-LSTM}{Target-Dependent LSTM}
\newacronym{tclstm}{TC-LSTM}{Target-Connection LSTM}
\newacronym{aelstm}{AE-LSTM}{Aspect Embedding with LSTM}
\newacronym{ataelstm}{ATAE-LSTM}{Aspect Embedding with Attention-based LSTM}
\newacronym{ram}{RAM}{Recurrent Attention on Memory}
\newacronym{aoa}{AOA}{Attention-over-Attention}
\newacronym{mgan}{MGAN}{Multi-Grained Attention Network}
\newacronym{tanh}{Tanh}{Hyperbolic Tangent}
\newacronym{relu}{ReLU}{Rectified Linear Unit}
\newacronym{igcn}{IGCN}{Interactive Gated Convolutional Network}
\newacronym{aen}{AEN}{Attention Encoder Network}
\newacronym{spc}{SPC}{Sentence Pair Classification}
\newacronym{intra-mha}{Intra-MHA}{Intra-Multi Head Attention}
\newacronym{inter-mha}{Inter-MHA}{Inter-Multi Head Attention}
\newacronym{pct}{PCT}{Pointwise Convolutional Transformer}
\newacronym{mha}{MHA}{Multi-Head Attention}
\newacronym{qa}{QA}{Question-Answering}
\newacronym{nli}{NLI}{Natural Language Inference}
\newacronym{sdgcn}{SDGCN}{Sentiment Dependency Graph Convolutional Network}
\newacronym{gcn}{GCN}{Graph Convolutional Network}
\newacronym{mhsa}{MHSA}{Multi-Head Self-Attention}
\newacronym{cdm}{CDM}{Context Dynamic Mask}
\newacronym{cdw}{CDW}{Context Dynamic Weights}
\newacronym{srd}{SRD}{Semantic-Relative Distance}
\newacronym{roberta}{RoBERTa}{Robustly Optimized BERT Pretraining Approach}
\newacronym{deberta}{DeBERTa}{Decoding-enhanced BERT with Disentangled Attention}
\newacronym{lsa}{LSA}{Local Sentiment Aggregation}
\newacronym{imn}{IMN}{Interactive Multi-Task Learning Network}
\newacronym{squad}{SQuAD}{Stanford Question Answering Dataset}
\newacronym{ate}{ATE}{Aspect Term Extraction}
\newacronym{lcf-atepc}{LCF-ATEPC}{Local Context Focus-Aspect Term Extraction and Polarity Classification}
\newacronym{m-t5}{m-T5}{Multilingual T5}
\newacronym{dapt}{DAPT}{Domain Adaptive Pretraining}
\newacronym{indapt}{In-DAPT}{In-Domain Adaptive Pretraining}
\newacronym{outdapt}{Out-DAPT}{Out-of-Domain Adaptive Pretraining}
\newacronym{tapt}{TAPT}{Task Adaptive Pretraining}
\newacronym{tf}{TF}{term frequency}
\newacronym{tfidf}{TF-IDF}{term frequency-inverse document frequency}
\newacronym{bi-indylstm}{Bi-IndyLSTM}{Bi-Directional Independent Long Short-Term Memory}

\newacronym{lcf-bert}{LCF-BERT}{Local Context Focus BERT}

\newacronym{tp}{TP}{True Positive}
\newacronym{tn}{TN}{True Negative}
\newacronym{fp}{FP}{False Positive}
\newacronym{fn}{FN}{False Negative}
\newacronym{ir}{IR}{Imbalance Ratio}

\begin{document}

\begin{frontmatter}

\title{Domain-Adaptive Pre-Training for Arabic Aspect-Based Sentiment Analysis: A Comparative Study of Domain Adaptation and Fine-Tuning Strategies }

%author1
%author1
\author[mymainaddress]{Salha Alyami \corref{mycorrespondingauthor}}
\cortext[mycorrespondingauthor]{Corresponding author.}
\ead{salyami0125@stu.kau.edu.sa}

\author[mymainaddress]{Amani Jamal}

\author[mymainaddress]{Areej Alhothali}

\address[mymainaddress]{ Department of Computer Science, Faculty of Computing and Information Technology, King Abdulaziza University, Jeddah, Saudi Arabia}


\begin{abstract}
Aspect-based sentiment analysis (ABSA) in natural language processing enables organizations to understand customer opinions on specific product aspects. While deep learning models are widely used for English ABSA, their application in Arabic is limited due to the scarcity of labeled data. Researchers have attempted to tackle this issue by using pre-trained contextualized language models, such as BERT. However, these models are often based on fact-based data, which can potentially introduce bias in domain-specific tasks, like ABSA. To our knowledge, no studies have applied adaptive pre-training with Arabic contextualized models for ABSA. This research proposes a novel approach using domain-adaptive pre-training for aspect-sentiment classification (ASC) and opinion target expression (OTE) extraction. We examine various fine-tuning strategies—feature extraction, full fine-tuning, and adapter-based methods—to enhance performance and computational efficiency, utilizing multiple adaptation corpora and contextualized models. Our results show that in-domain adaptive pre-training yields modest improvements. Adapter-based fine-tuning is a computationally efficient method that achieves competitive results. However, error analyses reveal considerable issues with model predictions and dataset labeling. In ASC, common problems include incorrect sentiment labeling, misinterpretation of contrastive markers, positivity bias for early terms, and challenges with conflicting opinions and subword tokenization. For OTE, issues involve mislabeling targets, confusion over syntactic roles, difficulty with multi-word expressions, and reliance on shallow heuristics. These findings underscore the need for syntax- and semantics-aware models, such as graph convolutional networks, to more effectively capture long-distance relations and complex aspect-based opinion alignments.}
\end{abstract}

\begin{keyword}
 Aspect-based sentiment analysis
 \sep 
  Arabic sentiment analysis
 \sep Opinion target extraction 
 \sep 
 Feature-based sentiment analysis
 \sep Aspect sentiment classification
 \sep Domain Adaptation

\end{keyword}

\end{frontmatter}

    %! Author = sbbfti
%! Date = 10/06/2020

% Document

\section{Introduction}
\label{sec:introduction}

The rise of reviews and social media platforms has made it easier for individuals to express their opinions and share their experiences with others. This presents an opportunity for businesses to gain insights from these online channels, better understanding their customers' concerns and improving the overall experience~\citep{IntrobusnessImpact}. Similarly, governments can gain valuable insights into public opinion and sentiment on various issues, thereby enhancing their policy-making and decision-making processes~\citep{pozzi2017sentiment}.

One of the \gls{nlp} methods that researchers have employed to analyze individuals' opinions and attitudes is \gls{sa}~\citep{Liu2012,SAecomeerce2019}. \gls{sa} is a specific \gls{nlp} task aimed at extracting expressed opinions from text and categorizing their polarity as positive, negative, neutral, or conflictual~\citep{Liu2012,liu_2020}. This sentiment analysis is further classified into various levels of granularity: document-level, sentence-level, and aspect-level~\citep{Liu2012}. At the document- and sentence- levels, \gls{sa} operates under the assumption that each review or sentence expresses an opinion about a single entity or topic. However, this assumption limits both document- and sentence-level \gls{sa} when faced with scenarios where documents or sentences contain opinions about multiple entities or aspects with differing polarities. In contrast, \gls{absa} can detect various opinions expressed within the same sentence about multiple entities or their aspects~\citep{Liu2012,pontiki-etal-2014-semeval}. This fine-grained analysis provides decision-makers with insights into the specific features or aspects that individuals like or criticize about a given entity.

Researchers have employed various techniques to address \gls{absa} tasks, including rule-based, lexicon-based, machine learning, and deep learning approaches~\cite{Zhou2019,Do2019}. The research community has favored \gls{dl} because it can learn more complex patterns directly from the data without the need for handcrafted features, unlike traditional \gls{ml}-based methods~\cite{Goodfellow-et-al-2016}. However, deep learning techniques require a large amount of labeled data for supervised learning tasks \citep{Goodfellow-et-al-2016}. Thus, most Arabic \gls{absa} studies have used traditional \gls{ml} due to the limited size of Arabic ABSA public benchmark datasets \citep{Obiedat_2021,Alyami2022,Bensoltane2023}.

Unidirectional and bidirectional \gls{rnn}s, including Vanilla \gls{rnn}, \gls{lstm}, and \gls{gru}, are commonly employed deep learning models in Arabic \gls{absa}~\citep{Obiedat_2021, Alyami2022, Bensoltane2023}. However, \gls{rnn}-based architectures face challenges in handling long sentences and are vulnerable to vanishing or exploding gradients~\citep{RNN_training_issues}. Furthermore, their sequential nature prevents parallel processing, resulting in slower training times~\citep{Sendilkkumaar2020aSNAQ}. Many Arabic \gls{absa} models also rely on context-independent word representations such as word2vec and fastText, which often encounter \gls{oov} issues and fail to effectively capture contextual meanings~\citep{miaschi-dellorletta-2020-contextual}. To address these limitations, recent research has explored the use of contextualized pre-trained large language models, such as BERT, which have demonstrated improved performance compared to context-independent \gls{rnn} models~\cite{Bensoltane2022_S50, Abdelgwad2022, Fadel2022}. Notably, studies by~\cite{Abdelgwad2022} and~\cite{Fadel2022} achieved state-of-the-art results in \gls{asc} and \gls{ote} extraction on the \gls{semeval}-2016 Arabic hotel reviews dataset.

Over the last few years, several language models have emerged that are pre-trained on large amount of data to enhance word representation and address the challenge of insufficient task data. Some of these models are \gls{elmo} \cite{Peters2018}, \gls{bert} \cite{devlin-etal-2019-bert}, \gls{gpt} \cite{Radfort2018}, and \gls{ulmfit} \cite{Howard2018}. However, most of these models were trained on fact-based corpora such as Wikipedia, which can introduce bias in tasks that require domain knowledge and opinionated data, such as \gls{absa}~\cite{Xu2019,Rietzler2020,gururangan-etal-2020-dont}. Researchers utilize domain adaptive pre-training, a domain adaptation technique, to mitigate bias. This involves continuing \gls{bert} pre-training on a corpus with the same domain as the end task~\cite{Xu2019},\cite{Rietzler2020},\cite{gururangan-etal-2020-dont}. This approach has proven effective for many English \gls{nlp} tasks, such as \gls{absa}~\citep{Rietzler2020,Xu2019,YANG2021344}.

Although domain adaptation has been shown to be effective in English \gls{absa}, it has not yet been explored in Arabic \gls{absa}. To address this gap,  we present the first comprehensive study on domain adaptation with pre-trained contextualized language models, such as \gls{bert}, for Arabic \gls{absa}. We evaluated two \gls{absa} tasks, \gls{asc} and \gls{ote}, through extensive experiments on various domain-specific corpora and multiple Arabic \gls{bert} models. We also explored the performance of end-task models under various factors, including the overlap between pre-trained model vocabularies and end-task datasets, dialect distribution, and the size of the adaptation corpus.
We apply both full fine-tuning and adapter-based fine-tuning, exploring the trade-off between the number of trainable parameters and performance. Beyond domain adaptation, we investigate multi-task learning with adapter fusion, apply advanced loss functions to handle class imbalance, and conduct error analysis with explainability methods. In general, this work presents the first comprehensive evaluation of domain and task adaptation strategies for Arabic \gls{absa}, demonstrating how domain relevance, corpus size, vocabulary overlap, and fine-tuning approaches can improve model performance across \gls{absa} tasks.

Together, these results constitute the first broad evaluation of domain and task adaptation for Arabic \gls{absa} and clarify how domain match, corpus scale, vocabulary coverage, and the chosen fine-tuning approach combine to improve performance across \gls{asc} and \gls{ote}.

The remainder of this work is structured into the following sections: % \cref{sec:background} provides background information on the Arabic language and aspect-based sentiment analysis. 
\cref{sec:RL} summarizes and discusses the related work. \cref{sec:methodology} describes the methodology used to perform this systematic review. \cref{sec:results} illustrates the findings. \cref{sec:discussion} discusses the findings, highlights the limitations of the primary studies, and identifies future research directions. \cref{sec:limitations} presents the limitation of this work. Finally, \cref{sec:conclusion} presents the conclusion and future work directions.

%The majority of research on ABSA is in English, with a small amount of work available in Arabic. Most previous Arabic research has relied on deep learning models that depend primarily on context-independent word embeddings (e.g. word2vec), where each word has a fixed representation independent of its context. This article explores the modeling capabilities of contextual embeddings from pre-trained language models, such as BERT, and making use of sentence pair input on Arabic aspect sentiment polarity classification task.

   %%\input{sections/Background}
    \section{Literature Review on Aspect-Based Sentiment Analysis (ABSA)}\label{sec:RL}

The magnitude of published research in \gls{absa} was highly limited before the introduction of \gls{absa} tasks at the \gls{semeval} workshop in 2014, particularly for Arabic. To encourage researchers to contribute to the solutions of \gls{absa} tasks, the \gls{semeval} workshop included \gls{absa} as a subtask in 2014, 2015, and 2016. They also introduced various tasks related to \gls{absa} and provided guidelines for annotating \gls{absa} datasets, which led to the release of several benchmark datasets.

In the \gls{absa} research, aspect sentiment is predicted at two levels of granularity. The first level involves predicting sentiment for specific aspect terms, while the second level involves predicting sentiment for both implicit and explicit aspect terms based on predefined aspect categories. For example, in the sentence (e.g., \enquote{But the staff was so nasty to us}), the aspect term is the token \enquote{staff}, while the aspect category is the \enquote{service}. 

In this work, we review ABSA studies that focus \gls{absa} studies that focus on either \gls{asc} or \gls{ote} extraction. We have addressed both aspect sentiment prediction levels within aspect sentiment classification. Additionally, we have reviewed works based on text in both English and Arabic.

\subsection{Arabic Aspect-Based Sentiment Analysis}

Recent reviews on Arabic \gls{absa} have revealed that most studies have handled \gls{asc} and \gls{ote} extraction tasks using traditional machine learning or deep learning models \cite{Bensoltane2023,Alyami2022,Obiedat2021}. In the case of traditional \gls{ml}, \gls{svm} is the most commonly used approach for \gls{asc}, while \gls{crf} has been the preferred method for \gls{ote} extraction. In this section, however, we will only review deep learning models. Please refer to our systematic review for further information \cite{Alyami2022}. 

\subsubsection{Opinion Target Expression (OTE) Extraction in Arabic}

Researchers employed three different \gls{dl} neural network architectures to extract \gls{ote} from Arabic text: \gls{rnn}, \gls{cnn}, and BERT. For example, \cite{ALSMADI2018_S23} compared \gls{svm} and vanilla \gls{rnn} in terms of performance and training time. They found that \gls{svm}'s performance outperformed that of the vanilla \gls{rnn}, but it required significantly more training time (8.81 hours vs. 1.67 hours). Other researchers, such as \cite{El_Kilany_2017_S21} and \cite{Al-Smadi2019c_S30}, utilized \gls{bi-lstm} coupled with \gls{crf} to address the vanishing gradient problem associated with vanilla \gls{rnn} and to capture contextual features within sentences for output labels \cite{info12110442}. Additionally, researchers like those in \cite{Al-Dabet2020a_S41} and \cite{Abdelgwad2021_S47} integrated \gls{rnn}-based models and \gls{cnn}, with \gls{cnn} being employed to generate character-level representation. \cite{Abdelgwad2021_S47} used an encoder-decoder with attention model using \gls{bi-lstm} attention, unidirectional \gls{lstm} and a \gls{crf} classifier, while \cite{Abdelgwad2021_S47} also utilized \gls{bi-gru} with a \gls{crf} classifier. 

Transformer-based pre-trained models were used only in the works of \cite{Bensoltane2022_S50} and \cite{Fadel2022}.  In \cite{Bensoltane2022_S50}, the authors employed AraBERT alongside \gls{bi-gru} and \gls{crf} to enhance model performance. Meanwhile, \cite{Fadel2022} integrated Arabic Flair string embeddings with AraBERT embeddings, which were then processed through \gls{bi-lstm} layers followed by \gls{crf}, creating a robust architecture for their tasks.

\subsubsection{Aspect Sentiment Classification (ASC) in Arabic}

For \gls{asc}, researchers used four different neural network architectures: \gls{cnn}, \gls{rnn}, \gls{bert}, and memory networks.

The \gls{cnn} neural network was tested by \cite{Masmoudi_2021_S44} and \cite{ruder-etal-2016-insight-1_S12}. \cite{Masmoudi_2021_S44} compared the performance of \gls{cnn}, \gls{lstm}, and \gls{bi-lstm}, with \gls{cnn} achieving the highest performance. Meanwhile, \cite{ruder-etal-2016-insight-1_S12} utilized a \gls{cnn}-based model, which represents the aspect information by concatenating the aspect category embedding with sentence term word embeddings.  

\cite{ALSMADI2018_S23}, \cite{ruder-etal-2016-hierarchical_S14}, and \cite{Liu_S42} employed different variations of \gls{rnn}-based models. For example, \cite{ALSMADI2018_S23} compared the performance of \gls{svm} and vanilla \gls{rnn}. The study found that while \gls{svm} outperformed \gls{rnn} in terms of accuracy, the vanilla \gls{rnn} required less training time. \cite{ruder-etal-2016-hierarchical_S14} utilized a hierarchical \gls{bi-lstm} model to capture relations within review sentences. The researchers achieved this by stacking \gls{bi-lstm} layers, with the first layer generating a representation of all review sentences and the second layer concatenating the aspect category embedding with the sentence-level \gls{bi-lstm} final hidden states to generate a review-level representation. \cite{Liu_S42} utilized multi-level \gls{bi-lstm} layers and \gls{rcnn}. The first-level \gls{bi-lstm} incorporated aspect and topic embeddings along with sentence word embeddings to create a representation of the sentence. The second-level \gls{bi-lstm} combined the sentence representation from the first-level \gls{bi-lstm} with an \gls{rcnn} feature map to capture the long-term dependencies between different aspects of the entire review.  

Several studies have employed attention mechanisms for \gls{asc} to identify the key terms that contribute to the sentiment of a specific aspect. These studies utilized \gls{rnn} with an attention mechanism where the aspect information is included through aspect embedding in the input layer or within the attention mechanism to capture correlations between the aspect and other input words. To create the aspect embedding, the researchers used \gls{ote} or the aspect category. For aspect embedding using \gls{ote}, they excluded any samples that contained implicit \gls{ote} with a null value within the dataset. \cite{Wang_Lu_2018_S32} integrated segment attention and \gls{bi-lstm} to capture essential words for sentiment prediction. Unlike soft attention, which computes the attention score for each term individually, segment attention uses a linear chain \gls{crf} to consider the structural associations between consecutive text fragments. \cite{Al-Smadi2019c_S30} utilized \gls{bi-lstm} followed by an attention mechanism, incorporating the aspect embedding with input word embeddings and the attention mechanism. \cite{Abdelgwad2021_S47} employed an \gls{ian} with \gls{bi-gru} to create aspect and context representations that can learn from each other. \cite{Almani2019_S35} experimented with various attention techniques combined with a \gls{gru} layer, including soft attention, self-attention, and interactive attention layers, and found that self-attention yielded the best performance. Inspired by \cite{chen-etal-2017-recurrent}, \cite{Al-Dabet_2021_S48} utilized a recurrent memory network with position weighting and an attention mechanism to handle \gls{asc}. One distinction between their work and that of \cite{chen-etal-2017-recurrent} is that they employed a bi-directional \gls{indylstm} for the memory module rather than \gls{bi-lstm}.  

A transformer-based pre-trained model such as \gls{bert} was used by \cite{Abdelgwad2022} and \cite{Chennafi2022}. \cite{Abdelgwad2022} used the AraBERT \citep{antoun-etal-2020-arabert} model and added a task-specific linear layer that takes the sentence and explicit \gls{ote}s as input. In their research, \cite{Chennafi2022} utilized an encoder-decoder sequence-to-sequence model to convert Egyptian dialect reviews to \gls{msa}. The converted text was then passed along with the opinion target term into Arabic BERT \cite{safaya-etal-2020-kuisail-arabicbert}, followed by a linear classification layer.

\subsection{English Aspect-Based Sentiment Analysis}

For English \gls{absa}, researchers follow two strategies for employing BERT-based models. The first strategy focuses on enhancing the model architecture by utilizing BERT as an embedding layer and adding various layers to improve the model further. The second strategy includes improving the BERT model by continuing the pre-training of the BERT model using in-domain data (domain-adaptive pre-training).  

\subsubsection{Aspect Sentiment Classification in English}

\cite{Song2019c} enhanced the \gls{bert}-base model by adding \gls{intra-mha} and \gls{inter-mha} to capture within-sentence context representation and aspect representation in relation to its context. These two attentions are followed by \gls{pct} to model local and positional patterns. \cite{sun-etal-2019-utilizing} treated \gls{asc} as \gls{qa} or \gls{nli} by developing four methods for generating auxiliary sentences in the form of \gls{qa} and \gls{nli} pseudo sentences. \cite{Zhaoa2019} employed \gls{gcn} with a sentiment-dependent graph to capture the sentiment dependency between different aspects. They utilized \gls{bert} with position weighting to generate contextual representations of aspect terms, which were then used to construct a sentiment graph. In this graph, the nodes represent aspects, while the edges denote the sentiment dependencies between aspect pairs. \cite{Zeng_lcf_first} utilized two parallel components composed of \gls{bert} and \gls{mhsa} to extract both global and local context features. The local context focus layer uses \gls{cdm} and \gls{cdw} layers, which mask or reduce the weights of non-relative context words using \gls{srd}. \cite{ABSALSA} combined \gls{bert} with Weighted \gls{lsa} to extract sentiment coherence features for different aspects. The \gls{lsa} was built based on the idea that adjacent aspects in text often share similar sentiments. They utilized techniques like \gls{bert}-\gls{spc} input \cite{sun-etal-2019-utilizing}, attention, and syntactical-based local context \cite{YANG2021344,phan-ogunbona-2020-modelling} for aspect-context modeling, and applied a sentiment aggregation window to integrate sentiment information from adjacent aspects to improve aspect sentiment analysis. \cite{Rietzler2020} experimented with domain adaptation using laptop and restaurant datasets, exploring in-domain training, cross-domain training, cross-domain adaptation, and joint-domain training scenarios. They found that joint training on both domains and increasing the number of sentences in the domain adaptation corpus improved model performance.

\subsubsection{Joint Aspect Sentiment Classification (ASC) and Opinion Target Expression (OTE) in English}

Several \gls{absa} studies have investigated various domain adaptation techniques using BERT on English texts. These studies often employed a single model to manage both \gls{asc} and \gls{ote} tasks, or implemented joint learning models to address both tasks simultaneously.

Xu et al. \cite{Xu2019} combined domain and task adaptation by jointly training \gls{bert} on an unlabeled in-domain corpus and on the SQuAD dataset. Although their model outperformed the baseline \gls{bert} across several tasks, it relied on large datasets and did not address limited-domain data scenarios. To tackle this, \cite{Xu2020c} introduced DomBERT, which used domain embeddings and a sampling technique to retrieve data from both target in-domain data and relevant source corpora. DomBERT outperformed the \cite{Xu2019} model in \gls{ote} extraction and end-to-end tasks. However, it obtained lower \gls{asc} accuracy, due to the larger data volume used in \cite{Xu2019}, where sentiment information is more consistent across domains than aspect terms. The authors in \cite{YANG2021344} enhanced a BERT-based model with domain adaptation and an improved version of the local context focus (LCF) technique introduced in \cite{Zeng_lcf_first}. They improved LCF by fusing the outputs of two layers: the memory-dependent window (CDM) and the context-dependent window (CDW). Notably, these models demonstrated accuracy and F1-score improvements over the \gls{bert} baseline across the laptop and restaurant SemEval-2014 and SemEval-2016 datasets, with gains ranging from 0.79\% \cite{YANG2021344} to 6.85\% \cite{Xu2019} for \gls{asc} and 0.92\% \cite{YANG2021344} to 5.81\% \cite{Xu2020c} for \gls{ote}.

Previous work on Arabic \gls{absa} has shown the effectiveness of the BERT-base model over those using Word2Vec or hand-crafted features with RNN or traditional ML algorithms. However, these models, such as AraBERT or Arabic BERT, are often pre-trained on fact-based data like Wikipedia, which may limit their ability to capture the significance of domain-specific terms. For example, aspects like \enquote{service} in restaurant reviews or \enquote{battery life} in electronics hold different significance. Therefore, English \gls{absa} researchers use adaptive pre-training by continuing \gls{bert} pre-training on a corpus with the same domain as the end task \citep{Xu2019, Rietzler2020, gururangan-etal-2020-dont}. However, based on our knowledge, no prior research has been conducted on Arabic \gls{absa} to evaluate domain adaptation on \gls{bert}-base models. 

It is worth noting that many Arabic \gls{dl} models rely on explicit \gls{ote} for \gls{bert}-based models, but exclude implicit \gls{ote} samples, which account for 8.5\% of the data. Additionally, studies using \gls{bert}-base models have only explored full fine-tuning, which is computationally expensive and can lead to catastrophic forgetting. Feature extraction by freezing all layers except the classifier can overcome this issue, but may decrease performance \citep{Abdelgwad2022, Wang_Lu_2018_S32, Al-Smadi2019c_S30, Abdelgwad2021_S47, Al-Dabet_2021_S48}.

This study addresses these limitations by developing various domain-adapted \gls{bert}-based models for Arabic \gls{absa} and experimenting with multiple fine-tuning techniques, including adapters. The adapters are designed to balance the disadvantages and advantages of full fine-tuning and feature extraction. Additionally, to handle both implicit and explicit aspects of the \gls{asc} task, we utilized aspect-category \gls{ea} to input aspect information during the fine-tuning of the \gls{bert} model.

    %! Author = Salha
%! Date = 5/03/2022

\section{Datasets}
\label{sec:datasets}
We employed two types of datasets: unlabeled datasets for domain- and task-adaptive pre-training, and labeled datasets for fine-tuning and evaluation. The end tasks included aspect-based sentiment classification (\gls{asc}) and opinion target expression (\gls{ote}) extraction.

\subsection{Domain Adaptation Corpora}
For adaptive pre-training, we leveraged reviews from the unbalanced version of the \gls{hard} dataset\footnote{\url{https://github.com/elnagara/HARD-Arabic-Dataset}}~\citep{Elnagar2018HotelAD} and the \gls{labr} dataset\footnote{\url{https://github.com/mohamedadaly/LABR}}~\citep{aly-atiya-2013-labr}, both for in-domain (\gls{indapt}) and out-of-domain (\gls{outdapt}) training. Additionally, we used the training portion of the \gls{semeval}-2016 hotel review dataset for task-adaptive pre-training (\gls{tapt}).  

Table~\ref{tab:DA_Data_Before} presents the statistics of the domain adaptation corpora before processing, including counts of unique samples and their sentiment distributions (positive, negative, and neutral).  

\begin{table}[htp!]
\centering
\caption{Overview of domain adaptation datasets before preprocessing.}
\label{tab:DA_Data_Before}
\begin{adjustbox}{max width=\columnwidth}
\begin{tabular}{|c|c|c|c|c|}
\hline
\textbf{Dataset Statistics} & \textbf{LABR} & \textbf{HARD} & \textbf{SemEval Train} & \textbf{SemEval Validation} \\ \hline
Original Reviews & 63,256 & 409,562 & 6,580 & 2,063 \\ \hline
No Duplication & 60,089 & 400,101 & 4,248 & 1,225 \\ \hline
Positive Samples & 40,528 & 268,409 & 2,399 & 660 \\ \hline
Negative Samples & 7,881 & 79,521 & 1,577 & 495 \\ \hline
Neutral Samples & 11,680 & 52,171 & 272 & 70 \\ \hline
\end{tabular}
\end{adjustbox}
\end{table}

\subsection{End-Task Corpora}
We evaluated our proposed model using the \gls{semeval}-2016 hotel review dataset \citep{Al-Smadi_2016_S7} for both \gls{ote} extraction and \gls{asc}. This dataset was originally created for the \gls{semeval} 2016 competition on Arabic text processing. It was derived from \citep{ElSahar2015} and adapted for various \gls{absa} tasks based on the guidelines of \gls{semeval}-2016 Task 5. In this work, we specifically utilized the sentence-level dataset.  

Table~\ref{tab:TaskData} summarizes the Arabic \gls{semeval}-2016 hotel review dataset, detailing training, validation, and test splits for both \gls{asc} and \gls{ote}. For \gls{asc}, sentiment distributions (positive, negative, neutral) are reported. For \gls{ote}, label counts are given for O (Outside), B-\gls{ote} (Beginning), and I-\gls{ote} (Inside). Table~\ref{tab:Asc_Pol_dist} further details sentiment distributions across aspect categories.  

\begin{table}[htp!]
\centering
\caption{Summary of the Arabic SemEval-2016 Hotel Review Dataset.}
\label{tab:TaskData}
\begin{adjustbox}{max width=\columnwidth}
\begin{tabular}{|c|c|c|c|}
\hline
\textbf{Task} & \textbf{Training Count} & \textbf{Validation Count} & \textbf{Testing Count} \\ \hline
\multicolumn{4}{|c|}{\textbf{Aspect Sentiment Classification (ASC)}} \\ \hline
\textbf{Total Examples} & 8461 & 2116 & 2604 \\ \hline
Positive Examples & 5002 & 1251 & 1508 \\ \hline
Negative Examples & 2911 & 728 & 927 \\ \hline
Neutral Examples & 548 & 137 & 169 \\ \hline
\multicolumn{4}{|c|}{\textbf{Opinion Target Expression (OTE) Extraction}} \\ \hline
\textbf{Total Examples} & 3841 & 961 & 1227 \\ \hline
O (Outside) Labels & 77041 & 19135 & 23969 \\ \hline
B-OTE (Beginning) Labels & 7613 & 2007 & 2372 \\ \hline
I-OTE (Inside) Labels & 1481 & 377 & 499 \\ \hline
\end{tabular}
\end{adjustbox}
\end{table}

\section{Methodology}
\label{sec:methodology}
This study evaluates the impact of domain-adaptive pre-training (\gls{dapt}) on the performance of Arabic pre-trained language models for \gls{absa} tasks. We fine-tuned seven Arabic \gls{bert}-based models for \gls{asc} and \gls{ote}, selecting the top three for further domain- and task-adaptive pre-training. The models were then evaluated against baselines without adaptation. We also explored different fine-tuning strategies, including full fine-tuning, adapter-based approaches, and methods to address dataset imbalance.  

\begin{figure*}[htbp]
    \centering
    \includegraphics[width=1\textwidth]{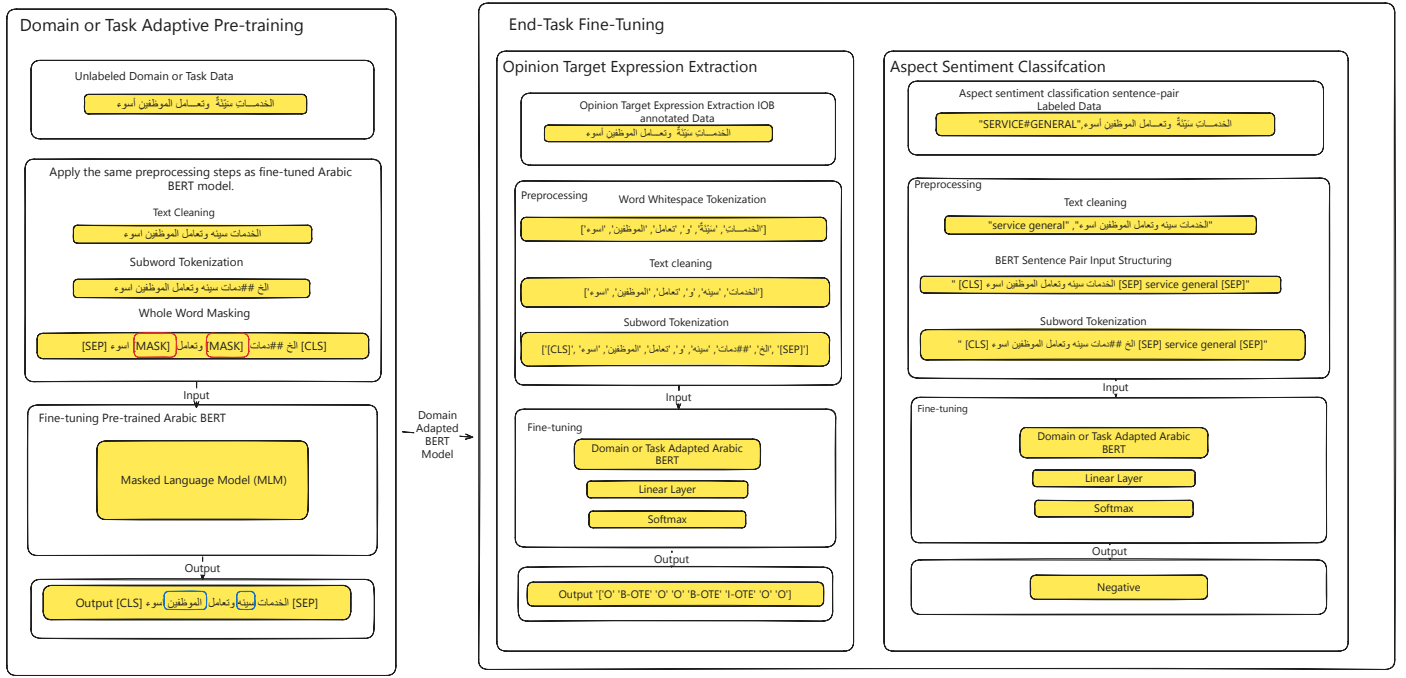}
    \caption{Overview of Domain and Task-Adaptive Pre-training with Fine-Tuning for End-Tasks}
    \label{fig:ModelOverView}
\end{figure*}

Figure~\ref{fig:ModelOverView} illustrates the overall workflow, including domain/task-adaptive pre-training and fine-tuning for end tasks.  

\subsection{Task Definitions}
We addressed two end tasks: (i) \gls{ote} extraction, framed as sequence labeling in IOB format (B = beginning of \gls{ote}, I = inside, O = outside), and (ii) \gls{asc}, framed as sentence-level classification of aspect categories into positive, neutral, or negative polarities.  

\subsection{Arabic BERT Models}
We fine-tuned seven Arabic \gls{bert}-based models pre-trained on diverse corpora:  
\begin{itemize}
    \item \textbf{MSA-based}: AraBERTv02~\citep{antoun-etal-2020-arabert}, Arabic-BERT~\citep{safaya-etal-2020-kuisail-arabicbert}, and CAMeLBERT-\gls{msa}~\citep{inoue-etal-2021-CAMELBERT}.  
    \item \textbf{Dialect-based}: QARiB~\citep{abdelali2021pretrainingQarib} and CAMeLBERT-\gls{da}.  
    \item \textbf{Mixed MSA and dialects}: CAMeLBERT-Mix and MARBERTv2~\citep{abdul-mageed-etal-2021-marbert}.  
\end{itemize}

\subsection{Domain and Task Adaptation}
Following prior work~\citep{Xu2019,gururangan-etal-2020-dont}, we employed masked language modeling (\gls{mlm}) as the objective for domain and task adaptation, given its superiority over next-sentence prediction (\gls{nsp}). For \gls{dapt}, we continued pre-training models on in-domain and out-of-domain unlabeled corpora. For \gls{tapt}, we used unlabeled samples from the task training set. Additionally, we experimented with sequential pre-training (\gls{dapt}+\gls{tapt}).  

\subsection{Fine-Tuning Strategies}
We tested three fine-tuning strategies:  
\begin{enumerate}
    \item \textbf{Feature extractor}: freezing all transformer layers except the classification head.  
    \item \textbf{Full fine-tuning}: updating all model parameters.  
    \item \textbf{Adapter-based fine-tuning}: updating only task-specific adapter layers.  
\end{enumerate}

\subsection{Adapters and Adapter Fusion}
Adapters strike a balance between efficiency and flexibility, enabling domain/task specialization without modifying the entire pre-trained model. We adopted the MAD-X architecture~\citep{pfeiffer-etal-2020-mad}, employing Pfeiffer invertible adapters for domain adaptation and comparing three adapter types for task fine-tuning: Pfeiffer, Houlsby~\citep{pmlr-v97-houlsby19a}, and parallel bottleneck adapters~\citep{he2022towards}.  

Given the correlation between \gls{ote} and \gls{asc}~\citep{Ruidan2019,YANG2021344}, we further applied adapter fusion~\citep{pfeiffer-etal-2021-adapterfusion} to share knowledge across tasks. Fusion parameters were trained while keeping individual adapters and base model weights frozen, enabling efficient multi-task transfer.

    %! Author = Salha
%! Date = 5/03/2021

\section{Experiments and Results}
\label{sec:results}

\subsection{Data Preparation}
\label{subsec:data-prep}

\subsubsection{General Preprocessing}
To ensure the quality of the datasets used in our experiments, we performed a sequence of common preprocessing steps that are frequently used on Arabic pre-trained models, including CAMeLBERT \citep{inoue-etal-2021-CAMELBERT} and AraBERT \citep{antoun-etal-2020-arabert}. These techniques include removing extra white spaces, diacritics, control characters, and elongation (kashida). We also replaced URLs and user mentions with ``URL'' and ``USER'' strings, respectively. These preprocessing steps were applied to both the domain adaptation datasets and the end-task dataset.

\subsubsection{Domain-Adaptation Data Preprocessing}
\begin{figure}[htp!]
    \centering
    \includegraphics[width=\columnwidth]{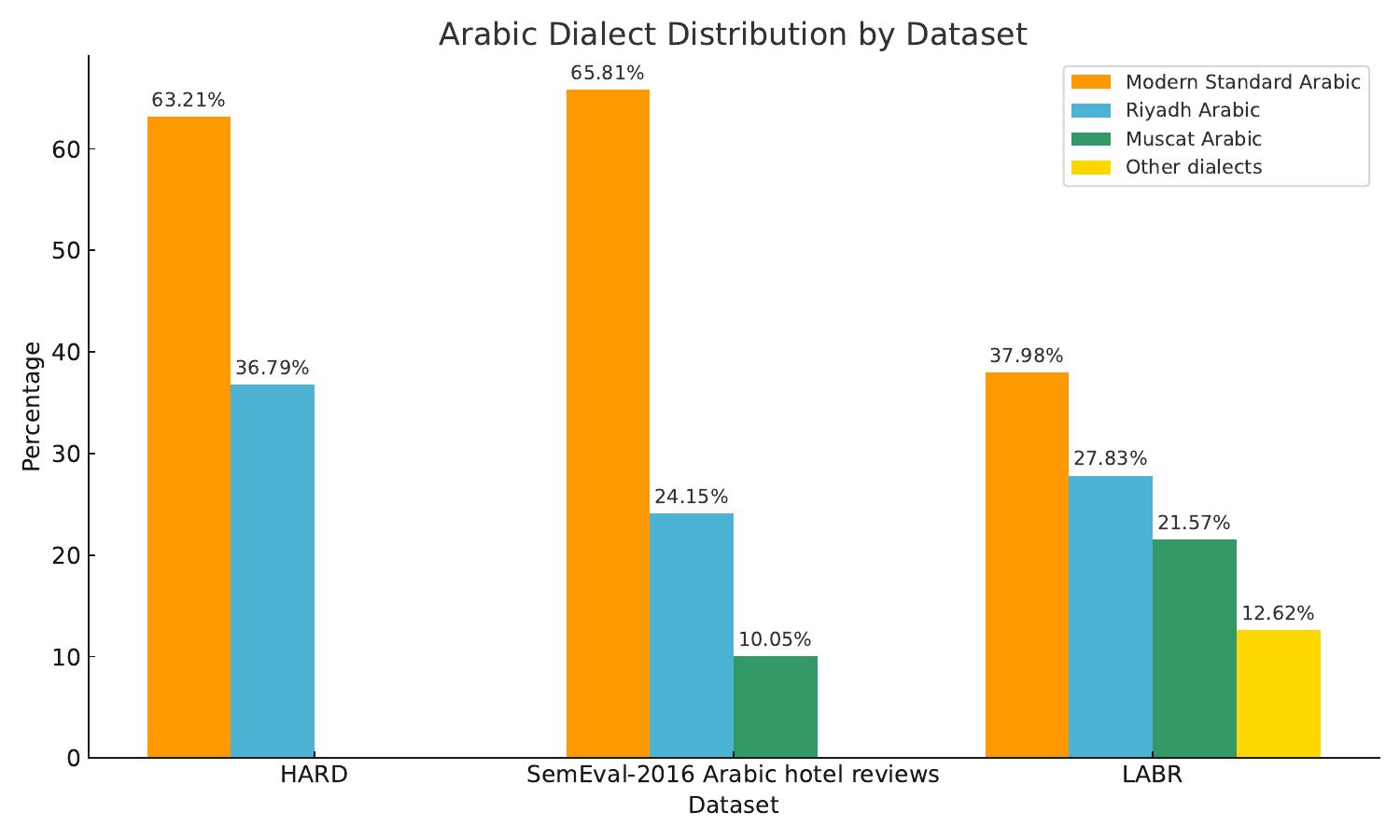}
    \caption{Arabic Dialect Distribution by Dataset: The chart shows the distribution of various Arabic dialects across \gls{hard}, \gls{semeval}-2016 hotel reviews, and LABR datasets. Data extracted with Camel Tools.}
    \label{fig:DialectsCount}
\end{figure}

\begin{figure}[htp!]
    \centering
    \begin{subfigure}[b]{\columnwidth}
        \centering
        \includegraphics[width=\columnwidth]{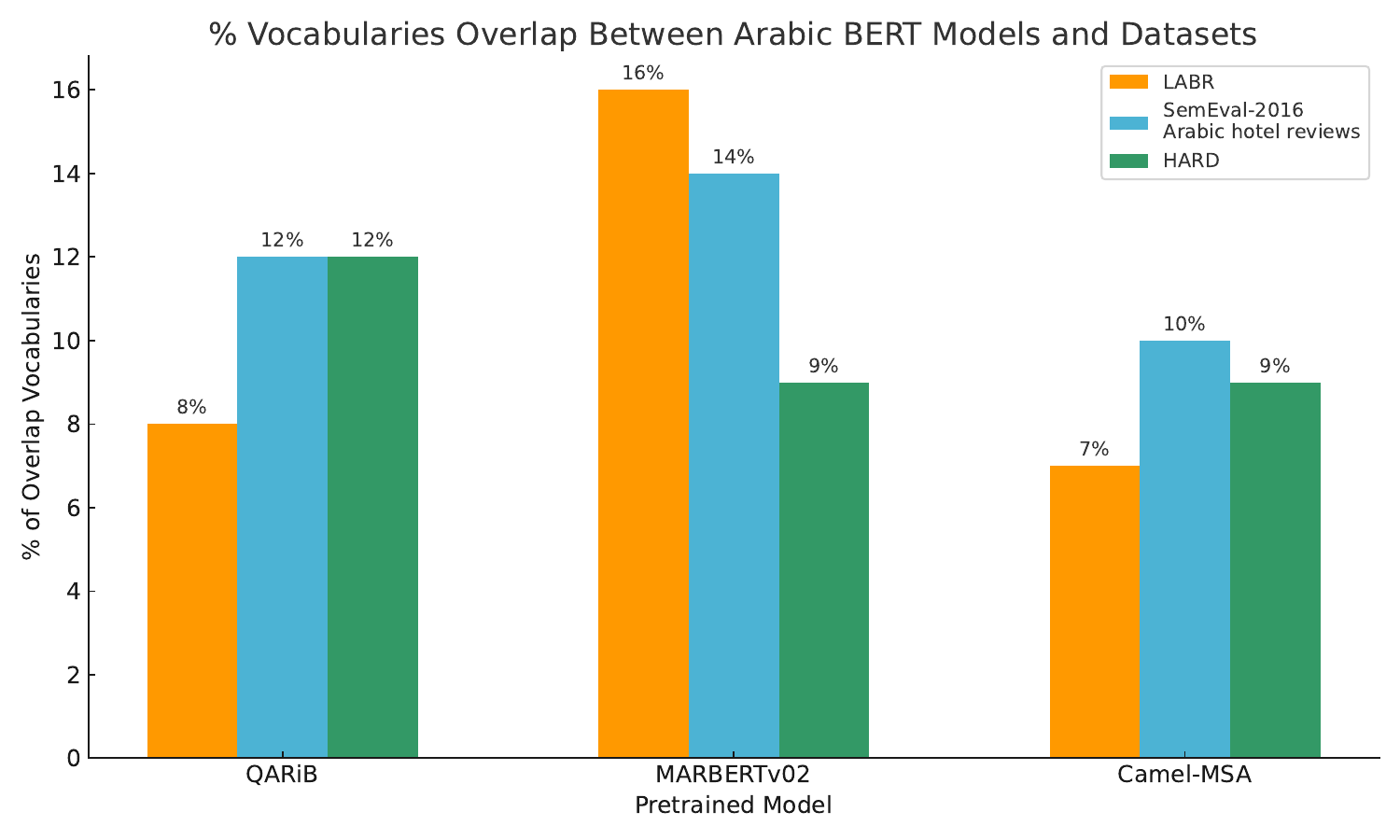}
        \caption{Percentage of vocabulary overlap between the domain adaptation datasets and pre-trained models.}
        \label{fig:bar_chart_overlapj}
    \end{subfigure}
    \vspace{1em}
    \begin{subfigure}[b]{\columnwidth}
        \centering
        \includegraphics[width=\columnwidth]{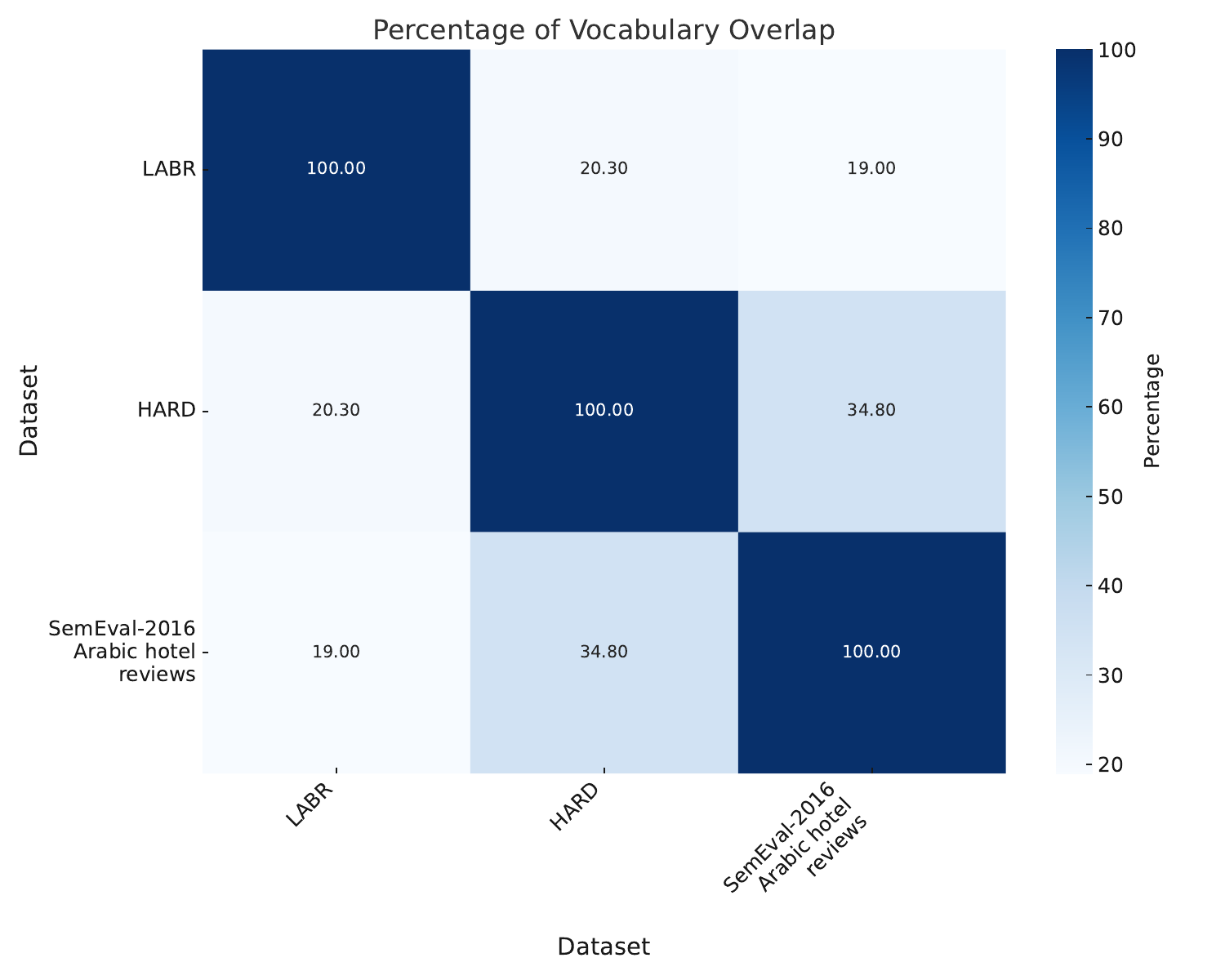}
        \caption{Percentage of vocabulary overlap between different datasets.}
        \label{fig:heatmap_overlapj}
    \end{subfigure}
    \caption{Percentages of vocabulary overlap. (a) Overlap between domain adaptation datasets and pre-trained models. (b) Overlap between domain adaptation datasets and \gls{semeval} data. The overlap is calculated using the heuristic from \citet{gururangan-etal-2020-dont}, considering the most frequent 10,000 words, excluding stop words.}
    \label{fig:OverlapDSMODEL}
\end{figure}

After selecting the top three Arabic \gls{bert} models with the best performance in the two tasks, we implemented the specific preprocessing steps mentioned in the model's research paper and GitHub repository for each selected model. The statistics and preprocessing steps for each dataset are provided in Tables \ref{tab:DA_Data_After} and \ref{tab:DAPrep}. Additionally, Figures \ref{fig:DialectsCount} and \ref{fig:OverlapDSMODEL} illustrate the dialect distributions and vocabulary overlap between the datasets, respectively.

\begin{table}[htp!]
\centering
\caption{Overview of domain adaptation datasets after preprocessing.}
\label{tab:DA_Data_After}
\begin{adjustbox}{max width=\columnwidth}
\begin{tabular}{|c|c|c|c|c|}
\hline
\textbf{Preprocessing Method} & \textbf{LABR} & \textbf{HARD} & \textbf{SemEval Train} & \textbf{SemEval Validation} \\ \hline
Min Preprocessing & 60,089 & 400,101 & 4,248 & 1,225 \\ \hline
QARiB Preprocessing & 60,089 & 400,101 & 4,248 & 1,225 \\ \hline
MARBERTV2 Preprocessing & 60,089 & 400,101 & 4,248 & 1,225 \\ \hline
CAMeLBERT-MSA Preprocessing & 60,084 & 399,839 & 4,248 & 1,225 \\ \hline
\end{tabular}
\end{adjustbox}
\end{table}

\begin{table}[ht!]
\centering
\caption{Preprocessing techniques for Arabic BERT models used on domain adaptation datasets.}
\label{tab:DAPrep}
\begin{adjustbox}{max width=\columnwidth}
\begin{tabular}{|c|p{0.75\textwidth}|}
\hline
\textbf{Model Name} & \textbf{Preprocessing Steps} \\ \hline
CAMeLBERT-MSA & Unicode normalization, removal of non-Arabic characters, extra spaces, control characters, punctuation, and kashida. \\ \hline
QARiB & Convert English letters to lowercase, remove diacritics, extra whitespace, and non-UTF-8 characters. Replace URLs, numbers, and mentions with 'URL', 'NUM', and 'USERNAME'. Reduce repetition of Arabic letters to a maximum of two instances. \\ \hline
MARBERTv2 & Remove extra spaces and control characters. Replace hashtags, URLs, and user mentions with 'HASHTAG', 'URL', and 'USER'. \\ \hline
\end{tabular}
\end{adjustbox}
\end{table}

\subsubsection{End-Task Dataset Preprocessing}
In addition to the preprocessing steps mentioned earlier at the beginning of Section \ref{subsec:data-prep}, we converted the \gls{semeval} dataset from XML to CSV format using scripts provided by\footnote{\url{https://github.com/howardhsu/ABSA_preprocessing}}~\citep{Xu2019}. For the extraction of \gls{ote}, we labeled each word as ``O'' if it was not part of \gls{ote}, ``B-\gls{ote}'' if it was the beginning of the \gls{ote}, and ``I-\gls{ote}'' for any word that was part of the \gls{ote} after the start. In terms of aspect categories, we replaced the hashtag (\#) sign with a space between entity and attribute pairs and then normalized them to lowercase. For example, ``ROOMS\#GENERAL'' became ``rooms general.''

\subsection{Addressing Class Imbalance}
\label{subsec:imbalance-method}
Table~\ref{tab:TaskData} reveals significant class imbalances, particularly in \gls{ote} extraction. Using the imbalance ratio (IR) formula~\ref{eq:IR}, the ratios between the majority class ``O'' and the minority classes ``B-\gls{ote}'' and ``I-\gls{ote}'' are 10.11 and 52.03, respectively. Similarly, the IR for aspect sentiment polarities between the majority (positive) and minority (neutral) classes is 9.13. Such imbalances can adversely affect model performance \citep{imbalnceDSANDHandleit}.

To mitigate this, we employed focal loss with class weighting. Focal loss emphasizes misclassified or difficult examples \citep{classImbalanceFocalLossPvsR}, while class weights adjust the loss based on class frequencies, giving more importance to minority classes.

We computed class weights using three formulas:
\begin{enumerate}
    \item \textbf{Imbalance Ratio (IR):}
    \begin{equation}
    \text{IR} = \frac{N_{\text{maj}}}{N_{\text{min}}} \label{eq:IR}
    \end{equation}
    \item \textbf{Inverse Frequency Class Weighting:}
    \begin{equation}
    w_j = \frac{N}{k \times n_j}
    \end{equation}
    \item \textbf{Inverse Frequency Class Weighting with Constant:}
    \begin{equation}
    w_j' = \frac{w_j + C}{\sum_{i=1}^{k} (w_i + C)}
    \end{equation}
\end{enumerate}
Where: \( w_j \) and \( w_j' \) are the weights for class \( j \); \( N \) is the total number of samples; \( k \) is the total number of classes; \( n_j \) is the number of samples in class \( j \); and \( C \) is a constant added in the last method.

\subsection{Hyper-parameter Selection}
\label{subsec:HP}
Initially, we utilized Optuna~\citep{Akiba2019-Optuna}\footnote{\url{https://optuna.org/}} for hyper-parameter selection, performing 500 trials per model using feature extraction and full fine-tuning techniques. However, due to computational constraints, each experiment took 2--4 days to complete, and some of the hyperparameters selected by Optuna led to suboptimal performance. Consequently, we decided to use the default hyperparameters recommended for \gls{mbert}~\citep{devlin-etal-2019-bert}, as well as those specified for sequence and token classification tasks in Arabic \gls{bert} pre-trained models. For instance, we employed hyperparameters designed for sentence-level sentiment analysis for the \gls{asc} task, and those selected for named-entity recognition for the \gls{ote} extraction task.

Tables~\ref{tab:hyperparameters} and \ref{tab:OptunaHP} show the details of the hyperparameter configuration for the Arabic \gls{bert} models evaluated for both the \gls{asc} and \gls{ote} extraction tasks. Specifically, Table~\ref{tab:hyperparameters} includes details on whether the model parameters are fine-tuned using full fine-tuning or feature extractions, as well as the learning rates, batch sizes, and number of epochs used during training. Table~\ref{tab:OptunaHP} presents the hyper-parameters search space optimized via Optuna, including the learning rate, batch size, and number of epochs.

\begin{table}[htp!]
\centering
\caption{Hyperparameter Configuration for Tested Arabic BERT Models. Abbreviations: ASC = Aspect Sentiment Classification, OTE = Opinion Target Expression, Full FT = Full Fine-tuning, LR = Learning Rate, Batch = Batch Size, MSA = Modern Standard Arabic, DA = Dialectal Arabic, Mix = Mixed.}

\label{tab:hyperparameters}
\scriptsize
\begin{tabular}{|c|c|c|c|c|c|}
\hline
\textbf{Model Name} & \textbf{Task} & \textbf{Full FT} & \textbf{LR} & \textbf{Batch} & \textbf{Epochs} \\ \hline
\multirow{3}{*}{All Models} & ASC, OTE & Yes & $5 \times 10^{-5}$ & 32 & 3, 5 \\ \cline{2-6}
 & ASC, OTE & No & 0.002 & 32 & 5, 10 \\ \cline{2-6}
 & ASC, OTE & No & 0.0094 & 32 & 5, 10 \\ \hline
QARiB & ASC, OTE & Yes & $8 \times 10^{-5}$ & 64 & 3 \\ \hline
MARBERTv2 & ASC, OTE & Yes & $2 \times 10^{-6}$ & 32 & 25 \\ \hline
Asafaya-Arabic BERT & ASC, OTE & Yes & $2 \times 10^{-5}$ & 32 & 10 \\ \hline
\multirow{2}{*}{\begin{tabular}[c]{@{}c@{}}CAMeLBERT-\\MSA, DA, Mix\end{tabular}} & ASC & Yes & $3 \times 10^{-5}$ & 32 & 3, 5 \\ \cline{2-6}
 & OTE & Yes & $5 \times 10^{-5}$ & 32 & 10 \\ \hline
AraBERTv0.2 & ASC, OTE & Yes & $2 \times 10^{-5}$ & 32 & 3, 5 \\ \hline
\end{tabular}
\end{table}

\begin{table}[htbp]
\centering
\caption{Optuna Hyperparameter Search-Space.}
\label{tab:OptunaHP}
\scriptsize
\begin{tabular}{|c|>{\centering\arraybackslash}p{0.6\columnwidth}|}
\hline
\textbf{Hyperparameter} & \textbf{Range} \\ \hline
Learning rate           & $1 \times 10^{-2}$ to $9 \times 10^{-5}$ \\ \hline
Batch size              & 4, 8, 16, 32, 64, 128, 256 \\ \hline
Number of epochs        & 2 to 10 \\ \hline
\end{tabular}
\end{table}

We compared the performance of models with Optuna-optimized hyperparameters to those using the default hyperparameters for \gls{mbert} or task-specific hyperparameters from Arabic \gls{bert} pre-trained models. Most models obtained higher performance with the default or task-specific hyperparameters, leading us to test \gls{ote} extraction models with these settings only. Only three fully fine-tuned models showed marginally higher performance: QARiB by 0.2884\%, Arabic \gls{bert} by 0.1939\%, and CAMeLBERT-Mix by 0.1453\% accuracy (see Table \ref{tab:OptunaASC_arabic_bert_performance}).

\subsection{Domain Adaptation: Setup}
\label{subsec:domain-setup}

\subsubsection{Full Fine-tuning}
We selected the three highest-performing Arabic \gls{bert} models for domain adaptation, including CAMeLBERT-\gls{msa}, QARiB, and MARBERTV02 (see Tables \ref{tab:arabic_bert_ote_full_hyperparameter_performance} and \ref{tab:arabic_bert_aspect_sentiment_classification}). We then continued pre-training these models using \gls{mlm} with a duplication factor of 10, whole-word masking at a 15\% rate, a batch size of 32, and a learning rate of 2.00E-05. We used Optuna to determine the number of epochs, considering more epochs for smaller datasets to balance dataset sizes. Experiments were conducted with in-domain (HARD), out-of-domain (LABR), and task adaptation (SemEval-2016) datasets. Table~\ref{tab:DAEPOCH} lists the number of training epochs for each model across adaptation datasets and tasks.

\begin{table}[htp!]
\centering
\caption{Epoch counts for domain-adapted models by dataset. Each model is listed with task-specific epochs in the format: Model (ASC: Epochs, OTE: Epochs).}
\label{tab:DAEPOCH}
\scriptsize
\begin{tabular}{|p{0.43\columnwidth}|p{0.52\columnwidth}|}
\hline
\textbf{Dataset} & \textbf{Number of Epochs} \\ \hline
\multirow{3}{*}{HARD} & CAMeLBERT-MSA (ASC: 14, OTE: 9) \\ \cline{2-2}
& QARiB (ASC: 10, OTE: 5) \\ \cline{2-2}
& MARBERTv2 (ASC: 6, OTE: 7) \\ \hline
\multirow{3}{*}{LABR} & CAMeLBERT-MSA (ASC: 14, OTE: 8) \\ \cline{2-2}
& QARiB (ASC: 4, OTE: 20) \\ \cline{2-2}
& MARBERTv2 (ASC: 24, OTE: 23) \\ \hline
\multirow{3}{*}{SemEval-2016 Hotel Reviews} & CAMeLBERT-MSA (ASC: 28, OTE: 39) \\ \cline{2-2}
& QARiB (ASC: 26, OTE: 8) \\ \cline{2-2}
& MARBERTv2 (ASC: 40, OTE: 4) \\ \hline
\multirow{3}{*}{HARD+SemEval-2016 Hotel Reviews} & CAMeLBERT-MSA (ASC: 33, OTE: 4) \\ \cline{2-2}
& QARiB (ASC: 12, OTE: 25) \\ \cline{2-2}
& MARBERTv2 (ASC: 7, OTE: 20) \\ \hline
\multirow{3}{*}{LABR+SemEval-2016 Hotel Reviews} & CAMeLBERT-MSA (ASC: 26, OTE: 40) \\ \cline{2-2}
& QARiB (ASC: 7, OTE: 7) \\ \cline{2-2}
& MARBERTv2 (ASC: 2, OTE: 6) \\ \hline
\end{tabular}
\end{table}

\subsubsection{Adapter-based Domain Adaptation}
We selected CAMeLBERT-\gls{msa} as the base model for adapter domain adaptation due to its superior performance in full fine-tuning experiments. We then continued pre-training it using a masked language model with a Pfeiffer invertible language adapter, a learning rate of 5.00E-05, and batch sizes of 64 and 128. To determine the optimal number of epochs, we evaluated accuracy and perplexity. Perplexity in \gls{mlm} measures a model's proficiency in accurately predicting masked tokens within a text sequence. Lower perplexity values indicate better predictive performance and a more robust understanding of the linguistic context. For the HARD dataset, we used 10\% of the data as the validation set, and for SemEval, we used 20\%. Our results indicated that increasing the number of epochs improved accuracy and reduced perplexity. Thus, we used 50 epochs for in-domain adaptation with the \gls{hard} dataset and 400 epochs for the smaller \gls{semeval} dataset. Table~\ref{tab:AdapterLMHP} presents the hyperparameters and performance metrics for language model adapters across datasets, including perplexity.

We fine-tuned the adapters (task or language-task stacked) for 20 epochs with a batch size of 32 and a learning rate of 1e-4. For fused adapter models, we used the same settings except for a learning rate of 5e-5. Following Focal Loss research guidelines, we explored gamma values from 1 to 5 and tested class weights with constants varying from 0.1 to 0.9.

\begin{table}[htp!]
\centering
\caption{Hyperparameters and Performance Metrics for Language Model Adapters across Datasets. The highest validation accuracy is highlighted in \textbf{bold}.}
\label{tab:AdapterLMHP}
\scriptsize
\begin{adjustbox}{width=\columnwidth}
\begin{tabular}{|c|c|c|c|}
\hline
\textbf{Batch Size} & \textbf{Epochs} & \textbf{Validation Accuracy} & \textbf{Perplexity} \\ \hline
\multicolumn{4}{|c|}{\textbf{HARD Dataset}} \\ \hline
64 & 20 & 57.18\% & 9.90 \\ \hline
128 & 50 & \textbf{58.30}\% & 9.11 \\ \hline
\multicolumn{4}{|c|}{\textbf{SemEval-2016 Hotel Reviews}} \\ \hline
64 & 20 & 49.29\% & 15.72 \\ \hline
64 & 100 & 50.70\% & 13.92 \\ \hline
128 & 400 & \textbf{52.53}\% & 13.45 \\ \hline
\end{tabular}
\end{adjustbox}
\end{table}

\subsubsection{Evaluation and Significance Testing}
We computed the average performance across three runs in each experiment and performed a t-test to evaluate statistical significance against the model without domain adaptation. The null hypothesis (H0) states that domain and task adaptive pre-training does not affect performance relative to the baseline (No-\gls{dapt}), while the alternative hypothesis (H1) claims that it either improves or decreases performance.

\subsection{Results: Domain Adaptation with Full Fine-tuning}
\label{subsec:results-full}

We evaluated single and sentence-pair \gls{bert} input formats for \gls{asc} across the three selected models. The sentence-pair format, which includes both the review text and the aspect as separate inputs, demonstrated better performance in two of the three models, leading us to adopt this format for subsequent experiments (see Tables \ref{tab:arabic_bert_aspect_sentiment_classification} and No-\gls{dapt} in \ref{tab:SPASC}).

After analyzing the results of task and domain adaptive pre-training, as presented in Tables \ref{tab:SPASC} and \ref{tab:OTEDA}, we observed that for the \gls{asc} task, the \gls{indapt} models outperformed other domain adaptation settings such as \gls{tapt}, \gls{outdapt}, and No-\gls{dapt} in terms of both accuracy and macro-F1. Regarding the \gls{ote} extraction task, both \gls{indapt}+\gls{tapt} QARiB and MARBERTv2 outperformed other settings, while \gls{indapt} CAMeLBERT-MSA achieved the highest results.

The CAMeLBERT-\gls{msa} model has consistently demonstrated superior performance in the \gls{asc} task compared to MARBERTv2 and QARiB in terms of both accuracy and macro-F1. Specifically, the \gls{indapt} and \gls{indapt}+\gls{tapt} versions of CAMeLBERT-MSA have achieved the highest performance in \gls{asc}, with the \gls{indapt} maximum accuracy at 89.67\% and macro-F1 of 75.16\%, and an average of 88.26\% accuracy and 73.77\% macro-F1. Even the baseline No-\gls{dapt} CAMeLBERT-\gls{msa} model has proven to be more effective than other models in terms of average accuracy and macro-F1. Regarding the \gls{ote} extraction task, \gls{indapt} CAMeLBERT-\gls{msa} has outperformed all QARiB and MARBERTv2 models, exhibiting a micro-F1 score of 77.02\%.

\begin{figure}[htp!]
\centering
\includegraphics[width=\linewidth]{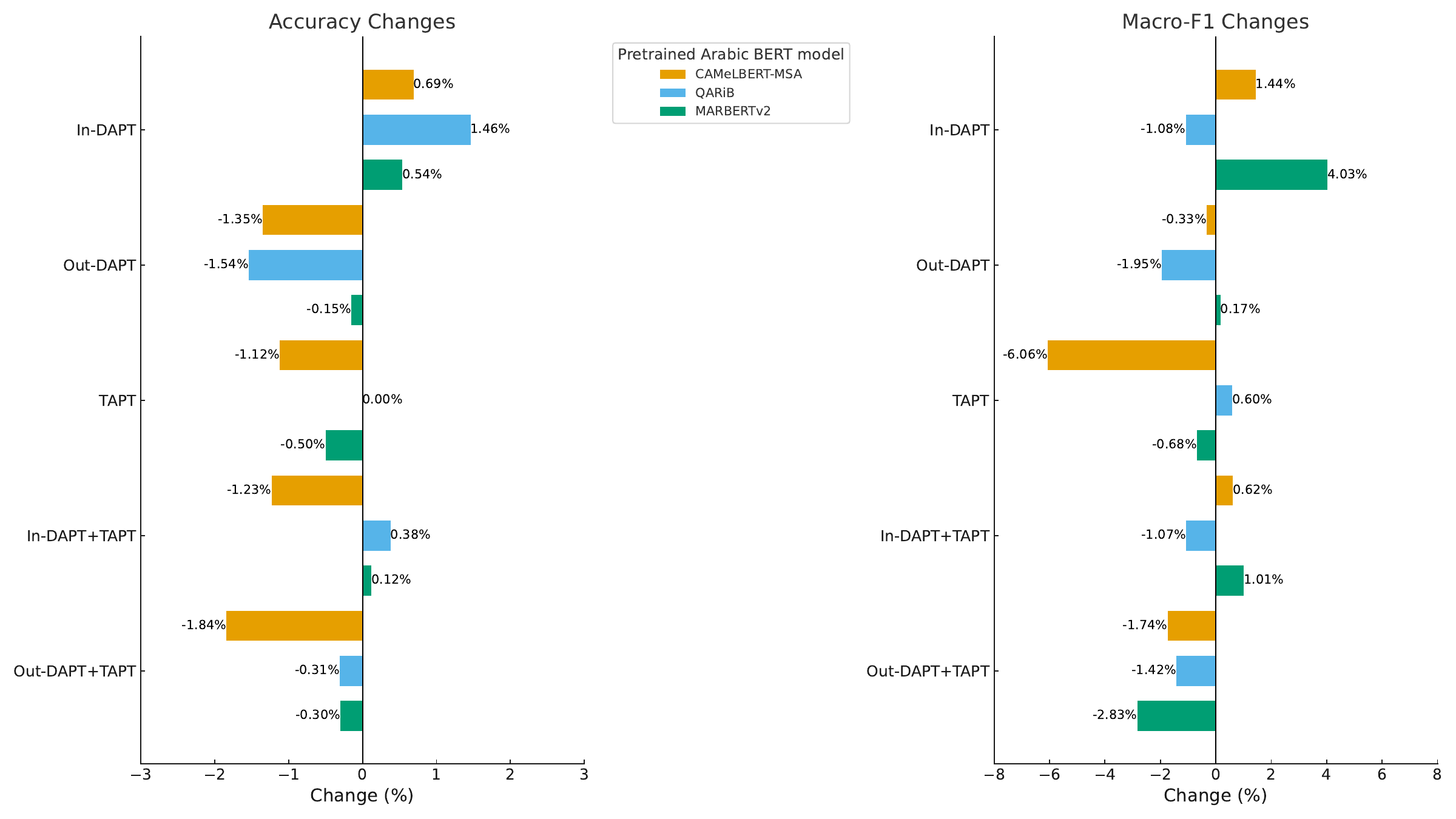}
\caption{Accuracy and Macro-F1 Change for pre-trained Arabic BERT models after domain adaptation for aspect-sentiment classification. The zero line represents the baseline without domain adaptation (No-\gls{dapt}). Bars indicate percentage changes, with positive values representing improvements and negative values representing declines.}
\label{fig:ASCdomain_adaptation_graph}
\end{figure}

\begin{figure}[htp!]
\centering
\includegraphics[width=\linewidth]{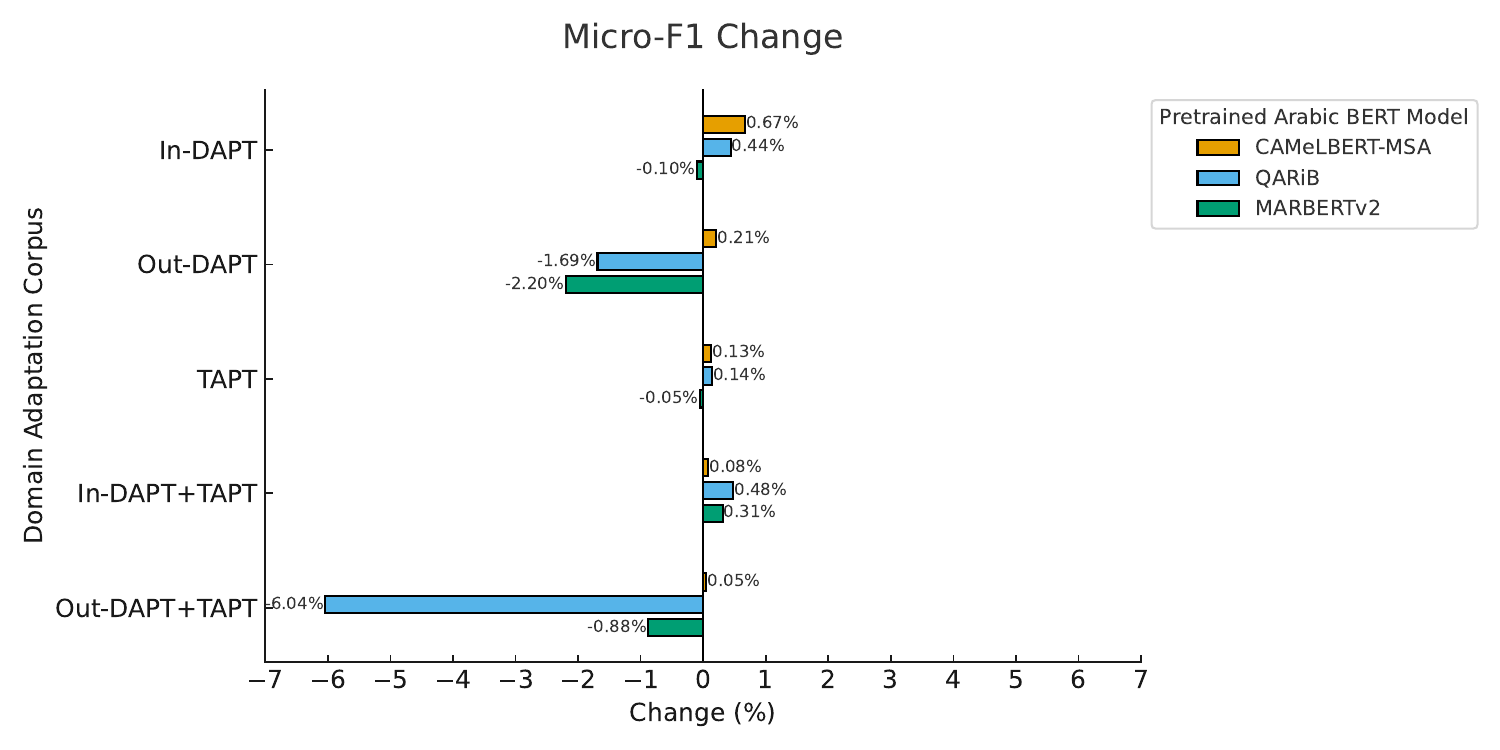}
\caption{Micro-F1 Change for pre-trained Arabic BERT models after domain adaptation for OTE extraction. The zero line represents the baseline without domain adaptation (No-\gls{dapt}). Bars indicate percentage changes, with positive values representing improvements and negative values representing declines.}
\label{fig:OTEdomain_adaptation_graph}
\end{figure}

Figures \ref{fig:ASCdomain_adaptation_graph} and \ref{fig:OTEdomain_adaptation_graph} show percentage differences in max run accuracy, macro-F1, and micro-F1 compared to the No-\gls{dapt} baseline for \gls{asc} and \gls{ote} tasks. For \gls{asc}, CAMeLBERT-MSA and QARiB with \gls{indapt} achieved the highest accuracy gains (0.69\% and 1.46\%, respectively). For macro-F1, only \gls{indapt} CAMeLBERT-MSA and MARBERTv2 showed improvements, while QARiB generally decreased, except for \gls{tapt}. CAMeLBERT-MSA and QARiB showed the greatest micro-F1 improvement on \gls{ote}, with CAMeLBERT-MSA (\gls{indapt}) leading at 0.67\% and QARiB (\gls{indapt}+\gls{tapt}) at 0.48\%. MARBERTv2 had the smallest gains, with 0.54\% in accuracy on \gls{asc} and 0.31\% in micro-F1 on \gls{ote} (\gls{indapt}+\gls{tapt}).

For \gls{ote} extraction, we tested \gls{indapt} CAMeLBERT-\gls{msa} with two popular sequence labelling classifiers: Softmax and \gls{crf}. These classifiers were enhanced by incorporating varying numbers of recurrent neural network layers, including \gls{bi-lstm} and \gls{bi-gru}. As summarized in Table \ref{tab:OTE_classifiers}, the Softmax classifier alone obtained the highest performance in F1-score (maximum 77.02\%, average 76.47\%) and precision (maximum 79.05\%, average 78.46\%). However, models with \gls{bi-lstm}, \gls{bi-gru}, or \gls{crf} classifiers achieved higher recall, with a single-layer \gls{bi-gru} followed by Softmax achieving the highest recall (maximum 79.85\%, average 79.19\%).

\begin{table*}[htp!]
\caption{Domain-Adapted Performance Metrics for Aspect Sentiment Classification with Full Fine-Tuning Across CAMeLBERT-MSA, QARiB, and MARBERTv2. The highest result among all models is \textbf{\underline{bold and underlined}}, while the highest result for each Arabic BERT model is \underline{underlined}. Significant p-values (<0.05) are dashed underlined. Metrics: Acc (Accuracy), M-F1 (Macro F1), M-P (Macro Precision), M-R (Macro Recall).}
\label{tab:SPASC}
\centering
\begin{adjustbox}{width=\textwidth}
\begin{tabular}{|l|c|c|c|c|c|c|c|c|c|c|c|c|c|c|c|c|}
\hline
\multirow{2}{*}{\begin{tabular}[c]{@{}c@{}}Adaptation\\ Technique\end{tabular}} & \multicolumn{4}{c|}{Maximum (\%)} & \multicolumn{4}{c|}{Average (\%)} & \multicolumn{4}{c|}{Variance (\%)} & \multicolumn{4}{c|}{P-value} \\ \cline{2-17}
 & \multicolumn{1}{c|}{Acc} & \multicolumn{1}{c|}{M-F1} & \multicolumn{1}{c|}{M-P} & \multicolumn{1}{c|}{M-R} & \multicolumn{1}{c|}{Acc} & \multicolumn{1}{c|}{M-F1} & \multicolumn{1}{c|}{M-P} & \multicolumn{1}{c|}{M-R} & \multicolumn{1}{c|}{Acc} & \multicolumn{1}{c|}{M-F1} & \multicolumn{1}{c|}{M-P} & \multicolumn{1}{c|}{M-R} & \multicolumn{1}{c|}{Acc} & \multicolumn{1}{c|}{M-F1} & \multicolumn{1}{c|}{M-P} & \multicolumn{1}{c|}{M-R} \\ \hline

\multicolumn{17}{|c|}{\textbf{CAMeLBERT-MSA}} \\ \hline
No-DAPT & 88.98 & 73.72 & 74.02 & 73.50 & 87.92 & 72.37 & 76.05 & 70.98 & 0.0093 & 0.0246 & 0.0590 & 0.0636 & N/A & N/A & N/A & N/A \\ \hline
\begin{tabular}[c]{@{}c@{}}In-DAPT\end{tabular} & \textbf{\underline{89.67}} & \textbf{\underline{75.16}} & \textbf{\underline{78.95}} & 73.33 & \textbf{\underline{88.26}} & \textbf{\underline{73.77}} & \textbf{\underline{75.42}} & 73.03 & 0.0017 & 0.0147 & 0.0966 & 0.0054 & 0.54 & 0.29 & 0.28 & 0.19 \\ \hline
\begin{tabular}[c]{@{}c@{}}Out-DAPT\end{tabular} & 87.63 & 73.39 & 73.76 & 73.09 & 87.15 & 67.19 & 73.57 & 67.09 & 0.0043 & 0.2886 & 0.0754 & 0.2715 & 0.49 & 0.17 & 0.47 & 0.16 \\ \hline
TAPT & 87.86 & 67.66 & 73.00 & 66.21 & 87.29 & 64.28 & 74.56 & 64.24 & 0.0028 & 0.1483 & 0.0272 & 0.0453 & 0.54 & 0.07 & 0.43 & 0.05 \\ \hline
\begin{tabular}[c]{@{}c@{}}In-DAPT+TAPT\end{tabular} & 87.75 & 74.34 & 74.21 & \textbf{\underline{74.47}} & 87.43 & 71.35 & 73.69 & 70.56 & 0.0012 & 0.0880 & 0.0089 & 0.1250 & 0.59 & 0.60 & 0.35 & 0.81 \\ \hline
\begin{tabular}[c]{@{}c@{}}Out-DAPT+TAPT\end{tabular} & 87.14 & 71.98 & 72.62 & 71.48 & 86.83 & 66.48 & 72.92 & 66.15 & 0.0008 & 0.2284 & 0.0364 & 0.2139 & 0.11 & 0.12 & 0.30 & 0.09 \\ \hline

\multicolumn{17}{|c|}{\textbf{QARiB}} \\ \hline
No-DAPT & 86.83 & 71.69 & 72.62 & 71.03 & 86.58 & 64.94 & 71.28 & 65.26 & 0.0009 & 0.3700 & 0.2164 & 0.2667 & N/A & N/A & N/A & N/A \\ \hline
\begin{tabular}[c]{@{}c@{}}In-DAPT\end{tabular} & \underline{88.29} & 70.61 & \underline{72.89} & 69.47 & \underline{87.17} & 65.39 & \underline{72.18} & 65.47 & 0.0095 & 0.2117 & 0.1736 & 0.1215 & 0.37 & 0.81 & 0.15 & 0.89 \\ \hline
\begin{tabular}[c]{@{}c@{}}Out-DAPT\end{tabular} & 85.29 & 69.74 & 69.59 & 69.91 & 84.90 & 66.40 & 67.91 & 66.34 & 0.0039 & 0.1272 & 0.0213 & 0.1517 & \dashuline{0.01} & 0.49 & 0.31 & 0.48 \\ \hline
TAPT & 86.83 & \underline{72.29} & 72.56 & \underline{72.17} & 86.33 & \underline{68.71} & 70.40 & \underline{68.86} & 0.0035 & 0.2678 & 0.0654 & 0.2566 & 0.66 & 0.31 & 0.85 & 0.29 \\ \hline
\begin{tabular}[c]{@{}c@{}}In-DAPT+TAPT\end{tabular} & 87.21 & 70.62 & 72.08 & 69.62 & 86.76 & 64.91 & 71.27 & 65.10 & 0.0017 & 0.2491 & 0.0447 & 0.1536 & 0.51 & 1.00 & 1.00 & 0.98 \\ \hline
\begin{tabular}[c]{@{}c@{}}Out-DAPT+TAPT\end{tabular} & 86.52 & 70.27 & 70.03 & 70.53 & 84.28 & 66.01 & 70.17 & 66.84 & 0.0630 & 0.2039 & 0.1933 & 0.1919 & 0.22 & 0.51 & 0.27 & 0.39 \\ \hline

\multicolumn{17}{|c|}{\textbf{MARBERTv2}} \\ \hline
No-DAPT & 87.40 & 68.32 & 70.76 & 67.34 & 87.12 & 67.65 & 69.63 & 65.48 & 0.0008 & 0.0413 & 0.0157 & 0.0275 & N/A & N/A & N/A & N/A \\ \hline
\begin{tabular}[c]{@{}c@{}}In-DAPT\end{tabular} & \underline{87.94} & \underline{72.35} & 72.83 & \underline{71.95} & \underline{87.39} & \underline{70.07} & \underline{72.62} & \underline{70.67} & 0.0025 & 0.0068 & 0.0030 & 0.0127 & 0.18 & 0.07 & 0.09 & 0.07 \\ \hline
\begin{tabular}[c]{@{}c@{}}Out-DAPT\end{tabular} & 87.25 & 68.49 & 71.72 & 67.55 & 86.93 & 68.50 & 70.61 & 66.85 & 0.0016 & 0.0132 & 0.0170 & 0.0126 & 0.34 & 0.48 & 0.18 & 0.48 \\ \hline
TAPT & 86.90 & 67.64 & 71.97 & 66.35 & 86.75 & 67.48 & 69.34 & 64.78 & 0.0002 & 0.0505 & 0.0592 & 0.0199 & 0.25 & 0.29 & 0.83 & 0.28 \\ \hline
\begin{tabular}[c]{@{}c@{}}In-DAPT+TAPT\end{tabular} & 87.52 & 69.33 & \underline{73.08} & 67.96 & 87.35 & 68.46 & 71.43 & 66.42 & 0.0002 & 0.0808 & 0.0356 & 0.0448 & 0.23 & 0.73 & 0.42 & 0.71 \\ \hline
\begin{tabular}[c]{@{}c@{}}Out-DAPT+TAPT\end{tabular} & 87.10 & 65.49 & 68.32 & 65.18 & 86.73 & 66.71 & 67.69 & 64.30 & 0.0021 & 0.0149 & 0.0112 & 0.0068 & 0.38 & 0.37 & 0.20 & 0.43 \\ \hline

\end{tabular}
\end{adjustbox}
\end{table*}

\begin{table*}[htp!]
\caption{Domain-Adapted Performance Metrics for Opinion Target Extraction (OTE) with Full Fine-Tuning. The highest result among all models is \textbf{\underline{bold and underlined}}, while the highest result for each Arabic BERT model is \underline{underlined}. Significant p-values (<0.05) are dashed underlined. F1 (Micro F1), P (Precision), R (Recall).}
\label{tab:OTEDA}
\centering
\tiny
\begin{adjustbox}{width=\textwidth}
\begin{tabular}{|l|c|c|c|c|c|c|c|c|c|c|c|c|}
\hline
\multirow{2}{*}{\begin{tabular}[c]{@{}c@{}}Adaptation\\ Technique\end{tabular}} & \multicolumn{3}{c|}{Maximum (\%)} & \multicolumn{3}{c|}{Average (\%)} & \multicolumn{3}{c|}{Variance (\%)} & \multicolumn{3}{c|}{P-value} \\ \cline{2-13}
 & \multicolumn{1}{c|}{F1} & \multicolumn{1}{c|}{P} & \multicolumn{1}{c|}{R} & \multicolumn{1}{c|}{F1} & \multicolumn{1}{c|}{P} & \multicolumn{1}{c|}{R} & \multicolumn{1}{c|}{F1} & \multicolumn{1}{c|}{P} & \multicolumn{1}{c|}{R} & \multicolumn{1}{c|}{F1} & \multicolumn{1}{c|}{P} & \multicolumn{1}{c|}{R} \\ \hline

\multicolumn{13}{|c|}{\textbf{CAMeLBERT-MSA}} \\ \hline
No-DAPT & 76.35 & 78.58 & 74.23 & 76.20 & 77.88 & 74.65 & 0.0002 & 0.0351 & 0.0340 & N/A & N/A & N/A \\ \hline
\begin{tabular}[c]{@{}c@{}}In-DAPT\end{tabular} & \textbf{\underline{77.02}} & 79.05 & 75.09 & \textbf{\underline{76.47}} & 78.46 & 74.58 & 0.0023 & 0.0048 & 0.0041 & 0.45 & 0.73 & 0.95 \\ \hline
\begin{tabular}[c]{@{}c@{}}Out-DAPT\end{tabular} & 76.56 & \textbf{\underline{79.72}} & 73.64 & 76.22 & \textbf{\underline{79.53}} & 73.18 & 0.0010 & 0.0048 & 0.0070 & 0.93 & 0.30 & 0.20 \\ \hline
TAPT & 76.48 & 77.11 & 75.86 & 76.29 & 77.63 & 75.01 & 0.0006 & 0.0048 & 0.0056 & 0.61 & 0.81 & 0.78 \\ \hline
\begin{tabular}[c]{@{}c@{}}In-DAPT+TAPT\end{tabular} & 76.43 & 78.63 & 74.36 & 76.32 & 78.79 & 73.99 & 0.0001 & 0.0007 & 0.0013 & 0.34 & 0.47 & 0.55 \\ \hline
\begin{tabular}[c]{@{}c@{}}Out-DAPT+TAPT\end{tabular} & 76.40 & 75.89 & \textbf{\underline{76.92}} & 76.05 & 76.05 & \textbf{\underline{76.05}} & 0.0011 & 0.0007 & 0.0076 & 0.29 & 0.21 & 0.30 \\ \hline

\multicolumn{13}{|c|}{\textbf{QARiB}} \\ \hline
No-DAPT & 76.31 & 76.94 & 75.69 & 76.09 & 76.78 & \underline{75.41} & 0.0004 & 0.0015 & 0.0013 & N/A & N/A & N/A \\ \hline
\begin{tabular}[c]{@{}c@{}}In-DAPT\end{tabular} & 76.75 & \underline{78.92} & 74.70 & \textbf{\underline{76.47}} & \underline{78.58} & 74.46 & 0.0010 & 0.0020 & 0.0005 & \dashuline{0.06} & \dashuline{0.00} & \dashuline{0.04} \\ \hline
\begin{tabular}[c]{@{}c@{}}Out-DAPT\end{tabular} & 74.62 & 76.60 & 72.74 & 74.13 & 76.19 & 72.19 & 0.0022 & 0.0167 & 0.0074 & \dashuline{0.01} & 0.38 & \dashuline{0.01} \\ \hline
TAPT & 76.45 & 77.11 & \underline{75.80} & 76.30 & 77.49 & 75.15 & 0.0004 & 0.0024 & 0.0069 & 0.45 & 0.10 & 0.72 \\ \hline
\begin{tabular}[c]{@{}c@{}}In-DAPT+TAPT\end{tabular} & \underline{76.79} & 78.37 & 75.26 & 76.45 & 78.47 & 74.54 & 0.0009 & 0.0006 & 0.0040 & 0.32 & \dashuline{0.01} & 0.13 \\ \hline
\begin{tabular}[c]{@{}c@{}}Out-DAPT+TAPT\end{tabular} & 70.27 & 75.76 & 71.48 & 73.06 & 75.59 & 70.70 & 0.0024 & 0.0011 & 0.0095 & \dashuline{0.01} & 0.07 & \dashuline{0.01} \\ \hline

\multicolumn{13}{|c|}{\textbf{MARBERTv2}} \\ \hline
No-DAPT & 76.64 & 77.19 & 76.10 & 76.20 & 75.67 & \underline{76.82} & 0.0037 & 0.0771 & 0.0306 & N/A & N/A & N/A \\ \hline
\begin{tabular}[c]{@{}c@{}}In-DAPT\end{tabular} & 76.54 & 76.60 & \underline{76.47} & 76.36 & \underline{77.54} & 75.23 & 0.0005 & 0.0090 & 0.0119 & 0.55 & 0.38 & 0.35 \\ \hline
\begin{tabular}[c]{@{}c@{}}Out-DAPT\end{tabular} & 74.44 & 73.10 & 75.82 & 74.04 & 74.07 & 74.04 & 0.0012 & 0.0090 & 0.0256 & 0.06 & 0.29 & \dashuline{0.00} \\ \hline
TAPT & 76.59 & 77.45 & 75.75 & 76.39 & 77.29 & 75.52 & 0.0003 & 0.0002 & 0.0005 & 0.56 & 0.40 & 0.37 \\ \hline
\begin{tabular}[c]{@{}c@{}}In-DAPT+TAPT\end{tabular} & \underline{76.95} & \underline{78.63} & 75.35 & \underline{76.44} & 76.73 & 76.20 & 0.0022 & 0.0404 & 0.0135 & 0.47 & 0.27 & 0.21 \\ \hline
\begin{tabular}[c]{@{}c@{}}Out-DAPT+TAPT\end{tabular} & 75.76 & 76.56 & 74.98 & 75.59 & 76.53 & 74.68 & 0.0002 & 0.0011 & 0.0011 & 0.21 & 0.68 & 0.21 \\ \hline

\end{tabular}
\end{adjustbox}
\end{table*}

\begin{table*}[htp!]
\caption{Performance Comparison of Classifiers for OTE extraction using In-DAPT CAMeLBERT-MSA. Micro F1 (F1), precision (P), and recall (R). The best values are highlighted in \textbf{bold}.}
\label{tab:OTE_classifiers}
\centering
\begin{adjustbox}{width=\textwidth}
\begin{tabular}{|l|c|c|c|c|c|c|c|c|c|c|c|c|c|c|c|c|}
\hline
\multirow{2}{*}{Classifier} & \multicolumn{4}{c|}{Maximum (\%)} & \multicolumn{4}{c|}{Average (\%)} & \multicolumn{4}{c|}{Variance (\%)} & \multicolumn{4}{c|}{P-value} \\ \cline{2-17}
 & \multicolumn{1}{c|}{F1} & \multicolumn{1}{c|}{P} & \multicolumn{1}{c|}{R} & & \multicolumn{1}{c|}{F1} & \multicolumn{1}{c|}{P} & \multicolumn{1}{c|}{R} & & \multicolumn{1}{c|}{F1} & \multicolumn{1}{c|}{P} & \multicolumn{1}{c|}{R} & & \multicolumn{1}{c|}{F1} & \multicolumn{1}{c|}{P} & \multicolumn{1}{c|}{R} & \\ \hline

Softmax & \textbf{77.02} & \textbf{79.05} & 75.09 & & 76.47 & 78.46 & 74.58 & & 0.0023 & 0.0048 & 0.0041 & & N/A & N/A & N/A & \\ \hline
1-Layer BiLSTM + Softmax & 76.57 & 74.89 & 78.33 & & 75.90 & 74.22 & 77.66 & & 0.0061 & 0.0047 & 0.0080 & & 0.34 & 0.03 & 0.01 & \\ \hline
2-Layer BiLSTM + Softmax & 75.95 & 76.41 & 76.60 & & 75.64 & 75.18 & 76.12 & & 0.0014 & 0.0159 & 0.0032 & & 0.06 & 0.01 & 0.12 & \\ \hline
1-Layer BiGRU + Softmax & 76.67 & 75.12 & \textbf{79.85} & & 76.41 & 73.83 & 79.19 & & 0.0005 & 0.0125 & 0.0065 & & 0.89 & 0.02 & 0.00 & \\ \hline
2-Layer BiGRU + Softmax & 76.34 & 75.83 & 78.54 & & 76.03 & 74.26 & 77.92 & & 0.0014 & 0.0204 & 0.0086 & & 0.30 & 0.03 & 0.00 & \\ \hline
CRF & 76.23 & 74.90 & 79.64 & & 76.08 & 74.19 & 78.09 & & 0.0002 & 0.0140 & 0.0191 & & 0.38 & 0.01 & 0.06 & \\ \hline
1-Layer BiLSTM + CRF & 76.38 & 74.34 & 78.79 & & 75.76 & 74.12 & 77.47 & & 0.0038 & 0.0004 & 0.0202 & & 0.06 & 0.01 & 0.08 & \\ \hline
2-Layer BiLSTM + CRF & 76.12 & 77.28 & 79.30 & & 75.90 & 74.70 & 77.23 & & 0.0011 & 0.0679 & 0.0464 & & 0.17 & 0.10 & 0.11 & \\ \hline
1-Layer BiGRU + CRF & 76.35 & 75.34 & 77.45 & & 76.18 & 75.32 & 77.07 & & 0.0003 & 0.0000 & 0.0016 & & 0.42 & 0.02 & 0.05 & \\ \hline
2-Layer BiGRU + CRF & 76.49 & 74.90 & 79.13 & & 76.24 & 74.40 & 78.19 & & 0.0015 & 0.0024 & 0.0165 & & 0.43 & 0.01 & 0.01 & \\ \hline

\end{tabular}
\end{adjustbox}
\end{table*}

\subsection{Results: Domain Adaptation with Adapters Fine-tuning}
\label{subsec:results-adapters}
Freezing all layers except the classification layer resulted in suboptimal performance, as detailed in Tables \ref{tab:arabic_bert_aspect_sentiment_classification} and \ref{tab:arabic_bert_ote_full_hyperparameter_performance}. Although fully fine-tuning all model parameters improved performance, the gains were not statistically significant when applied to domain-adapted models, as shown in Tables \ref{tab:SPASC} and \ref{tab:OTEDA}. To address this, we employ adapters as a more efficient alternative for fine-tuning, which requires fewer task-specific parameter updates. Furthermore, we utilized adapter fusing to enable parameter sharing between \gls{asc} and \gls{ote} tasks.

The adapter type with the highest performance varied for both \gls{asc} and \gls{ote}. For \gls{asc} (Table \ref{tab:ASC-Adapters}), Pfeiffer (No-\gls{dapt}) achieved the highest accuracy (87.52\%), while Houlsby (No-\gls{dapt}) had the highest average (87.39\%). The highest macro-F1 score was achieved by Parallel (73.64\%), with Pfeiffer \gls{tapt} stacked adapters achieving the highest average (71.82\%). For \gls{ote} (Table \ref{tab:OTEAdapters}), Houlsby with \gls{indapt} led both maximum and average F1 scores (76.31\% and 76.11\%). The lowest scores were recorded by Parallel for \gls{asc} (87.40\% accuracy) and Pfeiffer for \gls{ote} (75.87\% F1).

\begin{figure}[htp!]
    \centering
    \begin{minipage}{0.75\linewidth}
        \includegraphics[width=\linewidth]{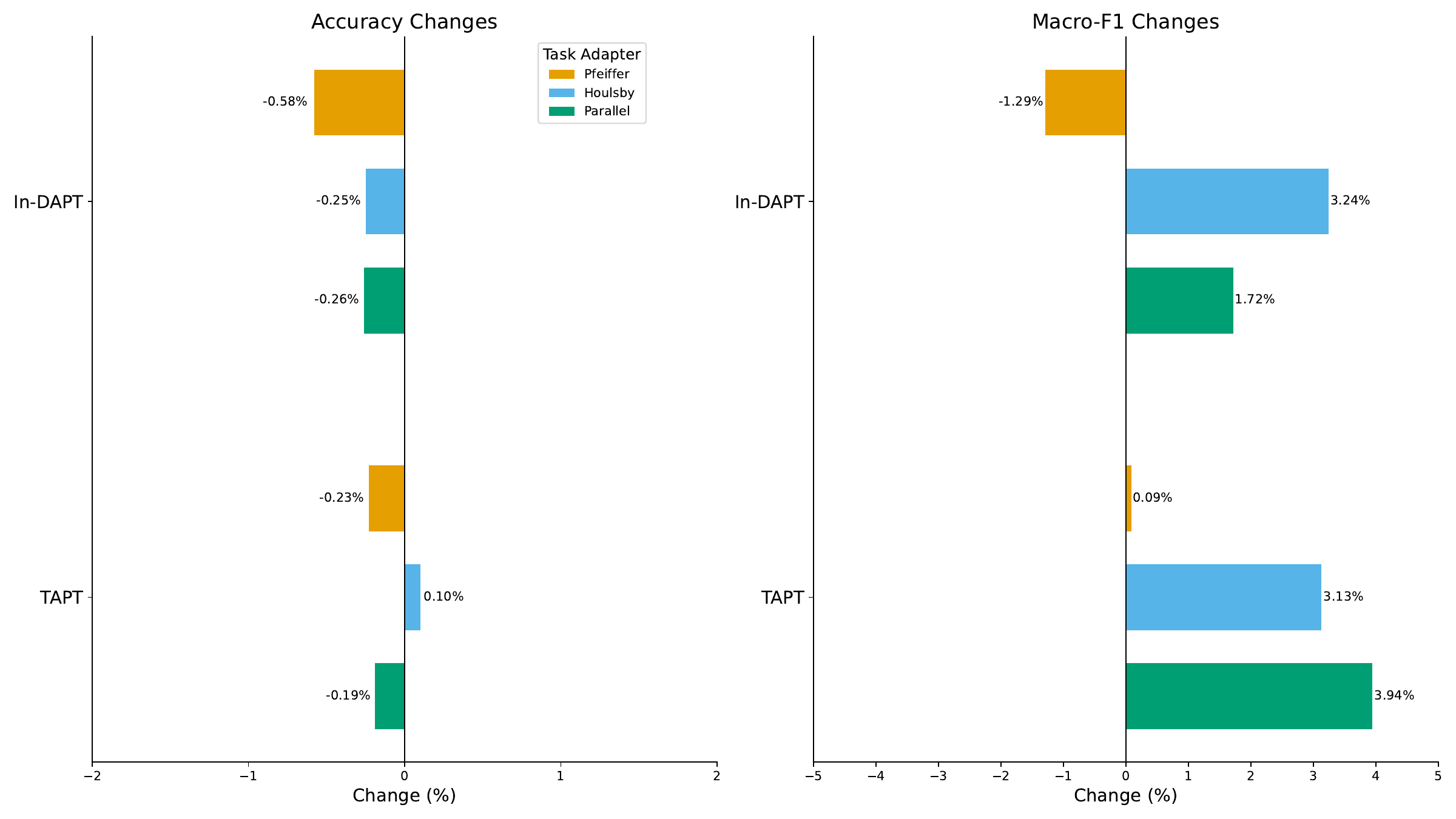}
       \caption{Accuracy and Macro-F1 change for CAMeLBERT-MSA with domain adaptation (\gls{indapt}, \gls{tapt}) using adapters for aspect-sentiment classification. The zero line represents the baseline task adapter without domain adaptation. Bars indicate percentage changes: positive for improvements and negative for declines.}
        \label{fig:ASC_Adapter_graph}
    \end{minipage}\hfill
    \begin{minipage}{0.75\linewidth}
        \includegraphics[width=\linewidth]{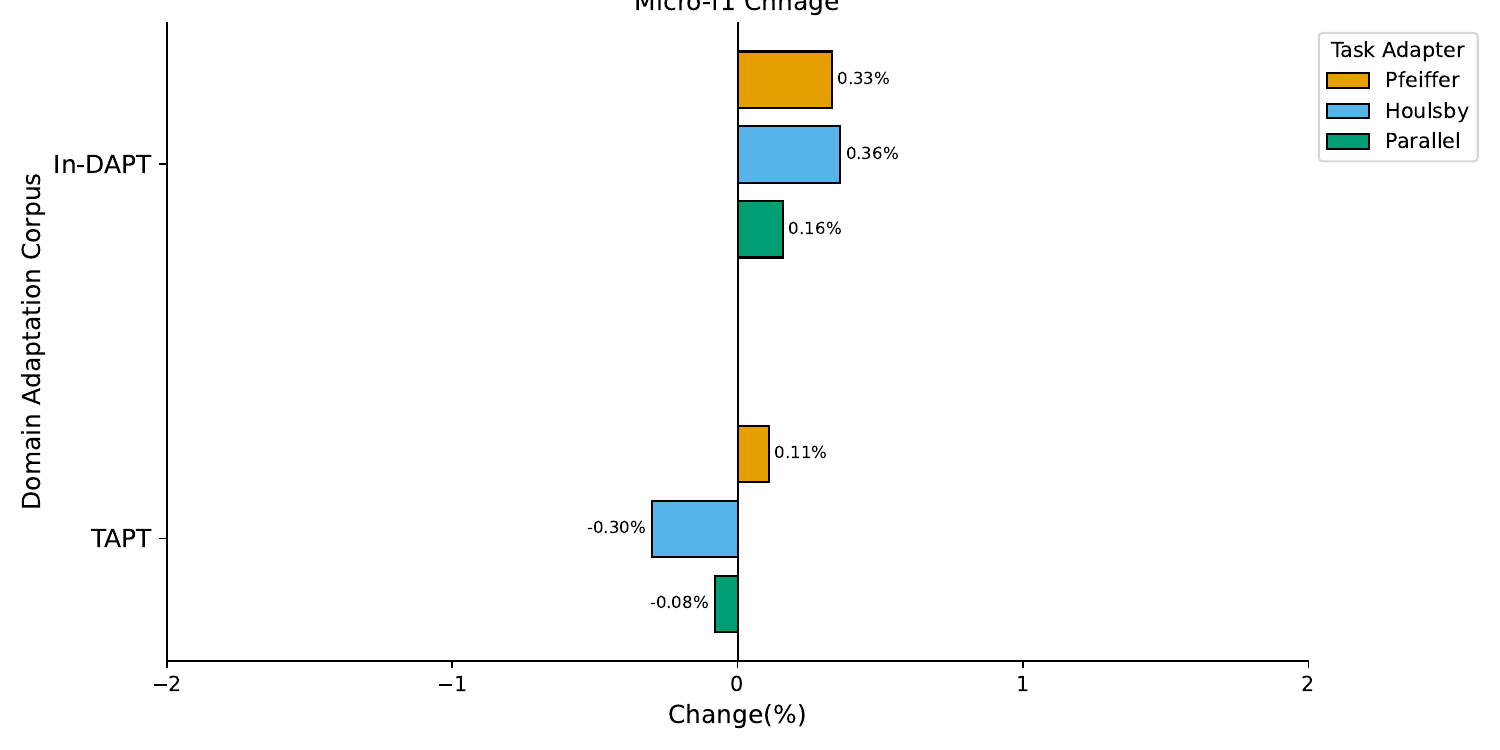}
        \caption{Micro-F1 change for CAMeLBERT-MSA with domain adaptation (\gls{indapt}, \gls{tapt}) using adapters for \gls{ote} extraction. The zero line represents the baseline task adapter without domain adaptation. Bars indicate percentage changes: positive for improvements and negative for declines.}
        \label{fig:OTE_Adapter_graph}
    \end{minipage}
\end{figure}

Figures \ref{fig:ASC_Adapter_graph} and \ref{fig:OTE_Adapter_graph} illustrate the effects of stacking domain-adapted language adapters (\gls{indapt}, \gls{tapt}) with task adapters compared to baseline task adapters alone. In the case of \gls{asc}, accuracy decreased across all configurations, with Pfeiffer showing the largest decrease (-0.58\%), while macro-F1 scores improved significantly for Houlsby (+3.24\%) and Parallel (+1.72\%), except for a decrease observed with Pfeiffer (-1.29\%). For \gls{ote}, micro-F1 showed slight improvements with \gls{indapt}, particularly for Houlsby (+0.36\%), whereas \gls{tapt} generally resulted in minimal or negative changes.

\begin{table*}[ht]
\caption{Aspect Sentiment Classification Adapter Performance Metrics. The highest result among all models is \textbf{\underline{bold and underlined}}, while the highest result for each adapter model is \underline{underlined}. Significant p-values (<0.05) are dashed underlined. Acc (Accuracy), M-F1 (Macro F1), M-P (Macro Precision), M-R (Macro Recall).}
\label{tab:ASC-Adapters}
\centering
\begin{adjustbox}{width=\textwidth}
\begin{tabular}{|l|c|c|c|c|c|c|c|c|c|c|c|c|c|c|c|c|}
\hline
\multirow{2}{*}{\begin{tabular}[c]{@{}c@{}}Adaptation\\ Technique\end{tabular}} & \multicolumn{4}{c|}{Maximum (\%)} & \multicolumn{4}{c|}{Average (\%)} & \multicolumn{4}{c|}{Variance (\%)} & \multicolumn{4}{c|}{P-value} \\ \cline{2-17}
 & \multicolumn{1}{c|}{Acc} & \multicolumn{1}{c|}{M-F1} & \multicolumn{1}{c|}{M-P} & \multicolumn{1}{c|}{M-R} & \multicolumn{1}{c|}{Acc} & \multicolumn{1}{c|}{M-F1} & \multicolumn{1}{c|}{M-P} & \multicolumn{1}{c|}{M-R} & \multicolumn{1}{c|}{Acc} & \multicolumn{1}{c|}{M-F1} & \multicolumn{1}{c|}{M-P} & \multicolumn{1}{c|}{M-R} & \multicolumn{1}{c|}{Acc} & \multicolumn{1}{c|}{M-F1} & \multicolumn{1}{c|}{M-P} & \multicolumn{1}{c|}{M-R} \\ \hline

\multicolumn{17}{|c|}{\textbf{Pfeiffer}} \\ \hline
Task Adapter & \underline{\textbf{87.52}} & 72.08 & 72.13 & 72.07 & 86.92 & 69.85 & 72.92 & 69.32 & 0.0027 & 0.0882 & 0.0304 & 0.1250 & N/A & N/A & N/A & N/A \\ \hline
\begin{tabular}[c]{@{}c@{}}In-DAPT Lang+\\ Task Stacked Adapter\end{tabular} & 86.94 & 70.79 & 71.79 & 70.16 & 86.75 & 70.23 & 72.58 & 69.11 & 0.0003 & 0.0029 & 0.0402 & 0.0095 & 0.50 & 0.85 & 0.43 & 0.91 \\ \hline
\begin{tabular}[c]{@{}c@{}}TAPT Lang+\\ Task Stacked Adapter\end{tabular} & 87.29 & \underline{72.17} & \underline{73.38} & 71.22 & \underline{87.14} & \underline{\textbf{71.82}} & \underline{73.16} & \underline{70.95} & 0.0002 & 0.0029 & 0.0005 & 0.0035 & 0.43 & 0.40 & 0.82 & 0.53 \\ \hline

\multicolumn{17}{|c|}{\textbf{Houlsby}} \\ \hline
Task Adapter & 87.39 & 68.39 & 73.02 & 67.14 & \underline{\textbf{87.39}} & \underline{67.27} & 74.44 & 66.27 & 0.0000 & 0.0212 & 0.0241 & 0.0124 & N/A & N/A & N/A & N/A \\ \hline
\begin{tabular}[c]{@{}c@{}}In-DAPT Lang+\\ Task Stacked Adapter\end{tabular} & 87.14 & 71.63 & 73.01 & \underline{70.71} & 87.05 & 67.11 & 74.52 & 66.46 & 0.0001 & 0.1673 & 0.0369 & 0.1422 & \dashuline{0.00} & 0.94 & 0.92 & 0.93 \\ \hline
\begin{tabular}[c]{@{}c@{}}TAPT Lang+\\ Task Stacked Adapter\end{tabular} & \underline{87.49} & \underline{71.52} & \underline{75.10} & 70.03 & 87.25 & 67.11 & \underline{75.90} & \underline{66.60} & 0.0007 & 0.1732 & 0.0401 & 0.0954 & 0.54 & 0.96 & 0.52 & 0.90 \\ \hline

\multicolumn{17}{|c|}{\textbf{Parallel}} \\ \hline
Task Adapter & \underline{87.40} & 69.70 & 71.71 & 68.75 & \underline{87.05} & 66.31 & 73.67 & 65.86 & 0.0027 & 0.1044 & 0.0440 & 0.0672 & N/A & N/A & N/A & N/A \\ \hline
\begin{tabular}[c]{@{}c@{}}In-DAPT Lang+\\ Task Stacked Adapter\end{tabular} & 87.14 & 71.42 & 71.04 & 71.94 & 86.64 & \underline{68.12} & \underline{74.63} & \underline{68.10} & 0.0059 & 0.2314 & 0.2315 & 0.2038 & 0.59 & 0.43 & 0.67 & 0.44 \\ \hline
\begin{tabular}[c]{@{}c@{}}TAPT Lang+\\ Task Stacked Adapter\end{tabular} & 87.21 & \textbf{\underline{73.64}} & \underline{72.69} & \underline{75.19} & 86.53 & 65.29 & 68.05 & 66.53 & 0.0035 & 0.5641 & 0.8603 & 0.5787 & 0.19 & 0.86 & 0.33 & 0.91 \\ \hline

\end{tabular}
\end{adjustbox}
\end{table*}

\begin{table*}[htp!]
\caption{Opinion Target Expression extraction Adapters Performance Results. The highest result among all models is \textbf{\underline{bold and underlined}}, while the highest result for each adapter model is \underline{underlined}. Significant p-values (<0.05) are dashed underlined. F1 (Micro F1), P (Precision), R (Recall).}
\centering
\tiny
\begin{adjustbox}{width=\textwidth}
\begin{tabular}{|l|l|l|l|l|l|l|l|l|l|l|l|l|l|l|l|l|}
\hline
\multirow{2}{*}{\begin{tabular}[c]{@{}c@{}}Adaptation\\ Technique\end{tabular}} & \multicolumn{3}{c|}{Maximum} & \multicolumn{3}{c|}{Average} & \multicolumn{3}{c|}{Variance} & \multicolumn{3}{c|}{P-value} \\ \cline{2-13}
 & \multicolumn{1}{c|}{F1} & \multicolumn{1}{c|}{P} & \multicolumn{1}{c|}{R} & \multicolumn{1}{c|}{F1} & \multicolumn{1}{c|}{P} & \multicolumn{1}{c|}{R} & \multicolumn{1}{c|}{F1} & \multicolumn{1}{c|}{P} & \multicolumn{1}{c|}{R} & \multicolumn{1}{c|}{F1} & \multicolumn{1}{c|}{P} & \multicolumn{1}{c|}{R} \\ \hline

\multicolumn{13}{|c|}{\textbf{Pfeiffer}} \\ \hline
Task Adapter & 75.54 & 75.04 & 76.04 & 75.23 & 75.35 & 75.11 & 0.0008 & 0.0059 & 0.0088 & N/A & N/A & N/A \\ \hline
\begin{tabular}[c]{@{}c@{}}In-DAPT Lang+\\ Task Stacked Adapter\end{tabular} & \underline{75.87} & 75.63 & 76.11 & \underline{75.74} & 76.21 & 75.31 & 0.0002 & 0.0201 & 0.0233 & 0.09 & 0.52 & 0.87 \\ \hline
\begin{tabular}[c]{@{}c@{}}TAPT Lang+\\ Task Stacked Adapter\end{tabular} & 75.65 & 76.01 & 75.28 & 75.33 & 75.96 & 74.72 & 0.0008 & 0.0079 & 0.0119 & \underline{0.05} & 0.59 & 0.70 \\ \hline

\multicolumn{13}{|c|}{\textbf{Houlsby}} \\ \hline
Task Adapter & 75.95 & 77.15 & 74.79 & 75.58 & 76.48 & 74.72 & 0.0011 & 0.0052 & 0.0035 & N/A & N/A & N/A \\ \hline
\begin{tabular}[c]{@{}c@{}}In-DAPT Lang+\\ Task Stacked Adapter\end{tabular} & \underline{\textbf{76.31}} & 77.07 & 75.57 & \underline{\textbf{76.11}} & 76.81 & 75.42 & 0.0005 & 0.0035 & 0.0069 & 0.25 & 0.12 & 0.30 \\ \hline
\begin{tabular}[c]{@{}c@{}}TAPT Lang+\\ Task Stacked Adapter\end{tabular} & 75.65 & 76.48 & 74.83 & 75.57 & 76.03 & 75.15 & 0.0001 & 0.0135 & 0.0109 & 0.97 & 0.61 & 0.68 \\ \hline

\multicolumn{13}{|c|}{\textbf{Parallel}} \\ \hline
Task Adapter & 75.85 & 75.30 & 76.42 & \underline{75.68} & 75.94 & 75.44 & 0.0003 & 0.0142 & 0.0205 & N/A & N/A & N/A \\ \hline
\begin{tabular}[c]{@{}c@{}}In-DAPT Lang+\\ Task Stacked Adapter\end{tabular} & \underline{76.01} & 76.22 & 75.81 & 75.52 & 75.89 & 75.16 & 0.0053 & 0.0043 & 0.0066 & 0.78 & 0.93 & 0.82 \\ \hline
\begin{tabular}[c]{@{}c@{}}TAPT Lang+\\ Task Stacked Adapter\end{tabular} & 75.77 & 77.11 & 74.47 & 75.52 & 75.28 & 75.80 & 0.0007 & 0.0250 & 0.0142 & 0.60 & 0.10 & 0.37 \\ \hline

\end{tabular}
\end{adjustbox}
\label{tab:OTEAdapters}
\end{table*}

\subsection{Full Fine-tuning vs.\ Adapter Fine-tuning}
\label{subsec:full-vs-adapter}
\begin{figure}[!ht]
    \centering
    \begin{minipage}{0.8\linewidth}
        \includegraphics[width=\linewidth]{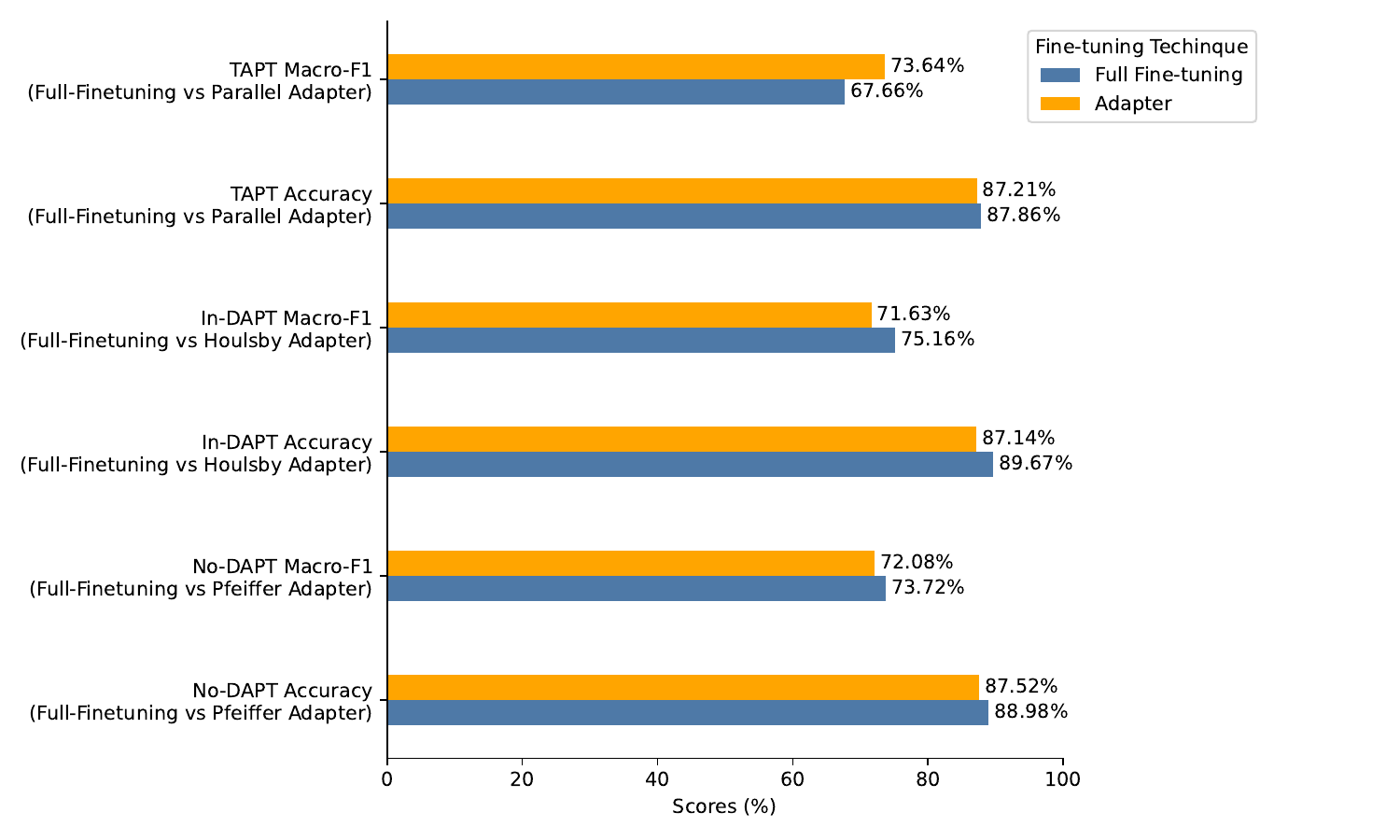}
        \caption{Comparison of accuracy and macro-F1 scores for aspect-sentiment classification using CAMeLBERT-MSA with full fine-tuning vs.\ adapters in different adaptation settings (\gls{indapt}, \gls{tapt}, No-\gls{dapt}).}
        \label{fig:ASC_Full_Adapter_Tunining_graph}
    \end{minipage}\hfill
    \begin{minipage}{0.8\linewidth}
        \includegraphics[width=\linewidth]{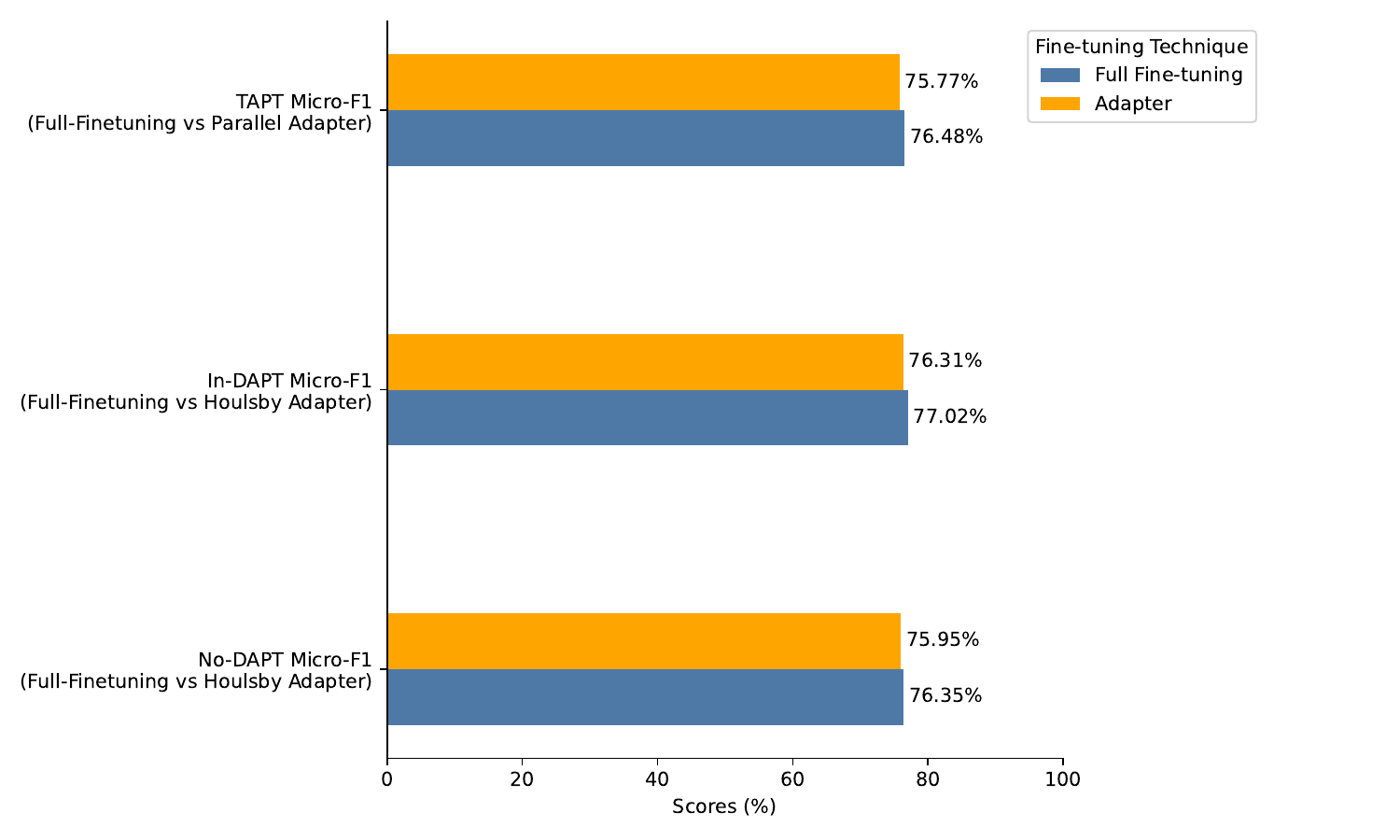}
        \caption{Comparison of micro-F1 scores for opinion target extraction using CAMeLBERT-MSA with full fine-tuning vs.\ adapters in different adaptation settings (\gls{indapt}, \gls{tapt}, No-\gls{dapt}).}
        \label{fig:OTE_Full_Adapter_Tunining_graph}
    \end{minipage}
\end{figure}

Figures \ref{fig:ASC_Full_Adapter_Tunining_graph} and \ref{fig:OTE_Full_Adapter_Tunining_graph} compare the performance of full fine-tuning and adapter fine-tuning for CAMeLBERT-MSA for \gls{asc} and \gls{ote} tasks across \gls{indapt}, \gls{tapt}, and baseline No-\gls{dapt} settings. For \gls{asc}, accuracy was consistently higher with full fine-tuning across all settings, with the largest margin in \gls{indapt} (+2.53\%), followed by No-\gls{dapt} (+1.46\%) and \gls{tapt} (+0.65\%). Macro-F1 scores improved significantly for adapters in \gls{tapt} (+6\%), while full fine-tuning performed better in \gls{indapt} (+3.53\%) and No-\gls{dapt} (+1.64\%). For \gls{ote}, micro-F1 showed minimal differences, with full fine-tuning leading by +0.71\% in both \gls{indapt} and \gls{tapt}, and +0.40\% in No-\gls{dapt}.

\subsection{Performance--Efficiency Trade-offs}
\label{subsec:efficiency}
\begin{figure}[htp!]
    \centering
    \begin{minipage}{0.80\linewidth}
        \includegraphics[width=\linewidth]{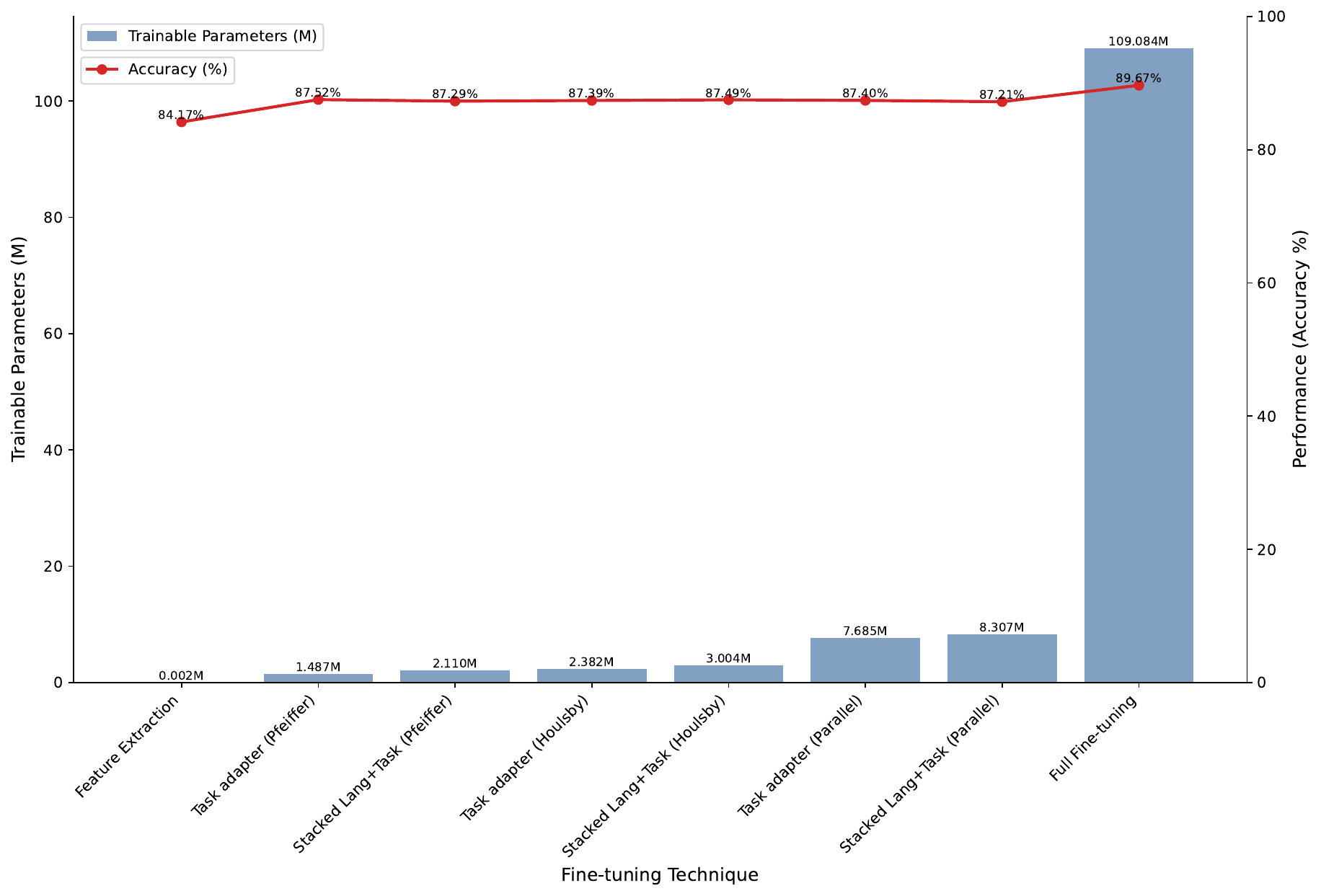}
        \caption{Trainable parameters (M) vs.\ Accuracy for aspect-sentiment classification using CAMeLBERT-MSA. Compares full fine-tuning, feature extraction, and adapters (task (No-\gls{dapt}), \gls{indapt}, \gls{tapt}).}
        \label{fig:ASC_Fine_Tuning_Paramters_graph}
    \end{minipage}\hfill
    \begin{minipage}{0.80\linewidth}
        \includegraphics[width=\linewidth]{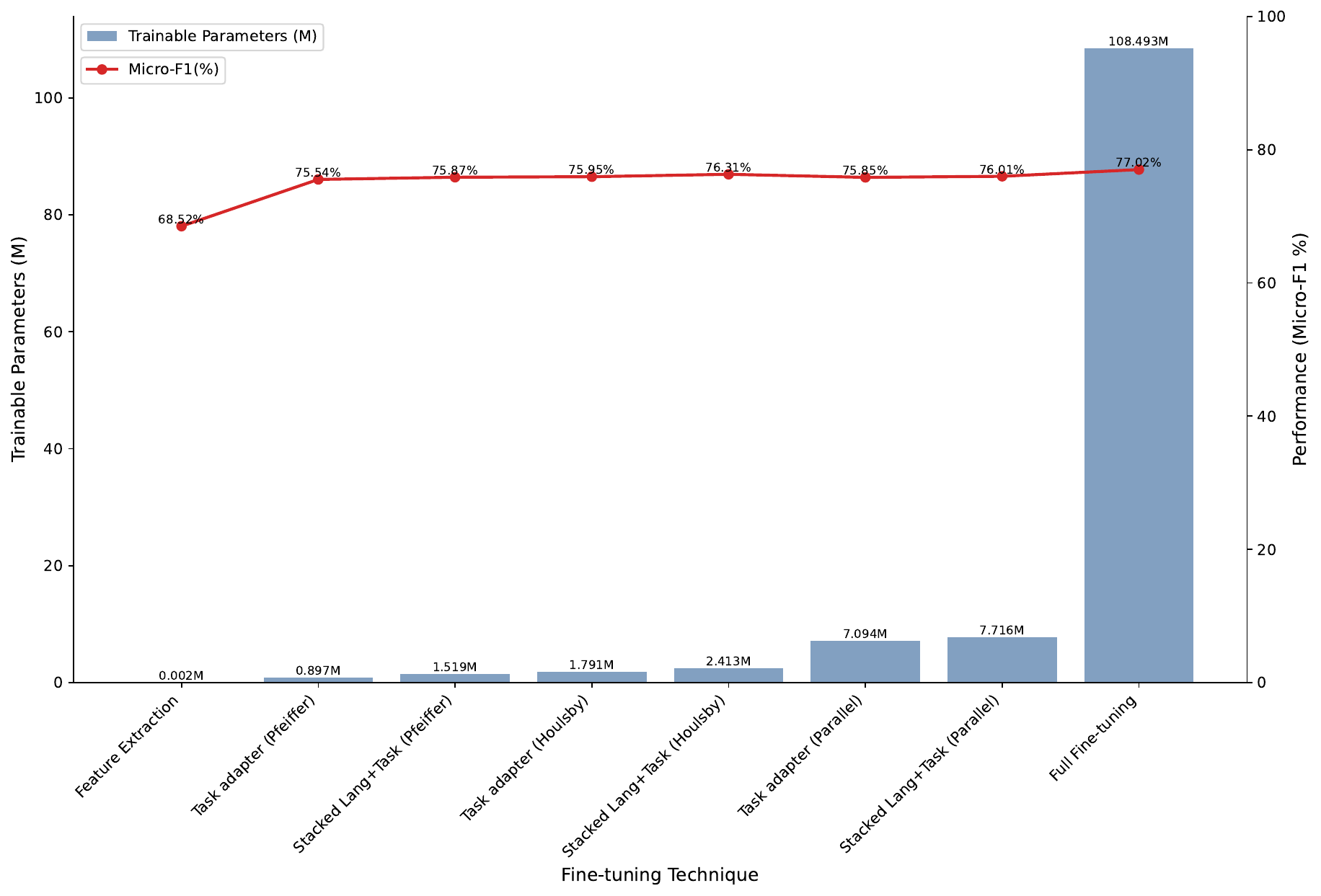}
       \caption{Trainable parameters (M) vs.\ Micro-F1 for \gls{ote} extraction using CAMeLBERT-MSA. Compares full fine-tuning, feature extraction, and adapters (task (No-\gls{dapt}), \gls{indapt}, \gls{tapt}).}
        \label{fig:OTE_Fine_Tuning_Paramters_graph}
    \end{minipage}
\end{figure}

Figures \ref{fig:ASC_Fine_Tuning_Paramters_graph} and \ref{fig:OTE_Fine_Tuning_Paramters_graph} compare fine-tuning techniques: feature extraction, adapter fine-tuning, and full fine-tuning for \gls{asc} (accuracy) and \gls{ote} (micro-F1).

For \gls{asc}, full fine-tuning (89.67\%, 109.084M parameters) outperforms feature extraction (84.17\%, 0.002M) and Pfeiffer adapter (87.52\%, 1.487M). Pfeiffer adapters reduce parameters by 98.6\% with only a 2.15\% accuracy loss. Similarly, for \gls{ote}, full fine-tuning leads with 77.02\% micro-F1 (108.493M parameters), followed by Houlsby adapters (76.31\%, 2.413M), which cut parameters by 97.8\% with just a 0.89\% performance drop. Although feature extraction is the most parameter-efficient method (0.002M), it sacrifices performance, with drops of 5.5\% and 8.5\% for \gls{asc} and \gls{ote}, respectively, compared to full fine-tuning. Adapters, on the other hand, balance efficiency and performance, achieving competitive results with far fewer trainable parameters.

\subsection{Adapter Fusion}
\label{subsec:adapter-fusion}
Adapters improve computational efficiency by reducing the number of trainable parameters. However, none of the adapters achieves higher performance than the fully fine-tuned \gls{indapt} CAMeLBERT-MSA. As highlighted in the literature, parameter sharing between \gls{ote} extraction and \gls{asc} tasks is essential due to their strong correlation \citep{Ruidan2019,YANG2021344}. Therefore, we used adapter fusion \citep{pfeiffer-etal-2021-adapterfusion}, which allows parameter sharing across tasks while maintaining computational efficiency. It achieves this by freezing task-specific adapters and the pre-trained model parameters, while adding only fusion parameters that learn to integrate adapter outputs. This approach optimizes the target task's performance by leveraging relevant knowledge.

\begin{figure*}[htp!]
    \centering
    \includegraphics[width=0.9\textwidth]{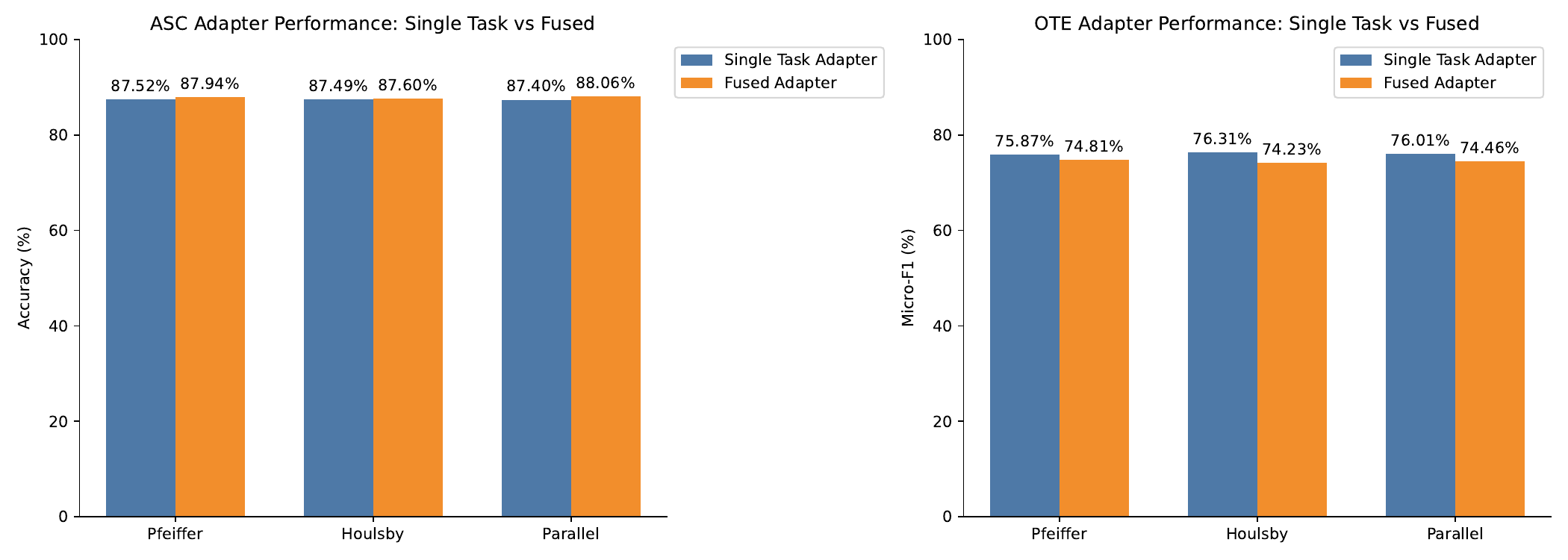}
    \caption{Comparison of single-task adapter vs.\ fused adapter performance. The results show the highest-performing adapters for each task (No-\gls{dapt}, \gls{tapt}, or \gls{indapt}) in aspect-sentiment classification (accuracy) and \gls{ote} extraction (Micro-F1) using CAMeLBERT-MSA.}
    \label{fig:Adapers_single_vs_fusion_graph}
\end{figure*}

As shown in Table \ref{tab:FusedAdapter} and Figure \ref{fig:Adapers_single_vs_fusion_graph}, independent adapters outperformed fused adapters for both \gls{asc} and \gls{ote} tasks. For \gls{asc}, the parallel fused adapter achieved the highest maximum and average accuracy, with values of 88.06\% and 87.43\%, respectively. For \gls{ote}, the fused Pfeiffer adapter yielded the highest F1-score of 74.59\%.

\begin{table*}[htp!]
\caption{Adapter Fusion Performance for Aspect Sentiment Classification and OTE Extraction. The highest result among all models is \textbf{bold}. Significant p-values (<0.05) are \underline{underlined}.}

\label{tab:FusedAdapter}
\centering
\tiny
\begin{adjustbox}{width=\textwidth}
\begin{tabular}{|l|cccc|cccc|cccc|cccc|}
\hline
\multirow{2}{*}{Adapter} & \multicolumn{4}{c|}{Maximum (\%)} & \multicolumn{4}{c|}{Average (\%)} & \multicolumn{4}{c|}{Variance (\%)} & \multicolumn{4}{c|}{P-value} \\ \cline{2-17}
 & \multicolumn{2}{c|}{F1} & \multicolumn{1}{c|}{P} & \multicolumn{1}{c|}{R} & \multicolumn{2}{c|}{F1} & \multicolumn{1}{c|}{P} & \multicolumn{1}{c|}{R} & \multicolumn{2}{c|}{F1} & \multicolumn{1}{c|}{P} & \multicolumn{1}{c|}{R} & \multicolumn{2}{c|}{F1} & \multicolumn{1}{c|}{P} & \multicolumn{1}{c|}{R} \\ \hline

\multicolumn{17}{|c|}{\textbf{Adapter Fusion OTE Extraction}} \\ \hline
Houlsby & 74.23 & & 75.55 & 72.96 & 74.20 && 75.55 & 72.90 & 0.0000 && 0.0009 & 0.0011 & \underline{0.01} && 0.09 & 0.05 \\ \hline
Pfeiffer & \textbf{74.81} && 75.00 & 74.62 & \textbf{74.60} && 75.37 & 73.87 & 0.0004 && 0.0130 & 0.0188 & 0.01 && 0.61 & 0.43 \\ \hline
Parallel & 74.46 && 73.19 & 75.77 & 74.26 && 74.13 & 74.42 & 0.0004 && 0.0127 & 0.0168 & \underline{0.02} && \underline{0.01} & 0.21 \\ \hline

\multicolumn{17}{|c|}{\textbf{Adapter Fusion Aspect Sentiment Classification}} \\ \hline
\multirow{2}{*}{} & \multicolumn{1}{c|}{Acc} & \multicolumn{1}{c|}{M-F1} & \multicolumn{1}{c|}{M-P} & \multicolumn{1}{c|}{M-R} & \multicolumn{1}{c|}{Acc} & \multicolumn{1}{c|}{M-F1} & \multicolumn{1}{c|}{M-P} & \multicolumn{1}{c|}{M-R} & \multicolumn{1}{c|}{Acc} & \multicolumn{1}{c|}{M-F1} & \multicolumn{1}{c|}{M-P} & \multicolumn{1}{c|}{M-R} & \multicolumn{1}{c|}{Acc} & \multicolumn{1}{c|}{M-F1} & \multicolumn{1}{c|}{M-P} & \multicolumn{1}{c|}{M-R} \\ \cline{2-17}
Houlsby & 87.60 & 72.83 & 74.67 & 71.65 & 87.12 & 72.18 & 73.39 & 71.38 & 0.0017 & 0.0044 & 0.0124 & 0.0035 & 0.33 & \underline{0.04} & 0.45 & \underline{0.03} \\ \hline
Pfeiffer & 87.94 & 74.03 & 75.31 & 73.19 & 86.85 & 72.77 & 72.96 & 72.57 & 0.0108 & 0.0142 & 0.0422 & 0.0175 & 0.64 & 0.18 & 0.54 & 0.05 \\ \hline
Parallel & \textbf{88.06} & 73.80 & 76.65 & 72.28 & \textbf{87.43} & 73.01 & 74.45 & 72.27 & 0.0049 & 0.0046 & 0.0446 & 0.0039 & 0.57 & 0.08 & 0.75 & 0.07 \\ \hline

\end{tabular}
\end{adjustbox}
\end{table*}

\subsection{Effects of Imbalance Mitigation (Results)}
\label{subsec:imbalance-results}
\begin{figure*}[tb]
    \centering
    \includegraphics[width=0.8\linewidth]{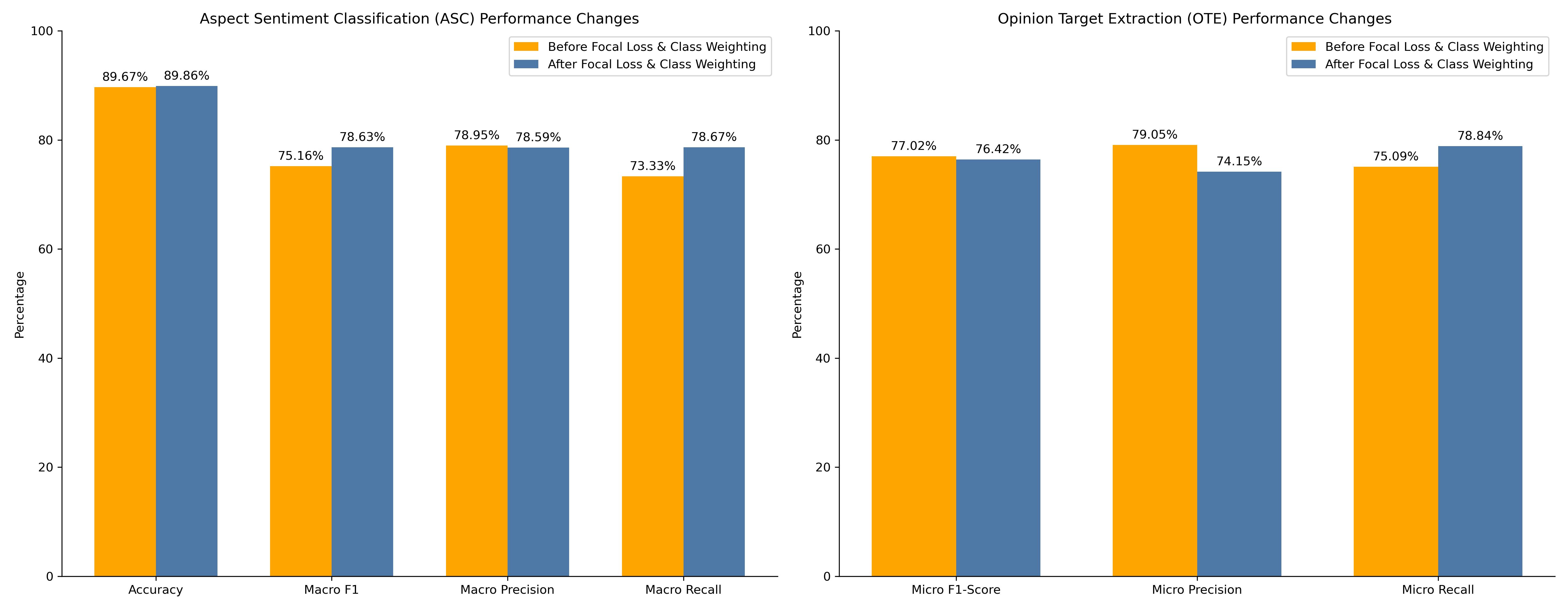}
    \caption{Comparison of performance before and after applying focal loss and class weighting for aspect-sentiment classification (accuracy, macro-F1, macro precision, macro recall) and opinion target extraction (micro-F1, micro precision, micro recall) using \gls{indapt} CAMeLBERT-MSA. The results highlight the impact of different training strategies on the model's performance.}
    \label{fig:Adapers_CE_vs_FL_graph}
\end{figure*}

Focal loss with a gamma value of 2, combined with inverse frequency class weighting with an added constant (0.5 for OTE and 0.8 for ASC), achieved the best performance for both tasks using the \gls{indapt} CAMeLBERT-MSA. As shown in Table \ref{tab:imbalncedata} and Figure \ref{fig:Adapers_CE_vs_FL_graph}, this approach achieved the highest accuracy for \gls{asc}, with a maximum of 89.86\% and an average of 89\%. Compared to cross-entropy without class weighting, this reflects a slight accuracy improvement from 89.67\% (+0.19\%) and a substantial increase in macro-F1 from 75.16\% to 78.63\% (+3.47\%). Recall showed the most significant gain, rising from 73.33\% to 78.67\% (+5.34\%), while precision saw a minor decrease from 78.95\% to 78.59\% (-0.36\%). For \gls{ote}, the impact was mixed. While focal loss decreased the micro-F1 from 77.02\% to 76.42\% (-0.60\%), recall improved significantly from 75.09\% to 78.84\% (+3.75\%). However, precision dropped from 79.05\% to 74.15\% (-4.90\%), indicating that the model adjusted its threshold to favor recall. The p-value in Table \ref{tab:imbalncedata} confirms the statistical significance of these changes.

\begin{table*}[htp!]
\caption{Impact of Focal Loss on In-Domain Adapted CAMeLBERT-MSA for ASC and OTE Extraction. Acc: Accuracy, M-F1: Macro F1, M-P: Macro Precision, M-R: Macro Recall, F1: Micro F1, P: Micro Precision, M-R: Micro Recall. The highest result among all models is \textbf{bold}. Significant p-values (<0.05) are \underline{underlined}. }
\label{tab:imbalncedata}
\centering
\begin{adjustbox}{width=\textwidth}
\begin{tabular}{|l|c|c|c|c|c|c|c|c|c|c|c|c|c|c|c|c|}
\hline
\multirow{2}{*}{\begin{tabular}[c]{@{}c@{}}Class Weight List (\(\alpha\))\end{tabular}} & \multicolumn{4}{c|}{Maximum (\%)} & \multicolumn{4}{c|}{Average (\%)} & \multicolumn{4}{c|}{Variance (\%)} & \multicolumn{4}{c|}{P-value} \\ \cline{2-17}
 & \multicolumn{2}{c|}{F1} & \multicolumn{1}{c|}{P} & \multicolumn{1}{c|}{R} & \multicolumn{2}{c|}{F1} & \multicolumn{1}{c|}{P} & \multicolumn{1}{c|}{R} & \multicolumn{2}{c|}{F1} & \multicolumn{1}{c|}{P} & \multicolumn{1}{c|}{R} & \multicolumn{2}{c|}{F1} & \multicolumn{1}{c|}{P} & \multicolumn{1}{c|}{R} \\ \hline

\multicolumn{17}{|c|}{\textbf{Opinion Target Expression (\gls{ote}) Extraction}} \\ \hline
\begin{tabular}[c]{@{}c@{}}\{'O': 0.2063, \\ 'B-OTE': 0.2641, \\ 'I-OTE': 0.5295\}\end{tabular} & \multicolumn{2}{c|}{76.42} & 74.15 & 78.84 & \multicolumn{2}{c|}{76.24} & 73.87 & 78.77 & \multicolumn{2}{c|}{0.0005} & 0.0056 & 0.0024 & \multicolumn{2}{c|}{0.36} & \underline{0.02} & \underline{0.02} \\ \hline

\multicolumn{17}{|c|}{\textbf{Aspect Sentiment Classification (\gls{asc})}} \\ \hline
\multirow{2}{*}{} & \multicolumn{1}{c|}{Acc} & \multicolumn{1}{c|}{M-F1} & \multicolumn{1}{c|}{M-P} & \multicolumn{1}{c|}{M-R} & \multicolumn{1}{c|}{Acc} & \multicolumn{1}{c|}{M-F1} & \multicolumn{1}{c|}{M-P} & \multicolumn{1}{c|}{M-R} & \multicolumn{1}{c|}{Acc} & \multicolumn{1}{c|}{M-F1} & \multicolumn{1}{c|}{M-P} & \multicolumn{1}{c|}{M-R} & \multicolumn{1}{c|}{Acc} & \multicolumn{1}{c|}{M-F1} & \multicolumn{1}{c|}{M-P} & \multicolumn{1}{c|}{M-R} \\ \cline{2-17}
\begin{tabular}[c]{@{}c@{}}\{'Negative': 0.2779, \\ 'Positive': 0.2601, \\ 'Neutral': 0.4619\}\end{tabular} & 89.86 & 78.63 & 78.59 & 78.67 & 89.00 & 76.91 & 77.03 & 76.82 & 0.0058 & 0.0295 & 0.0227 & 0.0366 & 0.80 & 0.32 & 0.87 & 0.35 \\ \hline

\end{tabular}
\end{adjustbox}
\end{table*}

% \textbf{Note:} The table presents performance metrics before and after applying focal loss. Metrics include accuracy (Acc), macro-F1 (M-F1), precision (P), recall (R) for ASC, and micro-F1 (F1), precision (P), recall (R) for OTE. P-values indicate the statistical significance of changes, highlighting the effectiveness of focal loss in improving domain-specific model performance. The highest result among all models is \textbf{bold}, while the highest result for each metric is \underline{underlined}. Significant p-values (<0.05) are dashed underlined.

\subsection{Benchmarking Against Existing Models}
\label{subsec:ComparingModels}
This section provides a summary of deep-learning models for \gls{asc} and \gls{ote} extraction evaluated on the Arabic \gls{semeval}-2016 hotel review dataset. It also includes summary tables comparing our models in different settings with other deep learning models tested on the same dataset for the extraction tasks \gls{asc} and \gls{ote}.

\subsubsection{\textbf{Aspect-Sentiment Classification}}
\label{subsec:ComparingASC}
\textbf{INSIGHT-1~\citep{ruder-etal-2016-insight-1_S12}}: employed a convolutional neural network (\gls{cnn}) that takes the concatenation of both aspect category and sentence word embedding as input.

\textbf{H-\gls{lstm}~\citep{ruder-etal-2016-hierarchical_S14}}: employed a multi-layer \gls{bi-lstm} for sentence and review-level representation, taking the aspect-category and sentence word embeddings as input.

\textbf{\gls{rnn}~\citep{ALSMADI2018_S23}}: utilized an \gls{rnn} that takes word embedding as input, along with handcrafted features such as N-grams, \gls{pos} tags, \gls{ner}, and \gls{tfidf}.

\textbf{AB-LSTM-PC~\citep{Al-Smadi2019c_S30}}: the authors utilized a \gls{bi-lstm} model with an attention mechanism to create a weighted representation of input words relative to aspect terms.

\textbf{\gls{ian}-\gls{bi-gru}~\citep{Al-Dabet_2021_S48}}: the researchers used an interactive attention network (\gls{ian}) with \gls{bi-gru}. The \gls{ian} interaction enables the generation of aspects and context representations that are learned interactively from each other.

\textbf{MBRA~\citep{Abdelgwad2021_S47}}: the researchers employed a positioned-weighted memory network consisting of multiple layers of \gls{indylstm} and recurrent attention mechanisms.

\textbf{AraBERT-linear-pair~\citep{Abdelgwad2022}}: The researchers utilized the AraBERT model with a task-specific linear layer that takes as input the text and explicit opinion target terms (\gls{ote}).

\textbf{Dialect Normalization + Arabic-\gls{bert}~\citep{Chennafi2022}}: in this research, the authors used an encoder--decoder sequence-to-sequence model to convert Egyptian dialect reviews to Modern Standard Arabic (\gls{msa}). The converted text was then passed along with the opinion target term into Arabic \gls{bert}, followed by a linear classification layer.

Table \ref{tab:comparedModelsASC} shows the comparison of our models in terms of accuracy to different \gls{asc} deep learning models on the \gls{semeval}-2016 Arabic hotel reviews dataset.

\begin{table}[htp!]
\centering
\caption{Performance comparison of Aspect-Sentiment classification models on the SemEval-2016 Arabic Hotel Review Dataset. The highest accuracy is highlighted in \textbf{bold}.}
\label{tab:comparedModelsASC}

\resizebox{\columnwidth}{!}{%
\begin{tabular}{|l|c|}
\hline
\textbf{Reference} & \textbf{Accuracy (\%)} \\ \hline
INSIGHT-1~\cite{ruder-etal-2016-insight-1_S12} & 82.72 \\ \hline
H-LSTM~\cite{ruder-etal-2016-hierarchical_S14} & 82.90 \\ \hline
RNN~\cite{ALSMADI2018_S23} & 87.00 \\ \hline
AB-LSTM-PC~\cite{Al-Smadi2019c_S30} & 82.60 \\ \hline
IAN-BGRU~\cite{Al-Dabet_2021_S48} & 87.31 \\ \hline
MBRA~\cite{Abdelgwad2021_S47} & 83.98 \\ \hline
AraBERT-linear-pair~\cite{Abdelgwad2022} & 89.51 \\ \hline
Dialect Normalization+Arabic-BERT~\cite{Chennafi2022} & 84.65 \\ \hline
\textbf{Ours (In-DAPT CAMeLBERT-MSA Full fine-tuning)} & 89.67 \\ \hline
Ours (In-DAPT CAMeLBERT-MSA Parallel Adapters Fusion) & 88.06 \\ \hline
\textbf{Ours (In-DAPT CAMeLBERT-MSA Full fine-tuning + Focal Loss)} & \textbf{89.86} \\ \hline
\end{tabular}%
}
\end{table}

\subsubsection{\textbf{Opinion-Target Expression Extraction}}
\textbf{\gls{rnn}-Softmax~\citep{ALSMADI2018_S23}}: utilized an \gls{rnn} that takes word embedding as input, along with handcrafted features such as N-grams, \gls{pos} tags, \gls{ner}, and \gls{tfidf}.

\textbf{\gls{bi-lstm}-\gls{crf}~\citep{Al-Smadi2019c_S30}}: the researchers utilized \gls{bi-lstm} followed by \gls{crf}, which used fastText character-level word representation.

\textbf{\gls{cnn}-\gls{bi-gru}-\gls{crf}~\citep{Al-Dabet2020a_S41}}: the author used \gls{cnn} to extract character representation, combined with word embeddings and input into the \gls{bi-gru} layer, followed by \gls{crf}.

\textbf{ABNM~\citep{Abdelgwad2021_S47}}: the researchers used an encoder--encoder with an attention mechanism between them followed by \gls{crf}.

\textbf{BF-BiLSTM-\gls{crf}~\citep{Fadel2022}}: the authors concatenated Arabic Flair string embedding and AraBERT embedding, which fed into \gls{bi-lstm} layers followed by \gls{crf}.

Table \ref{tab:ComparedModelsOTE} shows the comparison of our models in terms of micro-F1 to different \gls{ote} extraction deep learning models on the \gls{semeval}-2016 Arabic hotel reviews dataset.

\begin{table}[htp!]
\centering
\caption{Performance comparison of Opinion Target Expression Extraction models on the SemEval-2016 Arabic Hotel Review Dataset. The highest F1 score is highlighted in \textbf{bold}.}
\label{tab:ComparedModelsOTE}

\resizebox{\columnwidth}{!}{%
\begin{tabular}{|l|c|}
\hline
\textbf{Reference} & \textbf{F1 Score (\%)} \\ \hline
RNN-Softmax~\cite{ALSMADI2018_S23} & 48.00 \\ \hline
Bi-LSTM-CRF~\cite{Al-Smadi2019c_S30} & 69.98 \\ \hline
CNN-Bi-GRU-CRF~\cite{Al-Dabet2020a_S41} & 72.83 \\ \hline
ABNM~\cite{Abdelgwad2021_S47} & 70.67 \\ \hline
\textbf{BF-BiLSTM-CRF~\cite{Fadel2022}} & \textbf{79.70} \\ \hline
Ours (In-DAPT CAMeLBERT-MSA Full fine-tuning) & 77.02 \\ \hline
Ours (In-DAPT CAMeLBERT-MSA Lang+Task Stacked Adapter) & 76.31 \\ \hline
Ours (In-DAPT CAMeLBERT-MSA Full fine-tuning + Focal Loss) & 76.42 \\ \hline
\end{tabular}%
}
\end{table}

    \section{Discussion}
\label{sec:discussion}

\subsection{Overview of Findings and Fine-tuning Strategy}
Our study examined how domain-adaptive and task-adaptive pre-training affect Arabic \gls{absa} tasks, specifically \gls{asc} and \gls{ote} extraction. Using Arabic hotel reviews, we tested several pre-trained Arabic \gls{bert} models and fine-tuning strategies. The results showed that CAMeLBERT-MSA, QARiB, and MARBERTv2 outperformed other Arabic BERT models. Several factors may explain their strong performance. For example, CAMeLBERT-MSA is trained only on \gls{msa} data, which enhances its ability to represent this dialect. This is especially relevant since \gls{msa} makes up 65\% of the \gls{semeval}-2016 Arabic hotel reviews dataset (see Figure \ref{fig:DialectsCount}). On the other hand, QARiB and MARBERTv2 benefit from more vast and mixed datasets, including various dialects, providing a more comprehensive linguistic exposure. Their performance is further improved by undergoing a larger number of training iterations and using larger vocabulary dictionaries (see Table \ref{tab:ARBERT}). 

Notably, the fully fine-tuned Arabic \gls{bert} models outperformed those that used \gls{bert} as a feature extractor with all layers frozen. Thus, we conducted our subsequent experiments using full fine-tuning and adapters. These results suggest that pre-trained model parameters updates are necessary for these models to accurately comprehend \gls{absa} tasks, which aligns with the findings of \cite{Abdelgwad2022} and \cite{Fadel2022}. 

\subsection{Effects of Domain- and Task-Adaptive Pre-training}
The results show that for \gls{asc}, all three Arabic \gls{bert} models with \gls{indapt} outperform \gls{tapt}, \gls{outdapt}, and No-\gls{dapt} in both accuracy and macro-F1. In \gls{ote} extraction, \gls{indapt}+\gls{tapt} achieves the best performance for QARiB and MARBERTv2. Notably, across all adaptation settings, \gls{indapt} CAMeLBERT-MSA outperformed both QARiB and MARBERTv2 in both tasks, demonstrating its effectiveness in domain adaptation.

The strong performance of CAMeLBERT-\gls{msa} might be linked to two factors. First, CAMeLBERT-\gls{msa} is pre-trained on \gls{msa}, which constitutes the largest portion of \gls{semeval}-2016 and \gls{hard} datasets (see Figure \ref{fig:DialectsCount}). Second, CAMeLBERT-\gls{msa} has the fewest learnable parameters among the tested Arabic \gls{bert} pre-trained models, which makes it less prone to overfitting (see Table \ref{tab:ARBERT}).

CAMeLBERT-MSA with \gls{indapt} achieved the highest improvement over the baseline No-\gls{dapt}, with an accuracy increase of 0.69\% and a Macro-F1 increase of 6.44\% in \gls{asc}, and a Micro-F1 increase of 0.67\% in \gls{ote}. This improvement may be attributed to CAMeLBERT-MSA’s low vocabulary overlap (7\%–9\%) with the HARD and SemEval-2016 Arabic hotel review datasets (Figure \ref{fig:OverlapDSMODEL}). This observation aligns with the heuristic proposed by \cite{gururangan-etal-2020-dont}, suggesting that lower vocabulary overlap allows models to benefit more from domain adaptation by learning essential, previously unseen domain-specific vocabulary.

\subsection{Out-of-Domain TAPT/OutDAPT and Domain Relevance}
The impact of adaptive pre-training with \gls{outdapt} and \gls{outdapt}+\gls{tapt} corpora significantly reduced the performance of CAMeLBERT-MSA, QARiB, and MARBERTv2 across both tasks compared to baseline No-\gls{dapt}. This aligns with the findings of~\cite{gururangan-etal-2020-dont}, highlighting the effectiveness of using in-domain data for adaptive pretraining. Unlike the findings of Gururangan et al., where \gls{tapt} improved performance in the review-level sentiment analysis, our results indicated a decline in the \gls{asc} task. This decrease may be attributed to data imbalance or the limited size of the \gls{tapt} corpus, as evidenced by a more significant drop in macro-F1 compared to accuracy (see Figure \ref{fig:ASCdomain_adaptation_graph}). In contrast, for \gls{ote} extraction, \gls{tapt} improved performance for all models except MARBERTv2, which experienced only a slight reduction. This suggests that \gls{ote} extraction is more sensitive to domain relevance than sentiment classification, aligning with the observations by Xu et al. \cite{Xu2019,Xu2020c}. 

\subsection{Corpus Size vs.\ Domain Relevance (Ablation)}
\begin{figure}[t]
\centering
\includegraphics[width=\linewidth]{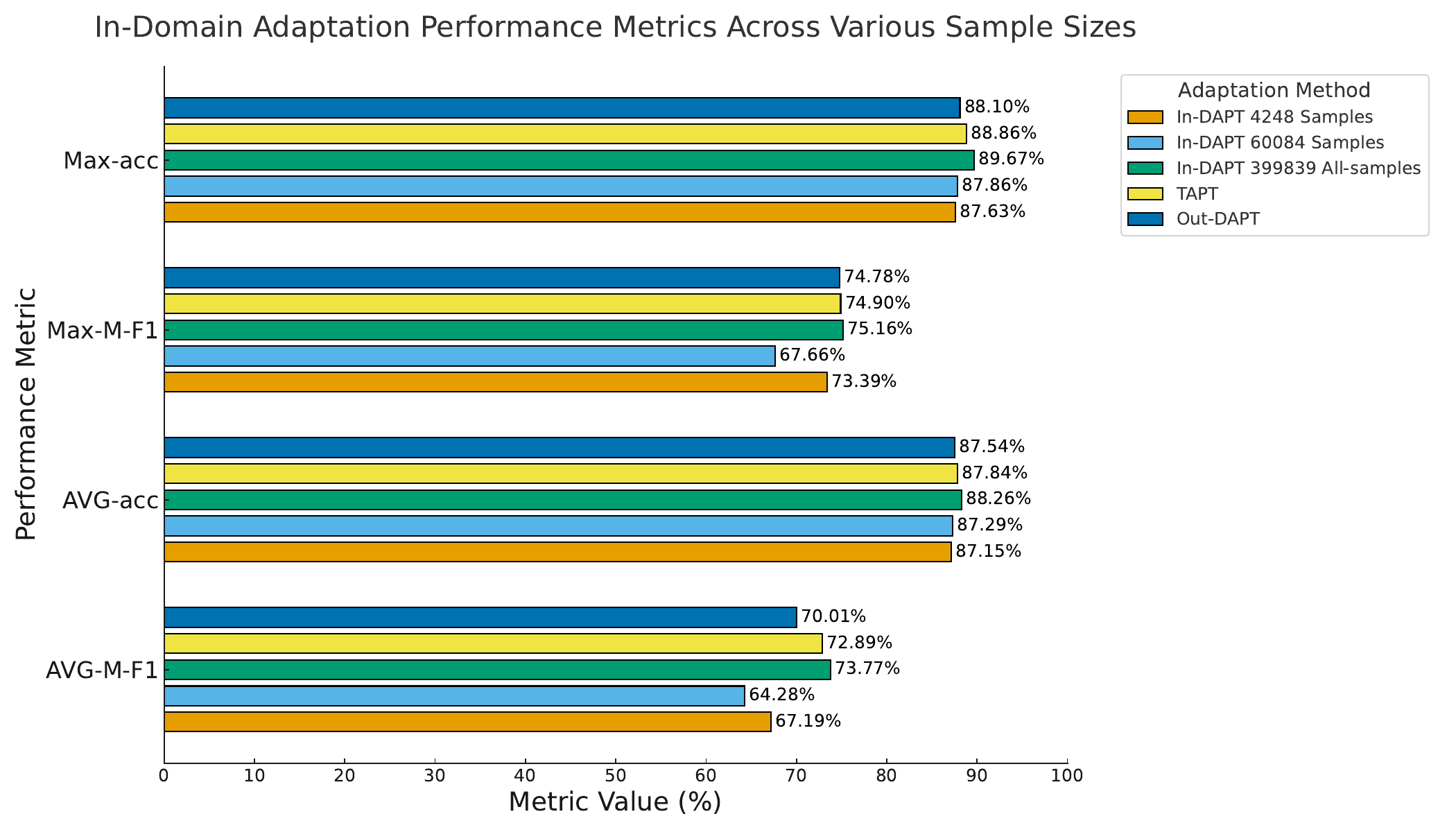}
\caption{Comparison of average and maximum accuracy, and macro F1 scores for aspect-sentiment classification across in-domain models (\gls{indapt} 4248, 60084, 399839 samples), \gls{tapt}, and \gls{outdapt}.}
\label{fig:ASC_In_domain_adaptation_SamplesVsPerformance_graph}
\end{figure}

To determine whether the superior performance of \gls{indapt} was due to domain relevance rather than corpus size, we performed adaptive pre-training on subsamples of the \gls{indapt} corpus, matching the sizes of the \gls{tapt} and \gls{outdapt} datasets (4,248 and 60,084 samples, respectively). At the same sample size, \gls{asc}  (see Figure~\ref{fig:ASC_In_domain_adaptation_SamplesVsPerformance_graph}), both \gls{tapt} and \gls{outdapt} outperformed \gls{indapt} . This suggests that even out-of-domain adaptive pretraining can be beneficial due to cross-domain sentiment similarities. However, for \gls{ote} extraction, only \gls{tapt} outperformed \gls{indapt}, while \gls{outdapt} underperformed \gls{indapt}, indicating that domain relevance is crucial, as \gls{ote} terms vary by domain \cite{Xu2019,Xu2020c}. Notably, the largest \gls{indapt} corpus, with 399,839 samples, achieved the best overall performance, suggesting that while domain relevance is important—especially for domain-sensitive tasks like \gls{ote} extraction—, increasing corpus size can also improve performance in both tasks.

% --- Place ablation figures at first mention ---

\begin{figure}[t]
\centering
\includegraphics[width=\linewidth]{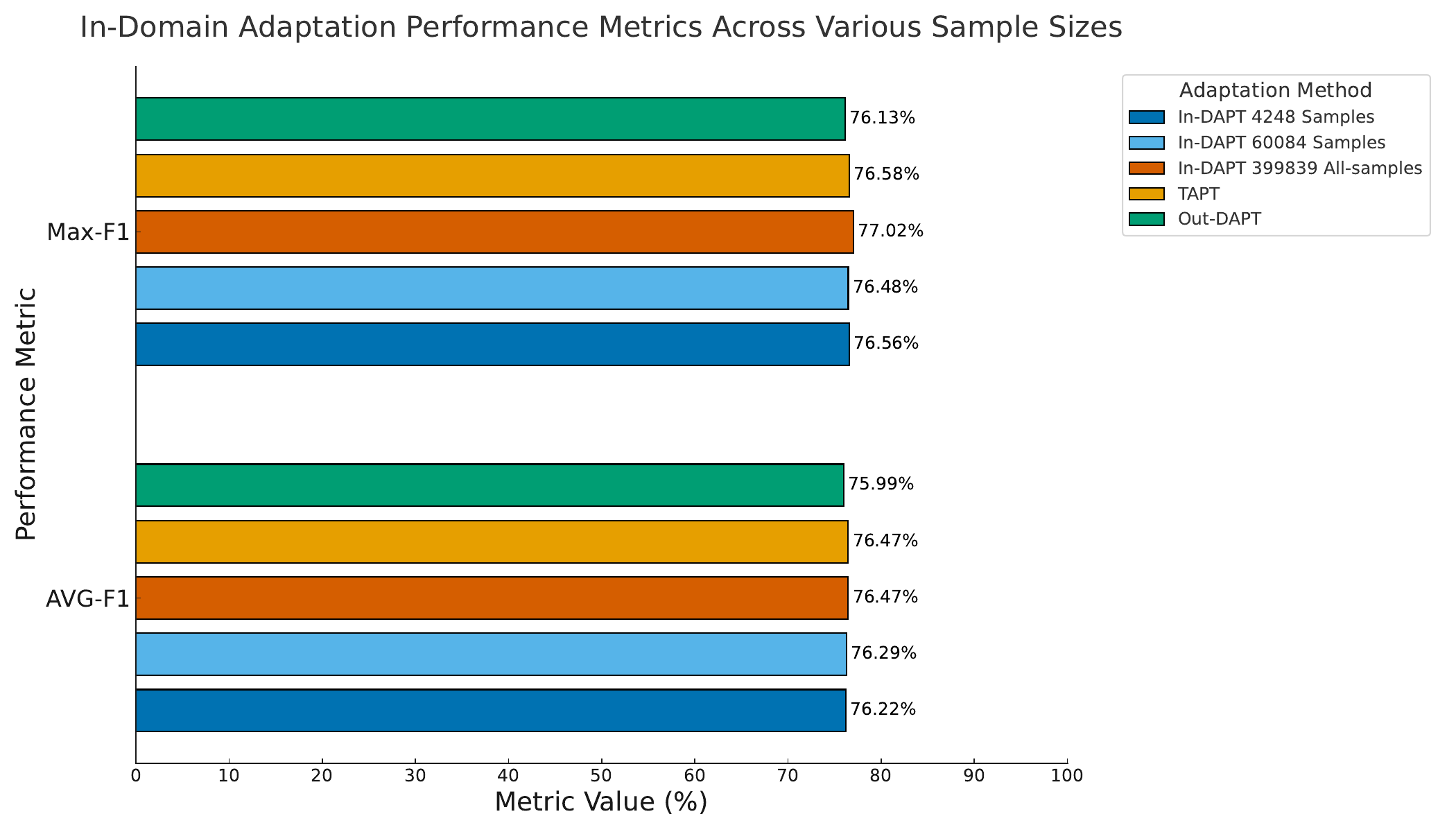}
\caption{Comparison of average and maximum micro F1 scores for opinion target extraction across in-domain models (\gls{indapt} 4248, 60084, 399839 samples), \gls{tapt}, and \gls{outdapt}.}
\label{fig:OTE_In_domain_adaptation_SamplesVsPerformance_graph}
\end{figure}
\FloatBarrier

\subsection{Statistical Significance and Training Choices}
Our results show that \gls{indapt} techniques led to performance improvements in both \gls{asc} and \gls{ote} extraction, but these gains were not statistically significant (p-values: 0.18–0.54 for accuracy, 0.07–0.81 for macro-F1 in \gls{asc}; 0.06–0.55 for micro-F1 in \gls{ote}). This is likely due to the small, imbalanced end-task dataset, which limits model generalization. To address this, we employed adapter-based fine-tuning, class weighting, and focal loss. For \gls{ote}, we also experimented with CRF, BiLSTM, and BiGRU layers in the CAMeLBERT-MSA model. Although these configurations did not improve overall F1, the BiGRU+Softmax setup increased recall to 79.85\%, improving span detection at the expense of precision.

\subsection{Adapters, Fusion, and Loss Functions}
For \gls{ote} extraction, \gls{indapt} significantly improved performance, while most \gls{tapt} adapters had a negative impact, contradicting the CAMeLBERT-MSA \gls{tapt}'s slight performance gain.

Figure~\ref{fig:ASC_Adapter_graph} shows that combining \gls{indapt} or \gls{tapt} language adapters with task adapters results in lower accuracy on the \gls{asc} task compared to using task adapters alone (No-\gls{dapt}). However, most models achieved higher macro-F1 scores, except for \gls{indapt} with the Pfeiffer adapter, suggesting better performance across all classes. This indicates that domain-adapted models may improve the balance of classes.

In the \gls{ote} task (Figure~\ref{fig:OTE_Adapter_graph}), \gls{indapt} adapters improved performance, likely because they helped the model better understand domain-specific expressions. However, most \gls{tapt} adapters led to lower performance than No-\gls{dapt}. These findings partly contrast with full fine-tuned \gls{tapt} CAMeLBERT-MSA’s , which showed a slight improvement (Figure~\ref{fig:OTEdomain_adaptation_graph}).

Our findings show that adapter fusion outperforms single-task adapters in the \gls{asc} task; however, it decreased performance in the \gls{ote} extraction task (see Figure \ref{fig:Adapers_single_vs_fusion_graph}). A possible reason for this could be that \gls{asc} benefits from shared representations that utilize broader contextual knowledge, improving the accuracy of the classification. In contrast, OTE tasks depend on task-specific features that can be negatively affected by generalized representations and feature dilution introduced through multi-task learning, as noted in previous research \citep{YANG2021344,Varia2022}.

Integrating focal loss and class weighting improved sensitivity to minority classes, particularly improving recall for both \gls{asc} and \gls{ote} extraction tasks, achieving the highest accuracy of 89.86\% on the \gls{asc} task (see Figure \ref{fig:Adapers_CE_vs_FL_graph}). Although precision slightly decreased, this deliberate trade-off prioritized recall is crucial for imbalanced datasets. These results demonstrate the efficacy of advanced loss functions and weighting techniques in handling skewed class distributions.

\subsection{Comparison to Prior Work}
Our two top-performing \gls{asc} models, based on \gls{indapt} CAMeLBERT-MSA and optimized with focal and cross-entropy losses, outperform previous deep learning approaches on the \gls{semeval}-2016 Arabic Hotel Reviews dataset. AraBERT-linear-pair by \cite{Abdelgwad2022} follows closely with an accuracy of 89.51\%, which is 0.35\% and 0.16\% lower than our models. Unlike their approach, which uses explicit OTE-review pairs as BERT inputs and excludes implicit aspects, our models leverage aspect-category–review pairs. This enables us to address both explicit and implicit aspects, which are commonly found in user reviews.
Unlike many previous models, our approach does not add additional layers or components beyond the base transformer. We fine-tune the pretrained model directly, without applying complex architectures such as memory networks~\citep{Abdelgwad2021_S47} or attention-based interaction modules~\citep{Al-Dabet_2021_S48}. Even with this simpler setup, our models outperform MBRA~\citep{Abdelgwad2021_S47}, IAN-BiGRU~\citep{Al-Dabet_2021_S48}, and earlier CNN-, RNN-, and BiLSTM-based models~\citep{ruder-etal-2016-insight-1_S12, ruder-etal-2016-hierarchical_S14, ALSMADI2018_S23, Al-Smadi2019c_S30, Chennafi2022}. 

For the \gls{ote} extraction task, our best model based on full fine-tuning of \gls{indapt} CAMeLBERT-MSA ranks second to the BF-BiLSTM-CRF model \citep{Fadel2022}, while our adapter-based variants still achieve higher performance than models such as CNN-BiGRU-CRF \citep{Al-Dabet2020a_S41}, ABNM \citep{Abdelgwad2021_S47}, BiLSTM-CRF \citep{Al-Smadi2019c_S30}, and RNN-Softmax \citep{ALSMADI2018_S23}. These results highlight the importance of contextualized embeddings adapted to the domain and balanced loss functions for enhanced performance in Arabic \gls{absa}.
\subsection{Error Analysis}
\label{subsec:error-analysis}

To investigate the limited performance of our model, we analyzed predictions from the best-performing \gls{indapt} CAMeLBERT, focusing on cases where predictions did not match the correct labels in the ASC and OTE tasks. We then analyzed the 50 samples with the highest loss, as computed by the model’s loss function, to identify where the model struggled and why.

We used the \textit{Transformers Interpret} library\footnote{\url{https://github.com/cdpierse/transformers-interpret}}—an explainability tool for Hugging Face Transformers built on PyTorch’s Captum—and applied Integrated Gradients (IG) to identify which words most influenced misclassifications. We chose IG because it is based on well-established theoretical principles: \textit{Sensitivity}, which ensures that important input features receive non-zero attribution, and \textit{Implementation Invariance}, which guarantees consistent attributions across models with identical behavior, regardless of internal differences \citep{Sundararajan2017}.

\subsubsection{Aspect Sentiment Classification (ASC) (see Table~\ref{tab:ASC_ErrorAnalysis})}
The \gls{asc} model showed recurring errors. First, annotation mistakes: explicit sentiment often conflicted with the assigned labels. For example, negative terms like ``\RL {قديمة}'' (old) and ``\RL {وبالية}'' (dilapidated) in \textbf{Ex1} were labeled as neutral; positive expressions such as ``\RL {رائعه}'' (wonderful) in \textbf{Ex7} and ``\RL {مؤدبين}'' (polite) in \textbf{Ex10} were also tagged as neutral; and strongly negative words like ``\RL {سيئة}'' (bad) in \textbf{Ex12} and ``\RL {أخطاء}'' (mistakes) in \textbf{Ex13} were incorrectly marked as positive. Second, the model missed contrastive shifts signaled by ``\RL {لكن}'' (but) and ``\RL {ولكن}'' (however), leading to errors in mixed sentiment: in \textbf{Ex11}, it ignored the positive ``\RL {فطور جيد}'' (good breakfast) after negative cues like ``\RL {سيئة... باهظا}'' (bad... expensive), and in \textbf{Ex7} and \textbf{Ex6} it overlooked key contrasts in service and location. Third, a positivity bias gave excess weight to early positive terms (e.g., ``\RL {رائع}'' in \textbf{Ex2}, ``\RL {أجمل}'' in \textbf{Ex6}) while downplaying later negatives such as ``\RL {بطيء}'' (slow) in \textbf{Ex9}. Fourth, the model misattributed sentiment across aspects: general hotel praise in \textbf{Ex2} overrode negative service feedback, and scenic location terms in \textbf{Ex6} obscured access issues. Fifth, when conflicting opinions appeared within the same aspect, it defaulted to the dominant tone: \textbf{Ex8} prioritized ``\RL {مزعج}'' (annoying) over ``\RL {نظيفة}'' (clean), \textbf{Ex11} effectively averaged opposing cues, and \textbf{Ex7} favored positive spatial descriptions over slow service. Finally, subword tokenization introduced noise: partial tokens such as ``\#\#\RL {ار}'' in \textbf{Ex5} and ``\#\#\RL {لسفة}'' in \textbf{Ex4} led to context-insensitive interpretations.

\subsubsection{Opinion Target Expression (OTE) Extraction (see Table~\ref{tab:OTE_error_analysis})}
Errors in the \gls{ote} extraction model mainly stem from shallow token-level heuristics rather than syntactic and semantic analysis. The model often confused roles, misidentifying subjects or locations as targets: in \textbf{Ex1} it tagged ``\RL {فريق}'' (team) instead of the true target ``\RL {الاستقبال}'' (reception), and in \textbf{Ex2} and \textbf{Ex3} it treated ``\RL {الحرم}'' (the sanctuary) and ``\RL {محطة الترام}'' (tram station) as targets. It also over-relied on the definite article heuristic (``\RL {ال}'', “the”), producing false positives such as ``\RL {البلكونة}'' (the balcony) in \textbf{Ex6}. Multi-word expressions—especially Idafa constructions—were not handled well, as with ``\RL {مستوى الاكل}'' (food quality) in \textbf{Ex4} and ``\RL {بار الماء الساخن}'' (hot water bar) in \textbf{Ex5}. Morphologically rich words were over-segmented, e.g., ``\RL {قذارة}'' (filthiness) in \textbf{Ex7} was split into meaningless sub-tokens. Finally, annotation inconsistencies added noise: ``\RL {حمامات}'' (bathrooms) was labeled a non-target in \textbf{Ex8}, while the conjunction ``\RL {و}'' (and) was incorrectly included as part of a target in \textbf{Ex9}.

% --- Place error-analysis tables at first mention ---
% --- helpers (place right before the table; no preamble edits needed) ---
% Make every \colorbox breakable (your earlier trick)

\begin{table*}[b!]
  \centering
\caption{Selected \gls{asc} examples from the 50 highest-loss cases, listing gold label (True), model prediction+probability (Predicted), attribution polarity (Attr.), IG score (Score), and token-level \emph{Word Importance}. Highlights are Integrated Gradients attributions (Transformers Interpret/Captum): green = positive, red = negative; darker shade = stronger contribution.}
% RTL helpers for Arabic tokens (pdfLaTeX + arabtex already in preamble)
\TeXXeTstate=1
\newcommand{\AR}[1]{\RL{#1}} % shape Arabic with arabtex
\newcommand{\ARCB}[2]{% color each token LTR to avoid overpainting
  \begingroup\beginL\colorbox[HTML]{#1}{\AR{#2}}\endL\endgroup
}
\newcommand{\rtlseq}[1]{{\beginR #1\endR}} % lay out a whole run RTL

  \begin{adjustbox}{max width=\textwidth}
    \scriptsize
    \setlength{\tabcolsep}{2.5pt}
    \setlength{\fboxsep}{0.5pt}
    \renewcommand{\arraystretch}{1.05}
    \begin{tabular*}{\textwidth}{@{\extracolsep{\fill}}c c c c c p{0.50\textwidth}@{}}
      \toprule
      \textbf{Ex.} & \textbf{True} & \textbf{Predicted} & \textbf{Attr.} & \textbf{Score} & \textbf{Word Importance} \\
      \midrule

      1 & neutral & negative (0.93) & negative & 1.25 &
      \colorbox[HTML]{FFFFFF}{[CLS]}%
      \rtlseq{%
        \ARCB{F8D2D2}{جودة}%
        \ARCB{FADFDF}{الغرفة}%
        \ARCB{FBE4E4}{غرفة}%
        \ARCB{DFFADF}{من}%
        \ARCB{FEFAFA}{الدرجة}%
        \ARCB{D6F9D6}{الثانية}%
        \ARCB{E4FBE4}{.}%
        \ARCB{FDF6F6}{غرفة}%
        \ARCB{DBF9DB}{قديمة}%
        \ARCB{B7F4B7}{جدا}%
        \ARCB{C0F6C0}{جدا}%
        \ARCB{79EB79}{وبال}%
        \ARCB{C4F6C4}{\#\#ية}%
        \ARCB{C4F6C4}{.}%
      }%
      \colorbox[HTML]{FFFFFF}{[SEP]}%
      \colorbox[HTML]{FEFAFA}{r}%
      \colorbox[HTML]{FDF6F6}{\#\#oo}%
      \colorbox[HTML]{FEFAFA}{\#\#ms}%
      \colorbox[HTML]{FBE8E8}{qu}%
      \colorbox[HTML]{FDF1F1}{\#\#al}%
      \colorbox[HTML]{FCEDED}{\#\#ity}%
      \colorbox[HTML]{FFFFFF}{[SEP]} \\

      \midrule
      2 & negative & positive (0.77) & positive & 1.27 &
      \colorbox[HTML]{FFFFFF}{[CLS]}%
      \rtlseq{%
        \ARCB{FFFFFF}{هيل}%
        \ARCB{FFFFFF}{\#\#تو}%
        \ARCB{FFFFFF}{\#\#ن}%
        \ARCB{E4FBE4}{دريم}%
        \ARCB{EDFCED}{فندق}%
        \ARCB{C0F6C0}{رائع}%
        \ARCB{FFFFFF}{و}%
        \ARCB{F6FDF6}{مكانة}%
        \ARCB{F1FDF1}{قريب}%
        \ARCB{FAFEFA}{من}%
        \ARCB{FAFEFA}{خليج}%
        \ARCB{F6FDF6}{نعمة}%
        \ARCB{FFFFFF}{و}%
        \ARCB{F6FDF6}{خدمة}%
        \ARCB{F1FDF1}{الفندق}%
        \ARCB{FFFFFF}{اكثر}%
        \ARCB{FFFFFF}{من}%
        \ARCB{AAF2AA}{رائعة}%
        \ARCB{FEFAFA}{خدمة}%
        \ARCB{FADFDF}{الغرف}%
        \ARCB{F09C9C}{بط}%
        \ARCB{F6C4C4}{\#\#يئة}%
        \ARCB{FADFDF}{و}%
        \ARCB{FFFFFF}{لكنها}%
        \ARCB{FFFFFF}{مست}%
        \ARCB{FDF6F6}{\#\#جيب}%
        \ARCB{F6FDF6}{\#\#ة}%
        \ARCB{FFFFFF}{ ...}%
      }%
      \colorbox[HTML]{FFFFFF}{[SEP]}%
      \colorbox[HTML]{F6FDF6}{s}%
      \colorbox[HTML]{FDF6F6}{\#\#erv}%
      \colorbox[HTML]{DFFADF}{\#\#ice}%
      \colorbox[HTML]{E4FBE4}{g}%
      \colorbox[HTML]{FFFFFF}{\#\#eneral}%
      \colorbox[HTML]{FFFFFF}{[SEP]} \\

      \midrule
      3 & negative & positive (0.98) & positive & 1.45 &
      \colorbox[HTML]{FFFFFF}{[CLS]}%
      \rtlseq{%
        \ARCB{C9F7C9}{الخدمة}%
        \ARCB{F7C9C9}{تبدو}%
        \ARCB{F6C0C0}{بط}%
        \ARCB{FBE4E4}{\#\#يئة}%
        \ARCB{FFFFFF}{ولكن}%
        \ARCB{FEFAFA}{المكان}%
        \ARCB{FBE4E4}{شا}%
        \ARCB{FEFAFA}{\#\#سع}%
        \ARCB{FBE4E4}{والس}%
        \ARCB{FFFFFF}{\#\#قا}%
        \ARCB{FDF6F6}{\#\#ة}%
        \ARCB{E4FBE4}{دائما}%
        \ARCB{CDF7CD}{على}%
        \ARCB{AAF2AA}{استعداد}%
        \ARCB{CDF7CD}{لتلبية}%
        \ARCB{EDFCED}{احتياجات}%
        \ARCB{E8FBE8}{\#\#نا}%
        \ARCB{FFFFFF}{.}%
        \ARCB{FFFFFF}{إذا}%
        \ARCB{FAFEFA}{سألت}%
        \ARCB{FFFFFF}{\#\#مون}%
        \ARCB{FFFFFF}{\#\#ي}%
        \ARCB{FAFEFA}{عما}%
        \ARCB{FFFFFF}{إذا}%
        \ARCB{FAFEFA}{كنت}%
        \ARCB{FFFFFF}{سأ}%
        \ARCB{FFFFFF}{\#\#عود}%
        \ARCB{FFFFFF}{سأ}%
        \ARCB{FFFFFF}{\#\#جيب}%
        \ARCB{FAFEFA}{\#\#كم}%
        \ARCB{FAFEFA}{بالطبع}%
        \ARCB{FEFAFA}{،}%
        \ARCB{FFFFFF}{نعم}%
        \ARCB{FAFEFA}{.}%
      }%
      \colorbox[HTML]{FFFFFF}{[SEP]}%
      \colorbox[HTML]{DFFADF}{s}%
      \colorbox[HTML]{C4F6C4}{\#\#erv}%
      \colorbox[HTML]{AEF3AE}{\#\#ice}%
      \colorbox[HTML]{D2F8D2}{g}%
      \colorbox[HTML]{D2F8D2}{\#\#eneral}%
      \colorbox[HTML]{FFFFFF}{[SEP]} \\

      \midrule
      4 & negative & positive (0.96) & positive & 1.94 &
      \colorbox[HTML]{FFFFFF}{[CLS]}%
      \rtlseq{%
        \ARCB{FDF1F1}{والإ}%
        \ARCB{8FEF8F}{\#\#فط}%
        \ARCB{FAFEFA}{\#\#ار}%
        \ARCB{D2F8D2}{متوسط}%
        \ARCB{F1FDF1}{على}%
        \ARCB{6BE96B}{أفضل}%
        \ARCB{F9DBDB}{تقدير}%
        \ARCB{FFFFFF}{،}%
        \ARCB{FEFAFA}{وهو}%
        \ARCB{FBE8E8}{فندق}%
        \ARCB{D6F9D6}{تجاري}%
        \ARCB{EDFCED}{يعمل}%
        \ARCB{FFFFFF}{بف}%
        \ARCB{D2F8D2}{\#\#لسفة}%
        \ARCB{D2F8D2}{الربح}%
        \ARCB{F6FDF6}{أولا}%
        \ARCB{FEFAFA}{والعم}%
        \ARCB{FFFFFF}{\#\#يل}%
        \ARCB{FCEDED}{يأتي}%
        \ARCB{FEFAFA}{بعد}%
        \ARCB{EDFCED}{ذلك}%
        \ARCB{FEFAFA}{.}%
      }%
      \colorbox[HTML]{FFFFFF}{[SEP]}%
      \colorbox[HTML]{FAFEFA}{h}\colorbox[HTML]{FAFEFA}{\#\#ote}\colorbox[HTML]{EDFCED}{\#\#l}\colorbox[HTML]{F6FDF6}{price}\colorbox[HTML]{E8FBE8}{\#\#s}\colorbox[HTML]{FFFFFF}{[SEP]} \\

      \midrule
      5 & neutral & positive (0.97) & positive & 0.58 &
      \colorbox[HTML]{FFFFFF}{[CLS]}%
      \rtlseq{%
        \ARCB{F1FDF1}{كما}%
        \ARCB{EDFCED}{أن}%
        \ARCB{F6FDF6}{حمام}%
        \ARCB{8AEE8A}{السباحة}%
        \ARCB{F8D2D2}{والإ}%
        \ARCB{FEFAFA}{\#\#فط}%
        \ARCB{9CF09C}{\#\#ار}%
        \ARCB{F2A5A5}{مقبول}%
        \ARCB{C9F7C9}{\#\#ان}%
        \ARCB{F7CDCD}{.}%
      }%
      \colorbox[HTML]{FFFFFF}{[SEP]}%
      \colorbox[HTML]{F6FDF6}{f}\colorbox[HTML]{FFFFFF}{\#\#ac}\colorbox[HTML]{FFFFFF}{\#\#il}\colorbox[HTML]{FEFAFA}{\#\#it}\colorbox[HTML]{FDF6F6}{\#\#ies}\colorbox[HTML]{F1FDF1}{g}\colorbox[HTML]{DFFADF}{\#\#eneral}\colorbox[HTML]{FFFFFF}{[SEP]} \\

      \midrule
      6 & negative & positive (0.97) & positive & 2.09 &
      \colorbox[HTML]{FFFFFF}{[CLS]}%
      \rtlseq{%
        \ARCB{B7F4B7}{جيد}%
        \ARCB{4CE54C}{افضل}%
        \ARCB{E4FBE4}{منش}%
        \ARCB{FFFFFF}{\#\#أة}%
        \ARCB{F1FDF1}{اخر}%
        \ARCB{F8D2D2}{\#\#ئ}%
        \ARCB{FDF6F6}{لقد}%
        \ARCB{FFFFFF}{قمت}%
        \ARCB{FAFEFA}{بزيارة}%
        \ARCB{EDFCED}{جزر}%
        \ARCB{FAFEFA}{ومنت}%
        \ARCB{FFFFFF}{\#\#جع}%
        \ARCB{FFFFFF}{\#\#ات}%
        \ARCB{F1FDF1}{اخر}%
        \ARCB{F9DBDB}{\#\#ئ}%
        \ARCB{FFFFFF}{في}%
        \ARCB{F1FDF1}{جزر}%
        \ARCB{EDFCED}{المالديف}%
        \ARCB{F1FDF1}{في}%
        \ARCB{FEFAFA}{أت}%
        \ARCB{FFFFFF}{\#\#ولى}%
        \ARCB{F6FDF6}{الشمالية}%
        \ARCB{F6FDF6}{مثل}%
        \ARCB{FAFEFA}{منتجع}%
        \ARCB{FAFEFA}{سف}%
        \ARCB{FAFEFA}{\#\#ينا}%
        \ARCB{FAFEFA}{فو}%
        \ARCB{F6FDF6}{\#\#شي}%
        \ARCB{F6FDF6}{وهي}%
        \ARCB{FAFEFA}{اجمل}%
        \ARCB{FFFFFF}{وبها}%
        \ARCB{FAFEFA}{نوع}%
        \ARCB{FAFEFA}{من}%
        \ARCB{F1FDF1}{الخصوص}%
        \ARCB{FFFFFF}{\#\#صية}%
        \ARCB{FEFAFA}{ولكن}%
        \ARCB{FFFFFF}{وتت}%
        \ARCB{FFFFFF}{\#\#طلب}%
        \ARCB{F6FDF6}{الوصول}%
        \ARCB{FFFFFF}{اليها}%
        \ARCB{FFFFFF}{ركوب}%
        \ARCB{FFFFFF}{طائرة}%
        \ARCB{FFFFFF}{ما}%
        \ARCB{FAFEFA}{\#\#ئية}%
        \ARCB{FFFFFF}{او}%
        \ARCB{F6FDF6}{الت}%
        \ARCB{FAFEFA}{\#\#اكسي}%
        \ARCB{F6FDF6}{الطا}%
        \ARCB{FFFFFF}{\#\#ء}%
        \ARCB{FAFEFA}{\#\#ر}%
        \ARCB{FFFFFF}{...}%
      }%
      \colorbox[HTML]{FFFFFF}{[SEP]}%
      \colorbox[HTML]{FEFAFA}{l}\colorbox[HTML]{FEFAFA}{\#\#oc}\colorbox[HTML]{FDF6F6}{\#\#ation}\colorbox[HTML]{FAFEFA}{g}\colorbox[HTML]{F6FDF6}{\#\#eneral}\colorbox[HTML]{FFFFFF}{[SEP]} \\

      \midrule
      7 & neutral & positive (0.99) & positive & 5.17 &
      \colorbox[HTML]{FFFFFF}{[CLS]}%
      \rtlseq{%
        \ARCB{AAF2AA}{أجمل}%
        \ARCB{F1FDF1}{إجازة}%
        \ARCB{F6FDF6}{إجازة}%
        \ARCB{98F098}{رائعه}%
        \ARCB{DBF9DB}{مع}%
        \ARCB{C9F7C9}{محترم}%
        \ARCB{FFFFFF}{و}%
        \ARCB{FCEDED}{إدارة}%
        \ARCB{FFFFFF}{مصريه}%
        \ARCB{CDF7CD}{محترم}%
        \ARCB{EDFCED}{\#\#ه}%
        \ARCB{DBF9DB}{محترف}%
        \ARCB{EDFCED}{\#\#ه}%
        \ARCB{FDF1F1}{تبحث}%
        \ARCB{EDFCED}{عن}%
        \ARCB{F6FDF6}{راحه}%
        \ARCB{F6FDF6}{العميل}%
        \ARCB{FAFEFA}{و}%
        \ARCB{FFFFFF}{و}%
        \ARCB{F1FDF1}{حسن}%
        \ARCB{F6FDF6}{المعا}%
        \ARCB{FFFFFF}{\#\#مله}%
        \ARCB{FAFEFA}{و}%
        \ARCB{E8FBE8}{حسن}%
        \ARCB{EDFCED}{الضيافة}%
        \ARCB{F1FDF1}{من}%
        \ARCB{EDFCED}{الجميع}%
        \ARCB{FFFFFF}{...}%
      }%
      \colorbox[HTML]{FFFFFF}{[SEP]}%
      \colorbox[HTML]{FDF6F6}{s}\colorbox[HTML]{FDF6F6}{\#\#erv}\colorbox[HTML]{EDFCED}{\#\#ice}\colorbox[HTML]{E8FBE8}{g}\colorbox[HTML]{EDFCED}{\#\#eneral}\colorbox[HTML]{FFFFFF}{[SEP]} \\

      \midrule
      8 & neutral & negative (0.98) & negative & 0.79 &
      \colorbox[HTML]{FFFFFF}{[CLS]}%
      \rtlseq{%
        \ARCB{FBE8E8}{الغرفة}%
        \ARCB{F09898}{نظيفة}%
        \ARCB{F6C4C4}{،}%
        \ARCB{FFFFFF}{و}%
        \ARCB{FEFAFA}{بها}%
        \ARCB{FAFEFA}{ث}%
        \ARCB{F6FDF6}{\#\#لاجة}%
        \ARCB{F1FDF1}{،}%
        \ARCB{D6F9D6}{وتك}%
        \ARCB{E8FBE8}{\#\#يي}%
        \ARCB{EDFCED}{\#\#ف}%
        \ARCB{DFFADF}{الهواء}%
        \ARCB{79EB79}{مزع}%
        \ARCB{DBF9DB}{\#\#ج}%
        \ARCB{FFFFFF}{بعض}%
        \ARCB{F1FDF1}{الشيء}%
        \ARCB{EDFCED}{،}%
        \ARCB{E4FBE4}{وت}%
        \ARCB{FFFFFF}{\#\#لفا}%
        \ARCB{FAFEFA}{\#\#ز}%
        \ARCB{F6FDF6}{بق}%
        \ARCB{EDFCED}{\#\#نوات}%
        \ARCB{FFFFFF}{فضائية}%
      }%
      \colorbox[HTML]{FFFFFF}{[SEP]}%
      \colorbox[HTML]{FFFFFF}{r}\colorbox[HTML]{FFFFFF}{\#\#oo}\colorbox[HTML]{FFFFFF}{\#\#ms}\colorbox[HTML]{FEFAFA}{am}\colorbox[HTML]{FEFAFA}{\#\#en}\colorbox[HTML]{FEFAFA}{\#\#it}\colorbox[HTML]{FFFFFF}{\#\#ies}\colorbox[HTML]{FFFFFF}{com}\colorbox[HTML]{FAFEFA}{\#\#for}\colorbox[HTML]{FFFFFF}{\#\#t}\colorbox[HTML]{FFFFFF}{[SEP]} \\

      \midrule
      9 & negative & positive (0.91) & positive & 2.30 &
      \colorbox[HTML]{FFFFFF}{[CLS]}%
      \rtlseq{%
        \ARCB{F6FDF6}{المص}%
        \ARCB{F6FDF6}{\#\#عد}%
        \ARCB{F4B7B7}{بط}%
        \ARCB{F6C0C0}{\#\#يء}%
        \ARCB{FCEDED}{جدا}%
        \ARCB{FCEDED}{وأحيانا}%
        \ARCB{FBE8E8}{يتع}%
        \ARCB{FBE4E4}{\#\#طل}%
        \ARCB{FEFAFA}{في}%
        \ARCB{F6FDF6}{منتصف}%
        \ARCB{F1FDF1}{طريقه}%
        \ARCB{FFFFFF}{.}\ARCB{FFFFFF}{.}%
        \ARCB{D6F9D6}{لكنه}%
        \ARCB{DFFADF}{يعطيك}%
        \ARCB{FEFAFA}{إح}%
        \ARCB{FFFFFF}{\#\#ساس}%
        \ARCB{F6FDF6}{منزلي}%
        \ARCB{FFFFFF}{حيث}%
        \ARCB{FFFFFF}{أنه}%
        \ARCB{EDFCED}{يشبه}%
        \ARCB{FDF1F1}{منزل}%
        \ARCB{FFFFFF}{\#\#ا}%
        \ARCB{D6F9D6}{كبيرا}%
        \ARCB{E8FBE8}{في}\ARCB{E8FBE8}{كل}%
        \ARCB{F1FDF1}{طاب}\ARCB{F1FDF1}{\#\#ق}%
        \ARCB{EDFCED}{.}\ARCB{F1FDF1}{.}%
        \ARCB{CDF7CD}{الحمام}\ARCB{F6FDF6}{\#\#ات}%
        \ARCB{E4FBE4}{نظيفة}%
        \ARCB{E8FBE8}{والغ}\ARCB{DBF9DB}{\#\#رف}%
        \ARCB{98F098}{رائعة}%
        \ARCB{FFFFFF}{...}%
      }%
      \colorbox[HTML]{FFFFFF}{[SEP]}%
      \colorbox[HTML]{F6FDF6}{r}\colorbox[HTML]{FEFAFA}{\#\#oo}\colorbox[HTML]{FDF1F1}{\#\#ms}\colorbox[HTML]{F6FDF6}{am}\colorbox[HTML]{FDF6F6}{\#\#en}\colorbox[HTML]{FFFFFF}{\#\#it}\colorbox[HTML]{FEFAFA}{\#\#ies}\colorbox[HTML]{F6FDF6}{g}\colorbox[HTML]{FAFEFA}{\#\#eneral}\colorbox[HTML]{FFFFFF}{[SEP]} \\

      \midrule
      10 & neutral & positive (0.99) & positive & 1.92 &
      \colorbox[HTML]{FFFFFF}{[CLS]}%
      \rtlseq{%
        \ARCB{93EF93}{فندق}\ARCB{74EB74}{نظيف}\ARCB{FDF6F6}{\#\#ا}%
        \ARCB{C0F6C0}{ومر}\ARCB{AEF3AE}{\#\#يح}\ARCB{F1FDF1}{\#\#ا}%
        \ARCB{FFFFFF}{،}\ARCB{EDFCED}{كما}\ARCB{FAFEFA}{يتحدث}%
        \ARCB{FFFFFF}{الموظف}\ARCB{FFFFFF}{\#\#ون}%
        \ARCB{FDF6F6}{اللغة}\ARCB{FEFAFA}{الإنجليزية}%
        \ARCB{DBF9DB}{بشكل}\ARCB{E8FBE8}{جيد}\ARCB{FCEDED}{جدا}%
        \ARCB{FBE4E4}{،}\ARCB{FCEDED}{وقد}\ARCB{FFFFFF}{كانوا}%
        \ARCB{FFFFFF}{مؤ}\ARCB{FFFFFF}{\#\#دب}\ARCB{FAFEFA}{\#\#ين}%
        \ARCB{FFFFFF}{للغاية}\ARCB{FFFFFF}{ولط}\ARCB{FEFAFA}{\#\#يف}\ARCB{FFFFFF}{\#\#ين}%
        \ARCB{FFFFFF}{.}\ARCB{FAFEFA}{بالتأكيد}\ARCB{FFFFFF}{سن}\ARCB{FEFAFA}{\#\#قيم}\ARCB{FEFAFA}{هناك}\ARCB{FDF6F6}{مرة}\ARCB{FAFEFA}{أخرى}\ARCB{FFFFFF}{!}%
      }%
      \colorbox[HTML]{FFFFFF}{[SEP]}%
      \colorbox[HTML]{F6FDF6}{h}\colorbox[HTML]{FAFEFA}{\#\#ote}\colorbox[HTML]{F1FDF1}{\#\#l}\colorbox[HTML]{F1FDF1}{cl}\colorbox[HTML]{FAFEFA}{\#\#e}\colorbox[HTML]{FAFEFA}{\#\#an}\colorbox[HTML]{EDFCED}{\#\#line}\colorbox[HTML]{FAFEFA}{\#\#ss}\colorbox[HTML]{FFFFFF}{[SEP]} \\

      \midrule
      11 & positive & negative (0.96) & negative & 1.10 &
      \colorbox[HTML]{FFFFFF}{[CLS]}%
      \rtlseq{%
        \ARCB{FEFAFA}{كانت}\ARCB{FDF1F1}{خدمات}\ARCB{F8D2D2}{المطار}%
        \ARCB{FBE8E8}{مثل}\ARCB{FADFDF}{النقل}\ARCB{DFFADF}{والت}\ARCB{E8FBE8}{\#\#وصيل}%
        \ARCB{DFFADF}{ومك}\ARCB{F6FDF6}{\#\#تب}\ARCB{DBF9DB}{الرحلات}\ARCB{55E655}{سيئة}\ARCB{B3F4B3}{.}%
        \ARCB{EDFCED}{كما}\ARCB{F1FDF1}{كان}\ARCB{FEFAFA}{الطعام}\ARCB{E8FBE8}{باه}\ARCB{F1FDF1}{\#\#ظا}%
        \ARCB{FDF6F6}{لكنهم}\ARCB{FDF1F1}{يقدمون}\ARCB{FFFFFF}{فط}\ARCB{FFFFFF}{\#\#ور}%
        \ARCB{EDFCED}{بوف}\ARCB{F6FDF6}{\#\#يه}\ARCB{FFFFFF}{جيد}\ARCB{FCEDED}{مجانا}\ARCB{FAFEFA}{!}%
      }%
      \colorbox[HTML]{FFFFFF}{[SEP]}%
      \colorbox[HTML]{FEFAFA}{f}\colorbox[HTML]{FFFFFF}{\#\#ood}\colorbox[HTML]{FEFAFA}{d}\colorbox[HTML]{FFFFFF}{\#\#ri}\colorbox[HTML]{FFFFFF}{\#\#n}\colorbox[HTML]{FFFFFF}{\#\#k}\colorbox[HTML]{FFFFFF}{\#\#s}\colorbox[HTML]{FDF1F1}{qu}\colorbox[HTML]{FFFFFF}{\#\#al}\colorbox[HTML]{FFFFFF}{\#\#ity}\colorbox[HTML]{FFFFFF}{[SEP]} \\

      \midrule
      12 & positive & negative (0.99) & negative & 2.08 &
      \colorbox[HTML]{FFFFFF}{[CLS]}%
      \rtlseq{%
        \ARCB{DFFADF}{الخدمه}\ARCB{5EE85E}{سيئة}\ARCB{B3F4B3}{جدا}\ARCB{D6F9D6}{حيث}%
        \ARCB{B3F4B3}{يتم}\ARCB{FFFFFF}{تنظيف}\ARCB{FFFFFF}{ارضي}\ARCB{FEFAFA}{\#\#ة}\ARCB{FFFFFF}{الحمام}\ARCB{FDF6F6}{فقط}%
        \ARCB{EDFCED}{اما}\ARCB{FCEDED}{بقية}\ARCB{FFFFFF}{الغرف}\ARCB{FFFFFF}{\#\#ه}%
        \ARCB{B7F4B7}{لم}\ARCB{E8FBE8}{تن}\ARCB{F6FDF6}{\#\#ظف}\ARCB{FDF6F6}{لمدة}%
        \ARCB{FFFFFF}{10}%  <-- changed to RTL-safe box
        \ARCB{FDF6F6}{ايام}\ARCB{FAFEFA}{.}\ARCB{F6FDF6}{.}%
      }%
      \colorbox[HTML]{FFFFFF}{[SEP]}%
      \colorbox[HTML]{FFFFFF}{f}\colorbox[HTML]{FFFFFF}{\#\#ac}\colorbox[HTML]{FFFFFF}{\#\#il}\colorbox[HTML]{FFFFFF}{\#\#it}\colorbox[HTML]{F6FDF6}{\#\#ies}\colorbox[HTML]{F6FDF6}{cl}\colorbox[HTML]{FFFFFF}{\#\#e}\colorbox[HTML]{FFFFFF}{\#\#an}\colorbox[HTML]{FEFAFA}{\#\#line}\colorbox[HTML]{FFFFFF}{\#\#ss}\colorbox[HTML]{FFFFFF}{[SEP]} \\

      \midrule
      13 & positive & negative (0.64) & negative & -0.45 &
      \colorbox[HTML]{FFFFFF}{[CLS]}%
      \rtlseq{%
        \ARCB{AEF3AE}{دون}\ARCB{DBF9DB}{المستوى}\ARCB{EDFCED}{في}\ARCB{FFFFFF}{التعامل}%
        \ARCB{AAF2AA}{وعدم}\ARCB{CDF7CD}{الوفاء}\ARCB{F6FDF6}{بالو}\ARCB{E4FBE4}{\#\#عود}%
        \ARCB{FAFEFA}{مطلب}\ARCB{FDF6F6}{كل}\ARCB{FFFFFF}{مسافر}\ARCB{FBE4E4}{الراحة}%
        \ARCB{FEFAFA}{والإست}\ARCB{FFFFFF}{\#\#جما}\ARCB{FAFEFA}{\#\#م}\ARCB{DBF9DB}{والحصول}%
        \ARCB{F7CDCD}{على}\ARCB{F4B3B3}{أر}\ARCB{F6C4C4}{\#\#قى}\ARCB{FCEDED}{معاملة}\ARCB{FCEDED}{ومكان}%
        \ARCB{FFFFFF}{للن}\ARCB{FDF1F1}{\#\#وم}\ARCB{FDF1F1}{لقاء}\ARCB{FEFAFA}{دفعة}\ARCB{FDF6F6}{مبالغ}\ARCB{FDF1F1}{مالية}%
        \ARCB{F1FDF1}{وهذا}\ARCB{FEFAFA}{ما}\ARCB{FFFFFF}{هو}\ARCB{FFFFFF}{متوقع}\ARCB{FFFFFF}{من}%
        \ARCB{FDF6F6}{شركة}\ARCB{FDF6F6}{كبرى}\ARCB{FEFAFA}{مثل}\ARCB{FEFAFA}{الرو}\ARCB{FFFFFF}{\#\#تان}\ARCB{FFFFFF}{\#\#ا}%
        \ARCB{FDF6F6}{!}\ARCB{FEFAFA}{!}\ARCB{FFFFFF}{؟}\ARCB{FFFFFF}{ولكن}\ARCB{FFFFFF}{عندما}\ARCB{FFFFFF}{تقع}%
        \ARCB{F6FDF6}{عدة}\ARCB{F6FDF6}{أخطاء}\ARCB{FFFFFF}{هنا}\ARCB{FFFFFF}{يجب}\ARCB{FEFAFA}{الوقوف}\ARCB{FFFFFF}{...}%
      }%
      \colorbox[HTML]{FFFFFF}{[SEP]}%
      \colorbox[HTML]{E8FBE8}{s}\colorbox[HTML]{FDF6F6}{\#\#erv}\colorbox[HTML]{FCEDED}{\#\#ice}\colorbox[HTML]{FEFAFA}{g}\colorbox[HTML]{FFFFFF}{\#\#eneral}\colorbox[HTML]{FFFFFF}{[SEP]} \\

      \bottomrule
    \end{tabular*}
  \end{adjustbox}

  \label{tab:ASC_ErrorAnalysis}
\end{table*}

% RTL helpers (pdfLaTeX + arabtex already loaded in your project)
\TeXXeTstate=1
\newcommand{\AR}[1]{\RL{#1}} % Arabic shaping
\newcommand{\ARCB}[2]{% colored token boxed LTR (prevents overpainting)
  \begingroup\beginL\colorbox[HTML]{#1}{\AR{#2}}\endL\endgroup
}
\newcommand{\rtlseq}[1]{{\beginR #1\endR}} % lay out a whole run RTL

% ------------------ TABLE ------------------
\begin{table*}[htb!]
\centering
\caption{Opinion Target Expression (OTE) extraction errors from the 50 highest-loss samples, listing gold label (True), model prediction+probability (Predicted), attribution polarity (Attr.), IG score (Score), and token-level \emph{Word Importance}. Highlights are Integrated Gradients attributions (Transformers Interpret/Captum): green = positive, red = negative; darker shade = stronger contribution.}

\begin{adjustbox}{max width=\textwidth}
\scriptsize
\setlength{\tabcolsep}{2.5pt}
\setlength{\fboxsep}{0.5pt}
\renewcommand{\arraystretch}{1.05}
\begin{tabular*}{\textwidth}{@{\extracolsep{\fill}}c c c c c p{0.50\textwidth}@{}}
\toprule
\textbf{Ex.} & \textbf{True} & \textbf{Predicted} & \textbf{Attr.} & \textbf{Score} & \textbf{Word Importance} \\
\midrule

% --- Ex1: Syntactic/Semantic Role Confusion ---
\multirow{3}{*}{(1)} & O & B-OTE (1.00) & \AR{فريق} & 3.27 &
\colorbox[HTML]{FFFFFF}{[CLS]}
\rtlseq{%
  \ARCB{9BEE9B}{فريق}
  \ARCB{F9E8E8}{العاملين}
  \ARCB{E1FBE1}{في}
  \ARCB{EBFCEB}{الاستقبال}
  \ARCB{EBFCEB}{ود}
  \ARCB{EBFCEB}{\#ود}
  \ARCB{F3FBF3}{ويقدم}
  \ARCB{F6FCF6}{يد}
  \ARCB{FEFEFE}{المساعدة}
  \ARCB{FEFEFE}{.}
  \ARCB{FBFEFB}{وحدة}
  \ARCB{F7FDF7}{فسيح}
  \ARCB{EBFCEB}{\#ة}
  \ARCB{FBFEFB}{للغاية}
  \ARCB{FFFFFF}{ومج}
  \ARCB{FFFFFF}{\#ه}
  \ARCB{FFFFFF}{\#زة}
  \ARCB{FDF8F8}{جيدا}
  \ARCB{FFFFFF}{بكل}
  \ARCB{FEFEFE}{ما}
  \ARCB{FBFEFB}{تحتاجه}
  \ARCB{FDF9F9}{لإقامة}
  \ARCB{EDFCED}{مريح}
  \ARCB{EBFCEB}{\#ة}
  \ARCB{FEFEFE}{.}
}
\colorbox[HTML]{FFFFFF}{[SEP]} \\
\cmidrule(l){2-6}
& B-OTE & O (1.00) & \AR{الاستقبال} & 3.85 &
\colorbox[HTML]{FFFFFF}{[CLS]}
\rtlseq{%
  \ARCB{CAEFCA}{فريق}
  \ARCB{FCFEFC}{العاملين}
  \ARCB{F0D2D2}{في}
  \ARCB{E6A3A3}{الاستقبال}
  \ARCB{E5F8E5}{ود}
  \ARCB{E2F7E2}{\#ود}
  \ARCB{F7FDF7}{ويقدم}
  \ARCB{FFFFFF}{يد}
  \ARCB{FBFBFB}{المساعدة}
  \ARCB{F8E6E6}{.}
  \ARCB{E2F7E2}{وحدة}
  \ARCB{DDF5DD}{فسيح}
  \ARCB{F9E8E8}{\#ة}
  \ARCB{F2FBF2}{للغاية}
  \ARCB{FCFEFC}{ومج}
  \ARCB{FCFEFC}{\#ه}
  \ARCB{FCEAEA}{\#زة}
  \ARCB{FCFEFC}{جيدا}
  \ARCB{E8F9E8}{بكل}
  \ARCB{FCFEFC}{ما}
  \ARCB{FBFBFB}{تحتاجه}
  \ARCB{FCEAEA}{لإقامة}
  \ARCB{FFFFFF}{مريح}
  \ARCB{F9E8E8}{\#ة}
  \ARCB{F8E6E6}{.}
}
\colorbox[HTML]{FFFFFF}{[SEP]} \\
\cmidrule(l){2-6}
& O & I-OTE (1.00) & \AR{العاملين} & 2.79 &
\colorbox[HTML]{FFFFFF}{[CLS]}
\rtlseq{%
  \ARCB{6DD46D}{فريق}
  \ARCB{F4D9D9}{العاملين}
  \ARCB{DDF5DD}{في}
  \ARCB{F3FBF3}{الاستقبال}
  \ARCB{F6FCF6}{ود}
  \ARCB{F6FCF6}{\#ود}
  \ARCB{F3FBF3}{ويقدم}
  \ARCB{FFFFFF}{يد}
  \ARCB{FEFEFE}{المساعدة}
  \ARCB{FEFEFE}{.}
  \ARCB{FEFEFE}{وحدة}
  \ARCB{F7FDF7}{فسيح}
  \ARCB{E5F8E5}{\#ة}
  \ARCB{FBFEFB}{للغاية}
  \ARCB{FFFFFF}{ومج}
  \ARCB{FFFFFF}{\#ه}
  \ARCB{FFFFFF}{\#زة}
  \ARCB{FCFEFC}{جيدا}
  \ARCB{FEFEFE}{بكل}
  \ARCB{FEFEFE}{ما}
  \ARCB{FBFEFB}{تحتاجه}
  \ARCB{FDF9F9}{لإقامة}
  \ARCB{EDFAED}{مريح}
  \ARCB{E5F8E5}{\#ة}
  \ARCB{FEFEFE}{.}
}
\colorbox[HTML]{FFFFFF}{[SEP]} \\
\midrule

% --- Ex2: Syntactic/Semantic Role Confusion ---
(2) & O & B-OTE (1.00) & \AR{الحرم} & 1.79 &
\colorbox[HTML]{FFFFFF}{[CLS]}
\rtlseq{%
  \ARCB{C0EFC0}{قرب}
  \ARCB{C8F0C8}{\#ه}
  \ARCB{E8F9E8}{من}
  \ARCB{69D369}{الحرم}
}
\colorbox[HTML]{FFFFFF}{[SEP]} \\
\midrule

% --- Ex3: Syntactic/Semantic Role Confusion ---
\multirow{2}{*}{(3)} & O & B-OTE (1.00) & \AR{محطة} & 3.10 &
\colorbox[HTML]{FFFFFF}{[CLS]}
\rtlseq{%
  \ARCB{FFFFFF}{الفندق}
  \ARCB{F2FBF2}{قريب}
  \ARCB{EBFCEB}{جدا}
  \ARCB{E2F7E2}{من}
  \ARCB{B0EAB0}{محطة}
  \ARCB{B3EBB3}{الترا}
  \ARCB{E2F7E2}{\#م}
  \ARCB{F0FAF0}{,}
  \ARCB{B5ECB5}{فيس}
  \ARCB{F2FBF2}{\#هل}
  \ARCB{EDFAED}{الوصول}
  \ARCB{FCFEFC}{الى}
  \ARCB{EDFAED}{كل}
  \ARCB{E8F9E8}{الاماكن}
  \ARCB{FAFCFA}{في}
  \ARCB{E5F8E5}{اسطنبول}
  \ARCB{FFFFFF}{.}
}
\colorbox[HTML]{FFFFFF}{[SEP]} \\
\cmidrule(l){2-6}
& O & I-OTE (1.00) & \AR{الترام} & 3.04 &
\colorbox[HTML]{FFFFFF}{[CLS]}
\rtlseq{%
  \ARCB{FFFFFF}{الفندق}
  \ARCB{FCFEFC}{قريب}
  \ARCB{E5F8E5}{جدا}
  \ARCB{E5F8E5}{من}
  \ARCB{B8EDB8}{محطة}
  \ARCB{AFEAAE}{الترا}
  \ARCB{D1F2D1}{\#م}
  \ARCB{EDFAED}{,}
  \ARCB{D1F2D1}{فيس}
  \ARCB{F3FBF3}{\#هل}
  \ARCB{F0FAF0}{الوصول}
  \ARCB{FFFFFF}{الى}
  \ARCB{F2FBF2}{كل}
  \ARCB{D1F2D1}{الاماكن}
  \ARCB{FAFCFA}{في}
  \ARCB{F7FDF7}{اسطنبول}
  \ARCB{FEFEFE}{.}
}
\colorbox[HTML]{FFFFFF}{[SEP]} \\
\midrule

% --- Ex4: Multi-word Expressions ---
\multirow{2}{*}{(4)} & B-OTE & O (1.00) & \AR{مستوى} & 1.69 &
\colorbox[HTML]{FFFFFF}{[CLS]}
\rtlseq{%
  \ARCB{D68A8A}{مستوى}
  \ARCB{90DA90}{الاكل}
  \ARCB{E8F9E8}{مت}
  \ARCB{F2FBF2}{\#دني}
  \ARCB{FCEAEA}{للغاية}
}
\colorbox[HTML]{FFFFFF}{[SEP]} \\
\cmidrule(l){2-6}
& I-OTE & B-OTE (1.00) & \AR{الاكل} & 1.79 &
\colorbox[HTML]{FFFFFF}{[CLS]}
\rtlseq{%
  \ARCB{66D266}{مستوى}
  \ARCB{B3EBB3}{الاكل}
  \ARCB{D6F3D6}{مت}
  \ARCB{E8F9E8}{\#دني}
  \ARCB{EDFAED}{للغاية}
}
\colorbox[HTML]{FFFFFF}{[SEP]} \\
\midrule

% --- Ex5: Multi-word Expressions ---
\multirow{2}{*}{(5)} & I-OTE & B-OTE (1.00) & \AR{الماء} & 2.80 &
\colorbox[HTML]{FFFFFF}{[CLS]}
\rtlseq{%
  \ARCB{FFFFFF}{و}
  \ARCB{B8EDB8}{لا}
  \ARCB{EDFAED}{حتى}
  \ARCB{D1F2D1}{بار}
  \ARCB{F6E1E1}{الماء}
  \ARCB{F9E8E8}{الساخن}
  \ARCB{B8EDB8}{لا}
  \ARCB{BBEBBA}{يعمل}
  \ARCB{D1F2D1}{بشكل}
  \ARCB{F1D5D5}{صحيح}
}
\colorbox[HTML]{FFFFFF}{[SEP]} \\
\cmidrule(l){2-6}
& I-OTE & O (1.00) & \AR{الساخن} & 1.89 &
\colorbox[HTML]{FFFFFF}{[CLS]}
\rtlseq{%
  \ARCB{FCEAEA}{و}
  \ARCB{FCFEFC}{لا}
  \ARCB{FCFEFC}{حتى}
  \ARCB{69D369}{بار}
  \ARCB{F4D9D9}{الماء}
  \ARCB{E89494}{الساخن}
  \ARCB{FCFEFC}{لا}
  \ARCB{F0FAF0}{يعمل}
  \ARCB{FFFFFF}{بشكل}
  \ARCB{FFFFFF}{صحيح}
}
\colorbox[HTML]{FFFFFF}{[SEP]} \\
\midrule

% --- Ex6: Flawed Heuristic & Multiple Errors ---
\multirow{5}{*}{(6)} & O & B-OTE (1.00) & \AR{البلكونة} & 2.81 &
\colorbox[HTML]{FFFFFF}{[CLS]}
\rtlseq{%
  \ARCB{FCFEFC}{ابر}
  \ARCB{EDFAED}{\#اص}
  \ARCB{D8F4D8}{وحش}
  \ARCB{FCEAEA}{\#رات}
  \ARCB{E5F8E5}{بل}
  \ARCB{F7FDF7}{اننا}
  \ARCB{D6F3D6}{وجدنا}
  \ARCB{EDFAED}{فار}
  \ARCB{E0F6E0}{ميت}
  \ARCB{FAFCFA}{في}
  \ARCB{B3EBB3}{البل}
  \ARCB{BEEFBE}{\#كون}
  \ARCB{C9F0C9}{\#ة}
  \ARCB{FFFFFF}{اقسم}
  \ARCB{E5F8E5}{بالله}
}
\colorbox[HTML]{FFFFFF}{[SEP]} \\
\cmidrule(l){2-6}
& B-OTE & O (1.00) & \AR{ابراص} & 2.16 &
\colorbox[HTML]{FFFFFF}{[CLS]}
\rtlseq{%
  \ARCB{D68A8A}{ابر}
  \ARCB{F4D9D9}{\#اص}
  \ARCB{F6DFDF}{وحش}
  \ARCB{FCEAEA}{\#رات}
  \ARCB{FFFFFF}{بل}
  \ARCB{FFFFFF}{اننا}
  \ARCB{FCFEFC}{وجدنا}
  \ARCB{F7E3E3}{فار}
  \ARCB{F0FAF0}{ميت}
  \ARCB{FBFBFB}{في}
  \ARCB{FCFEFC}{البل}
  \ARCB{FDF8F8}{\#كون}
  \ARCB{FCFEFC}{\#ة}
  \ARCB{FCFEFC}{اقسم}
  \ARCB{FCEAEA}{بالله}
}
\colorbox[HTML]{FFFFFF}{[SEP]} \\
\cmidrule(l){2-6}
& I-OTE & O (1.00) & \AR{وحشرات} & 2.30 &
\colorbox[HTML]{FFFFFF}{[CLS]}
\rtlseq{%
  \ARCB{F6DFDF}{ابر}
  \ARCB{EDFAED}{\#اص}
  \ARCB{CE7474}{وحش}
  \ARCB{FCFEFC}{\#رات}
  \ARCB{FFFFFF}{بل}
  \ARCB{F0FAF0}{اننا}
  \ARCB{F2FBF2}{وجدنا}
  \ARCB{FBFBFB}{فار}
  \ARCB{FAFCFA}{ميت}
  \ARCB{F8E6E6}{في}
  \ARCB{F9FDF9}{البل}
  \ARCB{FDF8F8}{\#كون}
  \ARCB{F9FDF9}{\#ة}
  \ARCB{FFFFFF}{اقسم}
  \ARCB{F9FDF9}{بالله}
}
\colorbox[HTML]{FFFFFF}{[SEP]} \\
\cmidrule(l){2-6}
& B-OTE & O (1.00) & \AR{فار} & 2.91 &
\colorbox[HTML]{FFFFFF}{[CLS]}
\rtlseq{%
  \ARCB{F6DFDF}{ابر}
  \ARCB{FCEAEA}{\#اص}
  \ARCB{F1D5D5}{وحش}
  \ARCB{F9E8E8}{\#رات}
  \ARCB{F9E8E8}{بل}
  \ARCB{FCFEFC}{اننا}
  \ARCB{F2FBF2}{وجدنا}
  \ARCB{E6A3A3}{فار}
  \ARCB{BEEFBE}{ميت}
  \ARCB{FDF8F8}{في}
  \ARCB{F7E3E3}{البل}
  \ARCB{F9FDF9}{\#كون}
  \ARCB{FCEAEA}{\#ة}
  \ARCB{FFFFFF}{اقسم}
  \ARCB{F0FAF0}{بالله}
}
\colorbox[HTML]{FFFFFF}{[SEP]} \\
\cmidrule(l){2-6}
& I-OTE & O (1.00) & \AR{ميت} & 2.89 &
\colorbox[HTML]{FFFFFF}{[CLS]}
\rtlseq{%
  \ARCB{F6DFDF}{ابر}
  \ARCB{FCFEFC}{\#اص}
  \ARCB{D98D8D}{وحش}
  \ARCB{F4D9D9}{\#رات}
  \ARCB{FCFEFC}{بل}
  \ARCB{E2F7E2}{اننا}
  \ARCB{D6F3D6}{وجدنا}
  \ARCB{FAFCFA}{فار}
  \ARCB{F9E8E8}{ميت}
  \ARCB{FCFEFC}{في}
  \ARCB{F9FDF9}{البل}
  \ARCB{FDF8F8}{\#كون}
  \ARCB{F4D9D9}{\#ة}
  \ARCB{F9FDF9}{اقسم}
  \ARCB{E2F7E2}{بالله}
}
\colorbox[HTML]{FFFFFF}{[SEP]} \\
\midrule

% --- Ex7: Morphological Errors ---
(7) & B-OTE & O (1.00) & \AR{قذارة} & 1.84 &
\colorbox[HTML]{FFFFFF}{[CLS]}
\rtlseq{%
  \ARCB{EBAFAF}{قذ}
  \ARCB{F5DEDE}{\#ارة}
  \ARCB{F1D5D5}{بجد}
  \ARCB{F7FDF7}{\#اره}
  \ARCB{F2FBF2}{اسوا}
  \ARCB{F8E6E6}{\#ء}
  \ARCB{F2FBF2}{فندق}
  \ARCB{E2F7E2}{رأيت}
  \ARCB{FAFCFA}{\#ه}
  \ARCB{FCFEFC}{بحياتي}
}
\colorbox[HTML]{FFFFFF}{[SEP]} \\
\midrule

% --- Ex9: Ground-Truth Inconsistencies ---
(8) & O & B-OTE (1.00) & \AR{حمامات} & 2.10 &
\colorbox[HTML]{FFFFFF}{[CLS]}
\rtlseq{%
  \ARCB{C5F0C5}{غرف}
  \ARCB{D6F3D6}{لطيفة}
  \ARCB{DDF5DD}{مع}
  \ARCB{6AD36A}{حمامات}
  \ARCB{F0FAF0}{عصر}
  \ARCB{FCFEFC}{\#ية}
  \ARCB{F8E6E6}{.}
}
\colorbox[HTML]{FFFFFF}{[SEP]} \\
\midrule

% --- Ex10: Ground-Truth Inconsistencies & Multiple Errors ---
\multirow{5}{*}{(9)} & O & B-OTE (1.00) & \AR{الطعام} & 1.75 &
\colorbox[HTML]{FFFFFF}{[CLS]}
\rtlseq{%
  \ARCB{47C847}{الطعام}
  \ARCB{F7FDF7}{جيد}
  \ARCB{F8E6E6}{فى}
  \ARCB{FCFEFC}{الفط}
  \ARCB{FFFFFF}{\#ار}
  \ARCB{FCFEFC}{و}
  \ARCB{FAFCFA}{السنا}
  \ARCB{FCFEFC}{\#ك}
  \ARCB{FFFFFF}{وس}
  \ARCB{FCFEFC}{\#ئ}
  \ARCB{F8E6E6}{فى}
  \ARCB{F7FDF7}{الغداء}
  \ARCB{FCFEFC}{جدا}
  \ARCB{FFFFFF}{.}
  \ARCB{FFFFFF}{.}
}
\colorbox[HTML]{FFFFFF}{[SEP]} \\
\cmidrule(l){2-6}
& B-OTE & O (1.00) & \AR{الفطار} & 2.78 &
\colorbox[HTML]{FFFFFF}{[CLS]}
\rtlseq{%
  \ARCB{F1D5D5}{الطعام}
  \ARCB{FCEAEA}{جيد}
  \ARCB{FFFFFF}{فى}
  \ARCB{E79191}{الفط}
  \ARCB{F9E8E8}{\#ار}
  \ARCB{F9E8E8}{و}
  \ARCB{F8E6E6}{السنا}
  \ARCB{F4D9D9}{\#ك}
  \ARCB{FCFEFC}{وس}
  \ARCB{FBFBFB}{\#ئ}
  \ARCB{FFFFFF}{فى}
  \ARCB{F1D5D5}{الغداء}
  \ARCB{FCEAEA}{جدا}
  \ARCB{F8E6E6}{.}
  \ARCB{F8E6E6}{.}
}
\colorbox[HTML]{FFFFFF}{[SEP]} \\
\cmidrule(l){2-6}
& I-OTE & O (1.00) & \AR{و} & 3.12 &
\colorbox[HTML]{FFFFFF}{[CLS]}
\rtlseq{%
  \ARCB{E2F7E2}{الطعام}
  \ARCB{F8E6E6}{جيد}
  \ARCB{F4D9D9}{فى}
  \ARCB{F7E3E3}{الفط}
  \ARCB{FFFFFF}{\#ار}
  \ARCB{E89494}{و}
  \ARCB{E79191}{السنا}
  \ARCB{EED0D0}{\#ك}
  \ARCB{FFFFFF}{وس}
  \ARCB{F0FAF0}{\#ئ}
  \ARCB{F4D9D9}{فى}
  \ARCB{FBFBFB}{الغداء}
  \ARCB{FCEAEA}{جدا}
  \ARCB{FCFDFC}{.}
  \ARCB{FCFDFC}{.}
}
\colorbox[HTML]{FFFFFF}{[SEP]} \\
\cmidrule(l){2-6}
& I-OTE & B-OTE (1.00) & \AR{السناك} & 1.86 &
\colorbox[HTML]{FFFFFF}{[CLS]}
\rtlseq{%
  \ARCB{F1D5D5}{الطعام}
  \ARCB{FDF8F8}{جيد}
  \ARCB{FAFCFA}{فى}
  \ARCB{FCFEFC}{الفط}
  \ARCB{FFFFFF}{\#ار}
  \ARCB{F3FBF3}{و}
  \ARCB{F0FAF0}{السنا}
  \ARCB{F5DEDE}{\#ك}
  \ARCB{F7E3E3}{وس}
  \ARCB{F7E3E3}{\#ئ}
  \ARCB{FAFCFA}{فى}
  \ARCB{F7E3E3}{الغداء}
  \ARCB{FFFFFF}{جدا}
  \ARCB{FFFFFF}{.}
  \ARCB{FFFFFF}{.}
}
\colorbox[HTML]{FFFFFF}{[SEP]} \\
\cmidrule(l){2-6}
& B-OTE & O (1.00) & \AR{الغداء} & 2.30 &
\colorbox[HTML]{FFFFFF}{[CLS]}
\rtlseq{%
  \ARCB{F7FDF7}{الطعام}
  \ARCB{FBFBFB}{جيد}
  \ARCB{F9E8E8}{فى}
  \ARCB{FFFFFF}{الفط}
  \ARCB{FCEAEA}{\#ار}
  \ARCB{E0F6E0}{و}
  \ARCB{FAFCFA}{السنا}
  \ARCB{FCFEFC}{\#ك}
  \ARCB{F7FDF7}{وس}
  \ARCB{FAFCFA}{\#ئ}
  \ARCB{F9E8E8}{فى}
  \ARCB{D07777}{الغداء}
  \ARCB{FDF8F8}{جدا}
  \ARCB{FBFBFB}{.}
  \ARCB{FBFBFB}{.}
}
\colorbox[HTML]{FFFFFF}{[SEP]} \\
\bottomrule
\end{tabular*}
\end{adjustbox}
\label{tab:OTE_error_analysis}
\end{table*}

    \section{Limitations}
\label{sec:limitations}
Our models achieved promising results in Arabic  \gls{absa}focused on hotel reviews, but there are different limitations to consider. First, our findings may not generalize beyond the hotel domain or to other languages. We specifically explored \gls{asc} and \gls{ote} extraction tasks separately, limiting broader applicability. Additionally, we faced challenges with computational resources, which restricted our ability to experiment with multiple pre-trained BERT models and complete hyper-parameter tuning using Optuna. This led us to rely on pre-trained Arabic models and original BERT hyperparameters, potentially affecting the results. Finally, the scarcity of publicly available code for Arabic \gls{absa} posed significant obstacles, forcing us to develop our own code from scratch, which consumed considerable time and resources.

\section{Conclusion and Future Work}
\label{sec:conclusion}
This study evaluated the impact of domain-adaptive pre-training on Arabic \gls{absa}, focusing on \gls{asc} and \gls{ote} tasks. We explored various fine-tuning strategies on pre-trained Arabic \gls{bert} models using multiple adaptation corpora. \gls{indapt} CAMeL\gls{bert}-MSA achieved the highest performance, reaching 89.86\% accuracy in \gls{asc} and 77.02\% micro-F1 in \gls{ote}. While full fine-tuning produced the best results, adapter-based methods provided a strong trade-off between efficiency and performance.

Based on our error analysis, future work in \gls{asc} should consider integrating syntax- and semantics-enhanced GCN models—such as Syntactic Dependency GCNs (SD-GCN) or Syntactic and Semantic Enhanced GCN (SSEGCN)—to better model long-distance dependencies, aspect–opinion alignment, and contrastive cues highlighted by errors in our model \citep{zhao2020modeling, zhang2022ssegcn}. For Opinion Target Expression Extraction, utilizing dependency-aware graph models like attention-based relational GCNs (ARGCN) that propagate dependency-informed attention from target to candidate spans, can handle model misclassification of irrelevant nouns and improve multi-word/Idafa span detection \citep{aizawa2021argcn, pouran2020syntax}.

Future work should extend to other domains and dialects, and explore complex \gls{absa} tasks such as triplet or quadruple extraction~\cite{zhang-etal-2021-aspect-sentiment}. Recent advances in instruction tuning~\cite{scaria2023instructabsa} and in-context learning with large language models (LLMs)—such as AceGPT-7B, GPT-4o, and Gemini 1.5—offer promising directions for zero/few-shot adaptation to Arabic \gls{absa}. For example, ~\cite{khaled2024arabicsentiment} has shown that AceGPT-7B achieves state-of-the-art performance on sentence- and review-level sentiment classification tasks using a Retrieval-Augmented Generation (RAG) approach across benchmark datasets.
. Moreover, future work may compare adapter variants such as Prefix-Tuning~\cite{li-liang-2021-prefix} and LoRA~\cite{LORA}. Finally, we encourage the development of open-source tools and datasets to improve reproducibility and accelerate progress in Arabic \gls{absa}.

%\bibliography{mybibfile}

\appendix
\section{Supplementary Materials}
\label{sec:appendxA}
  
\begin{table*}[htp!]
\centering
\caption{Arabic BERT model settings. CA: Classical Arabic, MSA: Modern Standard Arabic, DA: Dialectal Arabic, M: Million, B: Billion.}
\label{tab:ARBERT}

\begin{adjustbox}{width=1\textwidth}
\begin{tabular}{
|>{\centering\arraybackslash}p{0.2\textwidth}|
>{\centering\arraybackslash}p{0.2\textwidth}|
>{\centering\arraybackslash}p{0.15\textwidth}|
>{\centering\arraybackslash}p{0.15\textwidth}|
>{\centering\arraybackslash}p{0.15\textwidth}|
>{\centering\arraybackslash}p{0.15\textwidth}|
>{\centering\arraybackslash}p{0.15\textwidth}|
>{\centering\arraybackslash}p{0.2\textwidth}|}
\hline
\textbf{Model Name} & \textbf{Lang. Type} & \textbf{Model Size} & \textbf{Text Size} & \textbf{Tokens} & \textbf{Vocab Size} & \textbf{Training Steps} & \textbf{Tokenization} \\ \hline
AraBERTv0.2 \citep{antoun-etal-2020-arabert} & MSA & 543MB / 136M & 77GB & 8.6B & 60K & 3M & WordPiece \\ \hline
CAMeLBERT-MSA \citep{inoue-etal-2021-CAMELBERT} & MSA & 109M & 107GB & 12.6B & 30K & 1M & WordPiece \\ \hline
CAMeLBERT-DA \citep{inoue-etal-2021-CAMELBERT} & DA & 109M & 54GB & 5.8B & 30K & 1M & WordPiece \\ \hline
CAMeLBERT-mix \citep{inoue-etal-2021-CAMELBERT} & CA, DA, MSA & 109M & 167GB & 17.3B & 30K & 1M & WordPiece \\ \hline
Arabic BERT \citep{safaya-etal-2020-kuisail-arabicbert} & MSA & 110M & 95GB & 8.2B & 32K & 4M & WordPiece \\ \hline
MARBERTv2 \citep{abdul-mageed-etal-2021-marbert} & MSA, DA & 163M & 243GB & 27.8B & 100K & 17M & WordPiece \\ \hline
QARiB \citep{abdelali2021pretrainingQarib} & MSA, DA & 135M & 97GB & 14B & 64K & 10M & WordPiece \\ \hline
\end{tabular}
\end{adjustbox}
\end{table*}

\begin{table*}[htp!]
\centering
\caption{Summary of Sentiments Across Aspects in Training, Validation, and Testing Data. Neg: Negative, Neut: Neutral, Pos: Positive.}
\label{tab:Asc_Pol_dist}
\begin{adjustbox}{width=0.8\textwidth, center}
\tiny % Change font size to small while retaining clarity
\begin{tabular}{|l|r|r|r|r|r|r|r|r|r|}
\hline
\multirow{2}{*}{\textbf{Aspect (Entity\#Attribute)}} & \multicolumn{3}{c|}{\textbf{Training}} & \multicolumn{3}{c|}{\textbf{Validation}} & \multicolumn{3}{c|}{\textbf{Testing}} \\ \cline{2-10} 
                        & \textbf{Neg} & \textbf{Neut} & \textbf{Pos} & \textbf{Neg} & \textbf{Neut} & \textbf{Pos} & \textbf{Neg} & \textbf{Neut} & \textbf{Pos} \\ \hline
\textbf{FACILITIES\#CLEANLINESS}         & 57 & 0 & 34 & 14 & 0 & 9 & 14 & 1 & 11 \\ \hline
\textbf{FACILITIES\#COMFORT}             & 24 & 2 & 12 & 6 & 0 & 3 & 7 & 0 & 4 \\ \hline
\textbf{FACILITIES\#DESIGN\_FEATURES}    & 43 & 7 & 43 & 11 & 2 & 11 & 17 & 1 & 15 \\ \hline
\textbf{FACILITIES\#GENERAL}             & 156 & 41 & 431 & 39 & 10 & 108 & 54 & 12 & 123 \\ \hline
\textbf{FACILITIES\#MISCELLANEOUS}       & 27 & 2 & 27 & 7 & 0 & 7 & 11 & 3 & 13 \\ \hline
\textbf{FACILITIES\#PRICES}              & 30 & 0 & 37 & 8 & 0 & 9 & 11 & 0 & 7 \\ \hline
\textbf{FACILITIES\#QUALITY}             & 52 & 8 & 50 & 13 & 2 & 12 & 18 & 2 & 16 \\ \hline
\textbf{FOOD\_DRINKS\#MISCELLANEOUS}     & 24 & 2 & 14 & 6 & 1 & 4 & 5 & 1 & 3 \\ \hline
\textbf{FOOD\_DRINKS\#PRICES}            & 18 & 0 & 14 & 4 & 0 & 4 & 4 & 0 & 8 \\ \hline
\textbf{FOOD\_DRINKS\#QUALITY}           & 195 & 82 & 273 & 49 & 21 & 68 & 51 & 26 & 101 \\ \hline
\textbf{FOOD\_DRINKS\#STYLE\_OPTIONS}    & 54 & 10 & 82 & 13 & 3 & 20 & 14 & 1 & 22 \\ \hline
\textbf{HOTEL\#CLEANLINESS}              & 88 & 4 & 113 & 22 & 1 & 28 & 35 & 2 & 37 \\ \hline
\textbf{HOTEL\#COMFORT}                  & 60 & 3 & 89 & 15 & 1 & 24 & 29 & 1 & 36 \\ \hline
\textbf{HOTEL\#DESIGN\_FEATURES}         & 26 & 6 & 67 & 15 & 3 & 18 & 10 & 2 & 19 \\ \hline
\textbf{HOTEL\#GENERAL}                  & 305 & 82 & 773 & 84 & 21 & 205 & 101 & 33 & 220 \\ \hline
\textbf{HOTEL\#MISCELLANEOUS}            & 55 & 5 & 61 & 14 & 1 & 13 & 26 & 0 & 13 \\ \hline
\textbf{HOTEL\#PRICES}                   & 101 & 52 & 345 & 36 & 13 & 36 & 27 & 11 & 24 \\ \hline
\textbf{HOTEL\#QUALITY}                  & 104 & 10 & 266 & 23 & 5 & 22 & 40 & 2 & 25 \\ \hline
\textbf{LOCATION\#GENERAL}               & 86 & 54 & 582 & 22 & 14 & 183 & 22 & 11 & 270 \\ \hline
\textbf{ROOMS\#CLEANLINESS}              & 145 & 5 & 171 & 36 & 1 & 34 & 42 & 3 & 34 \\ \hline
\textbf{ROOMS\#COMFORT}                  & 47 & 2 & 66 & 12 & 0 & 15 & 4 & 0 & 17 \\ \hline
\textbf{ROOMS\#DESIGN\_FEATURES}         & 108 & 18 & 196 & 27 & 7 & 37 & 29 & 2 & 46 \\ \hline
\textbf{ROOMS\#GENERAL}                  & 85 & 38 & 330 & 21 & 10 & 33 & 19 & 13 & 45 \\ \hline
\textbf{ROOMS\#MISCELLANEOUS}            & 15 & 2 & 18 & 4 & 0 & 2 & 7 & 0 & 5 \\ \hline
\textbf{ROOMS\#PRICES}                   & 18 & 2 & 37 & 4 & 0 & 4 & 5 & 2 & 7 \\ \hline
\textbf{ROOMS\#QUALITY}                  & 74 & 3 & 87 & 19 & 1 & 9 & 24 & 4 & 4 \\ \hline
\textbf{ROOMS\_AMENITIES\#CLEANLINESS}   & 86 & 0 & 30 & 22 & 0 & 3 & 17 & 0 & 3 \\ \hline
\textbf{ROOMS\_AMENITIES\#COMFORT}       & 16 & 2 & 18 & 4 & 1 & 2 & 3 & 1 & 3 \\ \hline
\textbf{ROOMS\_AMENITIES\#DESIGN\_FEATURES} & 25 & 2 & 40 & 6 & 0 & 4 & 4 & 1 & 4 \\ \hline
\textbf{ROOMS\_AMENITIES\#GENERAL}       & 99 & 17 & 210 & 25 & 4 & 21 & 37 & 5 & 23 \\ \hline
\textbf{ROOMS\_AMENITIES\#MISCELLANEOUS} & 30 & 0 & 35 & 7 & 0 & 4 & 17 & 0 & 3 \\ \hline
\textbf{ROOMS\_AMENITIES\#PRICES}        & 4 & 0 & 15 & 1 & 0 & 2 & 2 & 0 & 0 \\ \hline
\textbf{ROOMS\_AMENITIES\#QUALITY}       & 113 & 8 & 100 & 28 & 2 & 5 & 33 & 1 & 3 \\ \hline
\textbf{SERVICE\#GENERAL}                & 542 & 78 & 2925 & 136 & 19 & 293 & 188 & 28 & 344 \\ \hline
\end{tabular}
\end{adjustbox}
\end{table*}

\begin{table*}
\centering
\caption{Detailed Performance of Arabic BERT Models with Optuna selected Hyperparameters on Aspect Sentiment Classification}
\label{tab:OptunaASC_arabic_bert_performance}
\begin{tabular}{|c|c|c|c|c|c|c|}
\toprule
Pre-trained BERT Model &  Number of Trials & Learning Rate &   Fine-Tune Method &  Batch Size &  Epochs & Accuracy \\
\midrule
                 QARiB &               500 &       5.0e-05 &   Full Fine-tuning &         256 &       3 &   87.54\% \\
                 QARiB &               500 &       9.4e-03 & Feature Extraction &          16 &       5 &   84.78\% \\
             MARBERTv2 &               500 &       5.1e-05 &   Full Fine-tuning &          64 &       2 &   87.30\% \\
             MARBERTv2 &               500 &       9.5e-03 & Feature Extraction &           8 &       5 &   83.08\% \\
         CAMeLBERT-MSA &               500 &       5.5e-05 &   Full Fine-tuning &          64 &       4 &   86.57\% \\
         CAMeLBERT-MSA &               500 &       9.4e-03 & Feature Extraction &          64 &       5 &   85.02\% \\
    Safaya Arabic BERT &               500 &       9.2e-05 &   Full Fine-tuning &          64 &       2 &   87.20\% \\
    Safaya Arabic BERT &               500 &       8.1e-03 & Feature Extraction &          32 &       8 &   82.40\% \\
           AraBERTv0.2 &               500 &       9.7e-03 & Feature Extraction &          64 &       6 &   85.46\% \\
           AraBERTv0.2 &               500 &       1.0e-04 &   Full Fine-tuning &         256 &       2 &   86.91\% \\
          CAMeLBERT-DA &               500 &       9.8e-05 &   Full Fine-tuning &          64 &       2 &   86.04\% \\
          CAMeLBERT-DA &               500 &       6.3e-03 & Feature Extraction &           8 &       8 &   81.68\% \\
         CAMeLBERT-Mix &               500 &       1.1e-04 &   Full Fine-tuning &         256 &       2 &   86.14\% \\
         CAMeLBERT-Mix &               500 &       9.9e-03 & Feature Extraction &          32 &       9 &   83.76\% \\
\bottomrule
\end{tabular}
\end{table*}

\begin{table*}[ht]

\centering
\caption{Detailed Performance of Arabic BERT Models with Various Hyperparameter Settings on Opinion Target Expression (OTE) Extraction. Note: Hyperparameter sources (HP sources) are based on configurations used in the original works of Arabic BERT models to test their performance on Named Entity Recognition (NER) tasks. Rows are ordered in descending order based on Micro-F1.}

\resizebox{\textwidth}{!}{%
\begin{tabular}{|l|l|l|c|c|c|c|c|c|c|}
\hline
\textbf{Pretrained Model} & \textbf{Fine-Tuning Method} & \textbf{Hyperparameter Source} & \textbf{Learning Rate} & \textbf{Batch Size} & \textbf{Epochs} & \textbf{Accuracy (\%)} & \textbf{Micro Precision (\%)} & \textbf{Micro Recall (\%)} & \textbf{Micro F1 Score (\%)} \\ \hline
MARBERTv2 & Full Fine-tuning & MARBERTv2-NER & $2.00 \times 10^{-6}$ & 32 & 25 & 95.76 & 77.49 & 76.36 & 76.92 \\ \hline
CAMEL-BERT-MSA & Full Fine-tuning & CAMEL-BERT-MSA-NER & $5.00 \times 10^{-5}$ & 32 & 10 & 95.66 & 77.74 & 75.67 & 76.69 \\ \hline
QARiB & Full Fine-tuning & mBERT NER & $5.00 \times 10^{-5}$ & 32 & 5 & 95.71 & 77.82 & 75.56 & 76.68 \\ \hline
CAMEL-BERT-MSA & Full Fine-tuning & CAMEL-BERT-MSA-NER & $5.00 \times 10^{-5}$ & 32 & 3 & 95.63 & 77.74 & 75.36 & 76.53 \\ \hline
AraBERTv0.2 & Full Fine-tuning & mBERT-NER & $5.00 \times 10^{-5}$ & 32 & 3 & 95.48 & 78.2 & 74.38 & 76.24 \\ \hline
AraBERTv0.2 & Full Fine-tuning & arabertv02-NER & $2.00 \times 10^{-5}$ & 32 & 3 & 95.62 & 76.6 & 75.65 & 76.12 \\ \hline
AraBERTv0.2 & Full Fine-tuning & mBERT NER & $5.00 \times 10^{-5}$ & 32 & 5 & 95.58 & 76.69 & 75.45 & 76.06 \\ \hline
CAMEL-BERT-MSA & Full Fine-tuning & CAMEL-BERT-MSA-NER & $5.00 \times 10^{-5}$ & 32 & 5 & 95.57 & 79.3 & 73.05 & 76.05 \\ \hline
QARiB & Full Fine-tuning & mBERT NER & $5.00 \times 10^{-5}$ & 32 & 3 & 95.74 & 72.51 & 79.63 & 75.9 \\ \hline
Asafaya Arabic-BERT & Full Fine-tuning & mBERT NER & $5.00 \times 10^{-5}$ & 32 & 3 & 95.5 & 76.35 & 74.65 & 75.49 \\ \hline
QARiB & Full Fine-tuning & bert-base-qarib-NER & $8.00 \times 10^{-5}$ & 64 & 3 & 95.42 & 76.85 & 74.14 & 75.47 \\ \hline
Asafaya Arabic-BERT & Full Fine-tuning & asafaya-bert-base-arabic-NER & $2.00 \times 10^{-5}$ & 32 & 10 & 95.37 & 76.77 & 74.05 & 75.39 \\ \hline
MARBERTv2 & Full Fine-tuning & mBERT-NER & $5.00 \times 10^{-5}$ & 32 & 3 & 95.58 & 73.31 & 77.39 & 75.3 \\ \hline
MARBERTv2 & Full Fine-tuning & mBERT-NER & $5.00 \times 10^{-5}$ & 32 & 5 & 95.47 & 76.43 & 74.18 & 75.29 \\ \hline
Asafaya Arabic-BERT & Full Fine-tuning & mBERT-NER & $5.00 \times 10^{-5}$ & 32 & 5 & 95.29 & 78.04 & 72.67 & 75.26 \\ \hline
Asafaya Arabic-BERT & Full Fine-tuning & mBERT-NER & $5.00 \times 10^{-5}$ & 32 & 3 & 95.39 & 76.18 & 73.79 & 74.96 \\ \hline
CAMEL-BERT-MIX & Full Fine-tuning & CAMEL-BERT-MIX-NER & $5.00 \times 10^{-5}$ & 32 & 10 & 95.35 & 77.45 & 72.24 & 74.75 \\ \hline
CAMEL-BERT-MIX & Full Fine-tuning & mBERT NER & $5.00 \times 10^{-5}$ & 32 & 5 & 95.23 & 78.5 & 70.72 & 74.41 \\ \hline
CAMEL-BERT-DA & Full Fine-tuning & mBERT NER & $5.00 \times 10^{-5}$ & 32 & 5 & 95.05 & 77.32 & 70.08 & 73.52 \\ \hline
CAMEL-BERT-DA & Full Fine-tuning & CAMEL-BERT-DA & $5.00 \times 10^{-5}$ & 32 & 10 & 94.95 & 77.82 & 69.16 & 73.24 \\ \hline
QARiB & Feature Extraction & Custom\_Selected & $2.00 \times 10^{-3}$ & 32 & 5 & 94.9 & 70.32 & 71.62 & 70.96 \\ \hline
QARiB & Feature Extraction & Custom\_Selected & $2.00 \times 10^{-3}$ & 32 & 10 & 94.85 & 70.57 & 71.33 & 70.95 \\ \hline
QARiB & Feature Extraction & Custom\_Selected & $9.40 \times 10^{-3}$ & 32 & 5 & 94.81 & 70.74 & 70.86 & 70.8 \\ \hline
CAMEL-BERT-MSA & Feature Extraction & Custom\_Selected & $2.00 \times 10^{-3}$ & 32 & 5 & 94.87 & 69.73 & 71.39 & 70.55 \\ \hline
CAMEL-BERT-MSA & Feature Extraction & Custom\_Selected & $9.40 \times 10^{-3}$ & 32 & 5 & 94.79 & 70.24 & 70.83 & 70.53 \\ \hline
CAMEL-BERT-MSA & Feature Extraction & Custom\_Selected & $2.00 \times 10^{-3}$ & 32 & 10 & 94.85 & 69.73 & 71.17 & 70.44 \\ \hline
AraBERTv0.2 & Feature Extraction & Custom\_Selected & $2.00 \times 10^{-3}$ & 32 & 10 & 94.72 & 70.11 & 70.65 & 70.38 \\ \hline
QARiB & Feature Extraction & Custom\_Selected & $9.40 \times 10^{-3}$ & 32 & 10 & 94.83 & 69.48 & 71.25 & 70.35 \\ \hline
CAMEL-BERT-MSA & Feature Extraction & Custom\_Selected & $9.40 \times 10^{-3}$ & 32 & 10 & 94.84 & 70.03 & 70.53 & 70.28 \\ \hline
AraBERTv0.2 & Feature Extraction & Custom\_Selected & $9.40 \times 10^{-3}$ & 32 & 10 & 94.75 & 69.27 & 71.31 & 70.27 \\ \hline
AraBERTv0.2 & Feature Extraction & Custom\_Selected & $9.40 \times 10^{-3}$ & 32 & 5 & 94.73 & 70.19 & 70.34 & 70.27 \\ \hline
CAMEL-BERT-MIX & Feature Extraction & Custom\_Selected & $9.40 \times 10^{-3}$ & 32 & 5 & 94.84 & 69.1 & 71.11 & 70.09 \\ \hline
CAMEL-BERT-MIX & Feature Extraction & Custom\_Selected & $2.00 \times 10^{-3}$ & 32 & 10 & 94.72 & 68.3 & 71.4 & 69.81 \\ \hline
AraBERTv0.2 & Feature Extraction & Custom\_Selected & $2.00 \times 10^{-3}$ & 32 & 5 & 94.68 & 68.76 & 70.67 & 69.7 \\ \hline
CAMEL-BERT-MIX & Feature Extraction & Custom\_Selected & $9.40 \times 10^{-3}$ & 32 & 10 & 94.6 & 70.7 & 68.17 & 69.41 \\ \hline
MARBERTv2 & Feature Extraction & Custom\_Selected & $9.40 \times 10^{-3}$ & 32 & 10 & 94.68 & 67.79 & 70.99 & 69.36 \\ \hline
Asafaya Arabic-BERT & Feature Extraction & Custom\_Selected & $9.40 \times 10^{-3}$ & 32 & 10 & 94.27 & 72.47 & 65.91 & 69.04 \\ \hline
Asafaya Arabic-BERT & Feature Extraction & Custom\_Selected & $2.00 \times 10^{-3}$ & 32 & 10 & 94.47 & 68.72 & 69.07 & 68.89 \\ \hline
Asafaya Arabic-BERT & Feature Extraction & Custom\_Selected & $9.40 \times 10^{-3}$ & 32 & 5 & 94.41 & 69.01 & 67.62 & 68.31 \\ \hline
MARBERTv2 & Feature Extraction & Custom\_Selected & $9.40 \times 10^{-3}$ & 32 & 5 & 94.51 & 67.12 & 69.55 & 68.31 \\ \hline
CAMEL-BERT-DA & Feature Extraction & Custom\_Selected & $9.40 \times 10^{-3}$ & 32 & 10 & 94.36 & 67.07 & 67.88 & 67.47 \\ \hline
CAMEL-BERT-DA & Feature Extraction & Custom\_Selected & $9.40 \times 10^{-3}$ & 32 & 5 & 94.3 & 67.45 & 67.2 & 67.33 \\ \hline
Asafaya Arabic-BERT & Feature Extraction & Custom\_Selected & $2.00 \times 10^{-3}$ & 32 & 5 & 94.39 & 65.39 & 69.3 & 67.29 \\ \hline
CAMEL-BERT-DA & Feature Extraction & Custom\_Selected & $2.00 \times 10^{-3}$ & 32 & 10 & 94.29 & 67.07 & 67.33 & 67.2 \\ \hline
CAMEL-BERT-DA & Feature Extraction & Custom\_Selected & $2.00 \times 10^{-3}$ & 32 & 5 & 94.03 & 67.83 & 64.8 & 66.28 \\ \hline
MARBERTv2 & Feature Extraction & Custom\_Selected & $2.00 \times 10^{-3}$ & 32 & 10 & 94.3 & 61.76 & 69.66 & 65.47 \\ \hline
MARBERTv2 & Feature Extraction & Custom\_Selected & $2.00 \times 10^{-3}$ & 32 & 5 & 93.62 & 52.95 & 67.75 & 59.44 \\ \hline
\end{tabular}%
}
\label{tab:arabic_bert_ote_full_hyperparameter_performance}
\end{table*}

\begin{table*}[ht]

\centering
\caption{Detailed Performance of Arabic BERT Models with Various Hyperparameter Settings on Aspect Sentiment Classification. Note: Hyperparameter sources (HP sources) are based on configurations used in the original works of Arabic BERT models to test their performance on sentence-level sentiment analysis tasks. Rows are ordered in descending order based on accuracy.}

\resizebox{\textwidth}{!}{%
\begin{tabular}{|l|l|l|c|c|c|c|c|c|c|}
\hline
\textbf{Pretrained Model} & \textbf{Fine-Tuning Method} & \textbf{Hyperparameter Source} & \textbf{Learning Rate} & \textbf{Batch Size} & \textbf{Epochs} & \textbf{Accuracy (\%)} & \textbf{Macro Precision (\%)} & \textbf{Macro Recall (\%)} & \textbf{Macro F1 Score (\%)} \\ \hline
MARBERTv2 & Full Fine-tuning & MARBERTv2-SA & $2.00 \times 10^{-6}$ & 32 & 25 & 87.35 & 69.58 & 64.15 & 64.16 \\ \hline
MARBERTv2 & Full Fine-tuning & m-BERT-SA & $5.00 \times 10^{-5}$ & 32 & 5 & 87.3 & 75.70 & 64.59 & 65.61 \\ \hline
QARiB & Full Fine-tuning & m-BERT-SA & $5.00 \times 10^{-5}$ & 32 & 5 & 87.25 & 73.52 & 69.14 & 70.60 \\ \hline
Asafaya Arabic BERT & Full Fine-tuning & m-BERT-SA & $5.00 \times 10^{-5}$ & 32 & 5 & 87.01 & 74.56 & 71.22 & 72.38 \\ \hline
AraBERTv0.2 & Full Fine-tuning & m-BERT-SA & $5.00 \times 10^{-5}$ & 32 & 5 & 87.01 & 73.92 & 69.18 & 70.48 \\ \hline
CAMeLBERT-MSA & Full Fine-tuning & m-BERT-SA & $5.00 \times 10^{-5}$ & 32 & 5 & 86.91 & 71.78 & 68.40 & 69.41 \\ \hline
Asafaya Arabic BERT & Full Fine-tuning & Asafaya Arabic BERT-SA & $2.00 \times 10^{-5}$ & 32 & 10 & 86.86 & 73.39 & 70.56 & 71.52 \\ \hline
CAMeLBERT-MSA & Full Fine-tuning & m-BERT-SA & $5.00 \times 10^{-5}$ & 32 & 3 & 86.57 & 71.52 & 69.17 & 69.90 \\ \hline
CAMeLBERT-DA & Full Fine-tuning & CAMeLBERT-DA-SA & $3.00 \times 10^{-5}$ & 32 & 5 & 86.43 & 71.70 & 70.54 & 71.03 \\ \hline
Asafaya Arabic BERT & Full Fine-tuning & m-BERT-SA & $5.00 \times 10^{-5}$ & 32 & 3 & 86.33 & 70.47 & 66.29 & 67.45 \\ \hline
AraBERTv0.2 & Full Fine-tuning & m-BERT-SA & $5.00 \times 10^{-5}$ & 32 & 3 & 86.23 & 72.09 & 71.43 & 71.72 \\ \hline
AraBERTv0.2 & Full Fine-tuning & AraBERTv0.2-SA & $2.00 \times 10^{-5}$ & 32 & 3 & 86.14 & 70.27 & 65.78 & 66.55 \\ \hline
MARBERTv2 & Full Fine-tuning & m-BERT-SA & $5.00 \times 10^{-5}$ & 32 & 3 & 86.09 & 71.44 & 70.40 & 70.51 \\ \hline
CAMeLBERT-MSA & Full Fine-tuning & CAMeLBERT-MSA-SA & $3.00 \times 10^{-5}$ & 32 & 5 & 86.09 & 70.61 & 69.24 & 69.69 \\ \hline
CAMeLBERT-DA & Full Fine-tuning & m-BERT-SA & $5.00 \times 10^{-5}$ & 32 & 5 & 85.99 & 72.59 & 72.87 & 72.53 \\ \hline
CAMeLBERT-MIX & Full Fine-tuning & m-BERT-SA & $5.00 \times 10^{-5}$ & 32 & 5 & 85.99 & 71.98 & 68.65 & 69.37 \\ \hline
AraBERTv0.2 & Full Fine-tuning & AraBERTv0.2-SA & $2.00 \times 10^{-5}$ & 32 & 5 & 85.8 & 69.96 & 67.45 & 68.07 \\ \hline
QARiB & Full Fine-tuning & m-BERT-SA & $5.00 \times 10^{-5}$ & 32 & 3 & 85.65 & 71.37 & 71.32 & 71.26 \\ \hline
AraBERTv0.2 & Feature Extraction & Custom-Selected & $9.40 \times 10^{-3}$ & 32 & 10 & 85.65 & 70.60 & 62.76 & 62.35 \\ \hline
CAMeLBERT-MSA & Full Fine-tuning & CAMeLBERT-MSA-SA & $3.00 \times 10^{-5}$ & 32 & 3 & 85.6 & 71.05 & 72.42 & 71.68 \\ \hline
AraBERTv0.2 & Feature Extraction & Custom-Selected & $2.00 \times 10^{-3}$ & 32 & 10 & 85.56 & 77.68 & 62.07 & 61.06 \\ \hline
AraBERTv0.2 & Feature Extraction & Custom-Selected & $9.40 \times 10^{-3}$ & 32 & 5 & 85.46 & 70.97 & 63.13 & 62.93 \\ \hline
AraBERTv0.2 & Feature Extraction & Custom-Selected & $2.00 \times 10^{-3}$ & 32 & 5 & 85.36 & 66.24 & 61.99 & 60.95 \\ \hline
QARiB & Feature Extraction & Custom-Selected & $9.40 \times 10^{-3}$ & 32 & 10 & 85.31 & 73.51 & 62.00 & 61.30 \\ \hline
CAMeLBERT-MIX & Full Fine-tuning & m-BERT-SA & $5.00 \times 10^{-5}$ & 32 & 3 & 85.17 & 71.07 & 71.88 & 71.29 \\ \hline
CAMeLBERT-MIX & Full Fine-tuning & CAMeLBERT-MIX-SA & $3.00 \times 10^{-5}$ & 32 & 3 & 85.02 & 70.22 & 70.96 & 70.57 \\ \hline
CAMeLBERT-DA & Feature Extraction & Custom-Selected & $9.40 \times 10^{-3}$ & 32 & 10 & 85.01 & 61.52 & 60.97 & 59.13 \\ \hline
QARiB & Feature Extraction & Custom-Selected & $2.00 \times 10^{-3}$ & 64 & 10 & 84.92 & 56.29 & 60.67 & 58.39 \\ \hline
QARiB & Feature Extraction & Custom-Selected & $2.00 \times 10^{-3}$ & 32 & 10 & 84.92 & 66.01 & 60.95 & 59.33 \\ \hline
QARiB & Feature Extraction & Custom-Selected & $9.40 \times 10^{-3}$ & 32 & 5 & 84.78 & 65.39 & 61.36 & 60.11 \\ \hline
CAMeLBERT-MIX & Full Fine-tuning & CAMeLBERT-MIX-SA & $3.00 \times 10^{-5}$ & 32 & 5 & 84.78 & 69.94 & 70.12 & 70.03 \\ \hline
CAMeLBERT-MIX & Feature Extraction & Custom-Selected & $9.40 \times 10^{-3}$ & 32 & 5 & 84.72 & 68.69 & 61.13 & 60.09 \\ \hline
CAMeLBERT-DA & Full Fine-tuning & m-BERT-SA & $5.00 \times 10^{-5}$ & 32 & 3 & 84.59 & 69.85 & 68.30 & 68.99 \\ \hline
CAMeLBERT-DA & Feature Extraction & Custom-Selected & $9.40 \times 10^{-3}$ & 32 & 5 & 84.49 & 61.00 & 60.45 & 58.61 \\ \hline
QARiB & Feature Extraction & Custom-Selected & $2.00 \times 10^{-3}$ & 32 & 5 & 84.44 & 72.65 & 60.35 & 58.45 \\ \hline
CAMeLBERT-MSA & Feature Extraction & Custom-Selected & $9.40 \times 10^{-3}$ & 32 & 10 & 84.29 & 55.88 & 60.21 & 57.96 \\ \hline
QARiB & Feature Extraction & Custom-Selected & $2.00 \times 10^{-3}$ & 64 & 5 & 84.29 & 55.92 & 60.20 & 57.97 \\ \hline
CAMeLBERT-MSA & Feature Extraction & Custom-Selected & $2.00 \times 10^{-3}$ & 32 & 10 & 84.2 & 55.94 & 59.93 & 57.87 \\ \hline
CAMeLBERT-MIX & Feature Extraction & Custom-Selected & $9.40 \times 10^{-3}$ & 32 & 5 & 84.2 & 68.17 & 60.61 & 59.57 \\ \hline
QARiB & Full Fine-tuning & QARiB-SA & $8.00 \times 10^{-5}$ & 64 & 3 & 84.1 & 69.54 & 71.21 & 70.22 \\ \hline
CAMeLBERT-DA & Feature Extraction & Custom-Selected & $2.00 \times 10^{-3}$ & 32 & 5 & 84.0 & 55.74 & 59.96 & 57.76 \\ \hline
CAMeLBERT-DA & Feature Extraction & Custom-Selected & $2.00 \times 10^{-3}$ & 32 & 10 & 84.0 & 55.87 & 59.68 & 57.71 \\ \hline
MARBERTv2 & Feature Extraction & Custom-Selected & $9.40 \times 10^{-3}$ & 32 & 10 & 83.91 & 75.13 & 60.11 & 59.36 \\ \hline
CAMeLBERT-MSA & Feature Extraction & Custom-Selected & $2.00 \times 10^{-3}$ & 32 & 5 & 83.91 & 55.62 & 59.92 & 57.68 \\ \hline
CAMeLBERT-MSA & Feature Extraction & Custom-Selected & $9.40 \times 10^{-3}$ & 32 & 5 & 83.76 & 55.77 & 59.52 & 57.58 \\ \hline
CAMeLBERT-MIX & Feature Extraction & Custom-Selected & $2.00 \times 10^{-3}$ & 32 & 10 & 83.76 & 55.82 & 59.31 & 57.50 \\ \hline
CAMeLBERT-MIX & Feature Extraction & Custom-Selected & $2.00 \times 10^{-3}$ & 32 & 5 & 83.57 & 55.54 & 59.22 & 57.31 \\ \hline
MARBERTv2 & Feature Extraction & Custom-Selected & $9.40 \times 10^{-3}$ & 32 & 5 & 83.52 & 72.30 & 59.26 & 57.73 \\ \hline
MARBERTv2 & Feature Extraction & Custom-Selected & $2.00 \times 10^{-3}$ & 32 & 10 & 82.26 & 54.83 & 58.03 & 56.34 \\ \hline
Asafaya Arabic BERT & Feature Extraction & Custom-Selected & $9.40 \times 10^{-3}$ & 32 & 10 & 82.26 & 59.29 & 58.19 & 56.89 \\ \hline
Asafaya Arabic BERT & Feature Extraction & Custom-Selected & $2.00 \times 10^{-3}$ & 32 & 10 & 81.58 & 54.68 & 57.34 & 55.88 \\ \hline
Asafaya Arabic BERT & Feature Extraction & Custom-Selected & $9.40 \times 10^{-3}$ & 32 & 5 & 81.29 & 60.44 & 58.28 & 57.11 \\ \hline
MARBERTv2 & Feature Extraction & Custom-Selected & $2.00 \times 10^{-3}$ & 32 & 5 & 81.24 & 54.31 & 56.97 & 55.50 \\ \hline
Asafaya Arabic BERT & Feature Extraction & Custom-Selected & $2.00 \times 10^{-3}$ & 32 & 5 & 80.56 & 53.65 & 56.67 & 55.07 \\ \hline
\end{tabular}%
}
\label{tab:arabic_bert_aspect_sentiment_classification}
\end{table*}

\end{document}